\documentclass{article}
\usepackage{caption}
\usepackage[accepted]{icml2025}
\usepackage{graphicx}
\usepackage{microtype}
\usepackage{hyperref}
\usepackage{xurl}
\usepackage{booktabs}
\usepackage{xcolor}
\usepackage{lipsum}
\usepackage{amsmath}
\usepackage{tabularx}
\usepackage{calc}
\usepackage{subcaption}
\usepackage[nomessages]{fp}
\usepackage{amsthm, amssymb}
\usepackage{caption, comment}
\usepackage{cleveref}

\newtheorem{definition}{Definition}

\date{}

\icmltitlerunning{Grokking vs. Learning}

\begin{document}

\twocolumn[
\icmltitle{Grokking vs. Learning: Same features, different encodings}




\begin{icmlauthorlist}
\icmlauthor{Dmitry Manning-Coe}{uiuc}
\icmlauthor{Jacopo Gliozzi}{uiuc}
\icmlauthor{Alexander G. Stapleton}{qm}
\icmlauthor{Edward Hirst}{qm}
\icmlauthor{Giuseppe De Tomasi}{uiuc,lb}
\icmlauthor{Barry Bradlyn}{uiuc}
\icmlauthor{David Berman}{qm}

\end{icmlauthorlist}

\icmlaffiliation{uiuc}{Anthony J. Leggett Institute for Condensed Matter Theory, University of Illinois Urbana-Champaign, Urbana, USA}
\icmlaffiliation{qm}{Centre for Theoretical Physics, Queen Mary University of London, UK}
\icmlaffiliation{lb}{CeFEMA-LaPMET, Departamento de Física, Instituto Superior Técnico, Universidade de Lisboa,
Lisboa, Portugal}

\icmlcorrespondingauthor{Dmitry Manning-Coe}{dmitry2@illinois.edu}

\icmlkeywords{Information Theory, Grokking}

\vskip 0.3in
]
\printAffiliationsAndNotice{}

\begin{abstract}
Grokking typically achieves similar loss to ordinary, “steady”, learning. We ask whether these different learning paths - grokking versus ordinary training - lead to fundamental differences in the learned models. To do so we compare the features, compressibility, and learning dynamics of models trained via each path in two tasks. We find that grokked and steadily trained models learn the same features, but there can be large differences in the efficiency with which these features are encoded. In particular, we find a novel ``compressive regime" of steady training in which there emerges a linear trade-off between model loss and compressibility, and which is absent in grokking. In this regime, we can achieve compression factors 25x the base model, and 5x the compression achieved in grokking. We then track how model features and compressibility develop through training. We show that model development in grokking is task-dependent, and that peak compressibility is achieved immediately after the grokking plateau. Finally, novel information-geometric measures are introduced which demonstrate that models undergoing grokking follow a straight path in information space.
\end{abstract}

\section{Introduction}
Grokking ~\cite{power2022grokking} is a mode of training in which generalisation emerges suddenly after a long period of overfitting. We empirically address two important questions that this raises. First, 
are there fundamental differences between models learned by grokking and through ordinary ``steady" training? Since grokking happens after a large number of epochs, it is expensive. Given this, it is natural to also ask: are there any compensating practical advantages?

Previous work has observed grokking in a variety of tasks ~\cite{tegmark2023omnigrok}, and several mechanisms have been identified as its possible origin~\cite{thilak2022slingshotmechanismempiricalstudy, tegmark2023omnigrok, kumar2024lazydynamics, tan2024understandinggrokkingrobustnessviewpoint, Humayun_2024, Kozyrev_2025}. For modular addition, the workhorse of grokking studies, an exactly solvable model of grokking has been proposed~\cite{gromov2024_grokking}, and network interpretability measures can track how overfitting gives way to generalisation~\cite{nanda2023progress}.

All of these works contrast the generalising, post-grokking model to the memorizing, pre-grokking model of the same training run. This leaves it unclear whether grokking is simply a slower form of ordinary learning, or whether there are more fundamental differences between these two learning paths. Explanations of grokking typically revolve around two points. First, that there are multiple representations that can solve the task, and second, that the model generalises as a result of compressing to a more efficient representation. This raises two immediate questions: do grokking and conventional learning lead to different representations, and are there differences in compression between grokking and learning?

We hence compare the features learned and the compressibility of models trained via grokking and conventional learning. To make this comparison, we choose two paradigmatic tasks which are known to have interpretable features. 
Our first task is to classify snapshots of the two-dimensional Ising model~\cite{Ising1925, baxter2007exactly}. This task can be interpreted in terms of physical variables - the energy and the magnetization of a snapshot~\cite{hu_ising_int_2017,isingint_suchs_2018}. Our second task is modular addition. Networks trained on modular addition are well known to learn a Fourier representation of the problem, and previous work has explicitly quantified how sharply the model is localized in this basis~\cite{nanda2023progress,gromov2024_grokking, gromov_pruning_paper}.

After comparing the terminal models, we then ask whether we can identify differences in model development through training. To do so, we define summary measures of model development and track their evolution. These measures are task-dependent and must be constructed by hand. We then show that important features of model development can be seen in the \textit{information geometry} of the model, which can be defined for a general task. Using the recently developed framework of Bayesian renormalization, ~\cite{amari_fimdiagapprox,Berman:2022uov, Berman:2023rqb,Berman:2024pax, Berman:2023rqb}, we introduce novel measures based on the Fisher Information Metric (FIM) to study model development in our example tasks.

We draw the following conclusions:
\begin{enumerate}
    \item In both tasks, the features learned via grokking and steady learning are the same.
    
    \item The encodings, however, are different. In the modular addition task, we discover what we call a ``compressive regime" of training. In this regime, tuning the weight scale at initialization results in a linear trade-off between the \textit{train} loss of a model and its compressibility.
    
     \item The modular addition and Ising tasks have different model development trajectories. In modular addition, consistent with previous literature, our interpretability tools allows us to define ``progress measures" in the grokking plateau. In the Ising task, however, we find no sign of model development before generalisation. This raises the question of whether we can define new progress measures that would indicate a coming jump in capability before generalisation.
    
    \item In both tasks, the grokking trajectory follows an approximately straight line in model space, as defined by the Fisher Information Metric. 
\end{enumerate}

The first two points are about the terminal model after training, and are addressed in Sections 4 and 5, respectively. The latter three points concern the dynamics during training and are the subject of Section 6. For convenience, an introduction to our two tasks is provided in Section 2, and our grokking methodology is outlined in Section 3. 

\section{Task Descriptions: Ising and Modular Addition}

\subsection{The Ising task}
Our first task is classifying the phases of the two-dimensional Ising model, perhaps the simplest physical system that exhibits a phase transition~\cite{Ising1925, baxter2007exactly}. On each vertex of a two-dimensional lattice, the model hosts a vector called a ``spin'' that can either point up or down.
This system has two distinct phases: an ``ordered" phase where the spins align with each other, and a ``disordered" phase where the spins are randomly oriented. 

\begin{figure}[t]
    \includegraphics[width=1\linewidth]{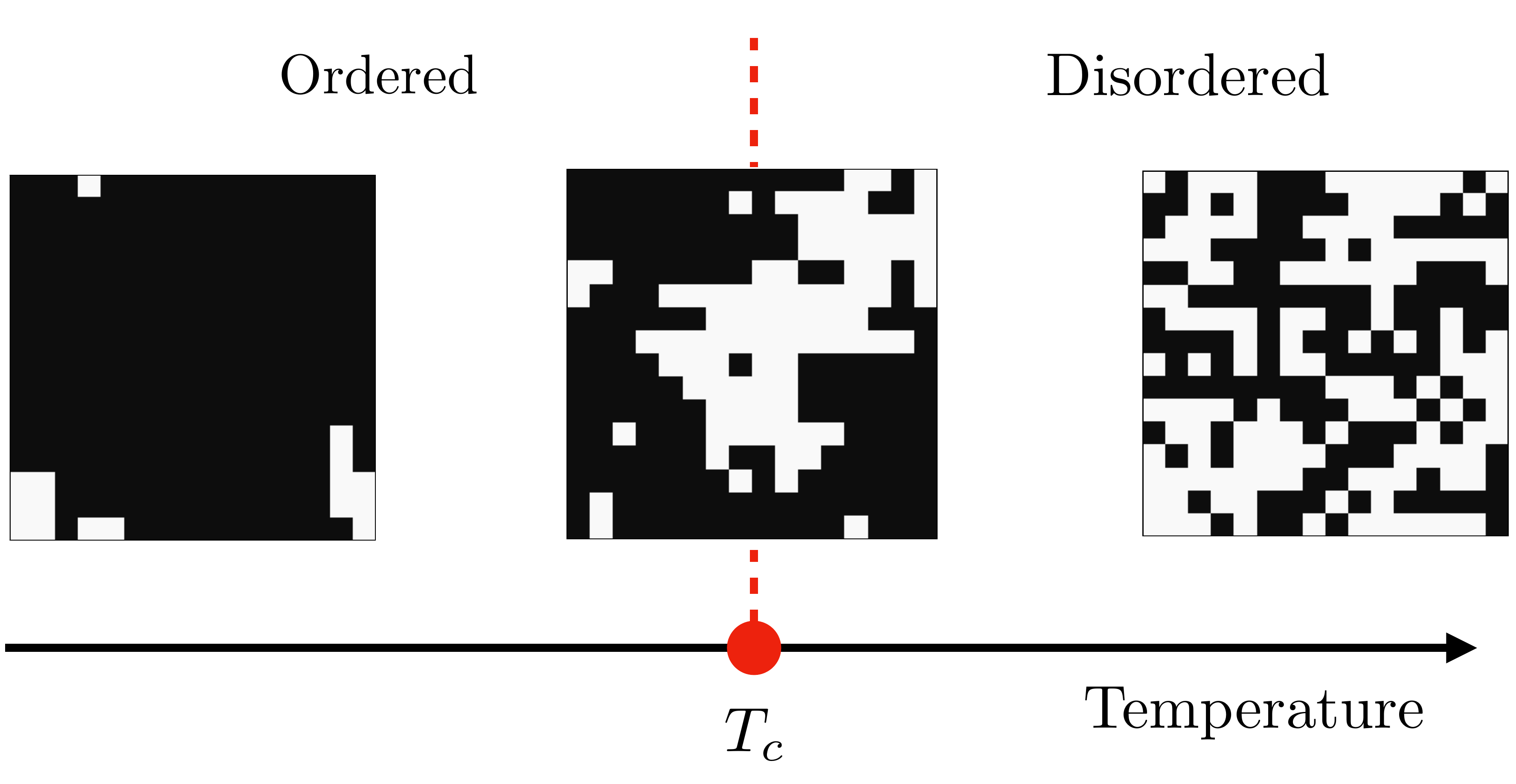}
    \caption{Snapshots of the 16 x 16 Ising lattice, where white and black squares indicate $\pm 1$ spin assignments. As temperature increases, snapshots become more disordered,  passing through a phase transition at temperature $T_c$.}
    \label{fig:snapshots}
\end{figure}

The two phases are separated by a transition that occurs at a critical temperature, $T_c$. For $T < T_c$, the system is in an ordered phase, while for $T > T_c$, it is in a disordered phase. In our model, we consider a square grid where each site $i$ contains a spin of value $\sigma_i=\pm 1$, corresponding to up and down. We call an assignment of $\sigma_i=\pm 1$ to every site $i$ in the grid a ``snapshot". A snapshot is hence just a binary image, and in Fig.~\ref{fig:snapshots} we show representative snapshots for the different phases.
 
Snapshots are characterized by two physical quantities: energy and magnetization. The energy is defined as a sum over nearest-neighbour pairs of spins,
\begin{equation}
\label{eq:Energy}
    E=-\sum_{\langle i,j\rangle}\sigma_i\sigma_j,
\end{equation}
where $\langle i, j\rangle$ are adjacent sites and periodic boundary conditions are used. The energy determines the probability distribution of different snapshot configurations (see Appendix B for more details).
Based on Eq.~\eqref{eq:Energy}, neighbouring spins that are misaligned ``cost'' positive energy. Disordered snapshots have many misaligned spins, and therefore a higher energy than ordered snapshots.  
The magnetization of a snapshot is the sum of all of its spins:
\begin{equation}
\label{eq:Magnetization}
    M=\sum_{i}\sigma_i
\end{equation}
In the disordered phase, opposite spins cancel in the sum and $M$ is small compared to the number of spins. In the ordered phase, spins align with each other and $|M|$ is large. Viewing a snapshot as an image, the magnetization quantifies its brightness and the energy captures the number of light to dark interfaces. Classifying snapshots is therefore analogous to the problem of image recognition. 
\\

In our first task, we train a Convolutional Neural Network (CNN) to classify a given snapshot of the Ising model as ``ordered" or ``disordered". Concretely, our input data $\mathbf{x}$ is a $16\times 16$ square grid of $\pm 1 $ values, and our labels are $y=0,1$ corresponding to whether the phase is ordered or disordered. 
For all Ising classification tasks in the main text, we use two convolutional layers and one fully connected layer, ReLU activations, and optimize the cross entropy loss with the Adam optimizer \cite{adampaper2017}.
Our training data are 300 snapshots equally divided between disordered and ordered configurations. The snapshots are generated by running a local-update Monte Carlo simulation at different temperatures surrounding the critical point between the two phases (see Fig.~\ref{fig:snapshots}). 

In an infinite system at equilibrium, the magnetization of a snapshot uniquely characterizes its phase. However, because our snapshots are of a small system that may not be fully equilibrated due to the local update Monte Carlo simulations, the classification problem is more difficult than simply counting how many spins have a positive sign. In particular, ordered snapshots with an anomalously low magnetization can be misclassified if their energy is not taken into account.

\subsection{Modular Addition}
Our second task is modular addition. We train models to solve modular equations of the form $c=(a+b)\text{ } \% \text{ }P$. Our input data is the set of $P^2$ two-hot encoded vectors corresponding to the pair $a$ and $b$, i.e. the length $2P$ vectors $\mathbf{x}_{a,b}=(0_1,...,1_a,...,0_{P},0_{P+1},...,1_{P+b},...,0_{2P})$, and our output vectors are the one-hot encoded length $P$ vectors corresponding to $c$, $\mathbf{y_c}=(0_1,...,1_{c},...,0_P)$ . We train a fully connected network with a single hidden layer and ReLU activation, minimizing the cross-entropy loss using the Adam optimizer for $P=113$. Our training data is a fixed fraction ($70\%$ in the main text) of randomly chosen samples from the set of all possible pairs. The full details of the model and data for both tasks is in the Appendix \cref{tab:model-params}.

\section{Inducing grokking}
We first show that we can tune between grokking and conventional learning regimes while maintaining the same training data, architecture, and regularization. This allows us to generate comparable models in each regime which differ in whether they were trained via grokking or steady learning. In subsequent sections, we will compare the features, compression, and development of these models.

\begin{figure}[t]
    \centering
    \includegraphics[width=1\linewidth]{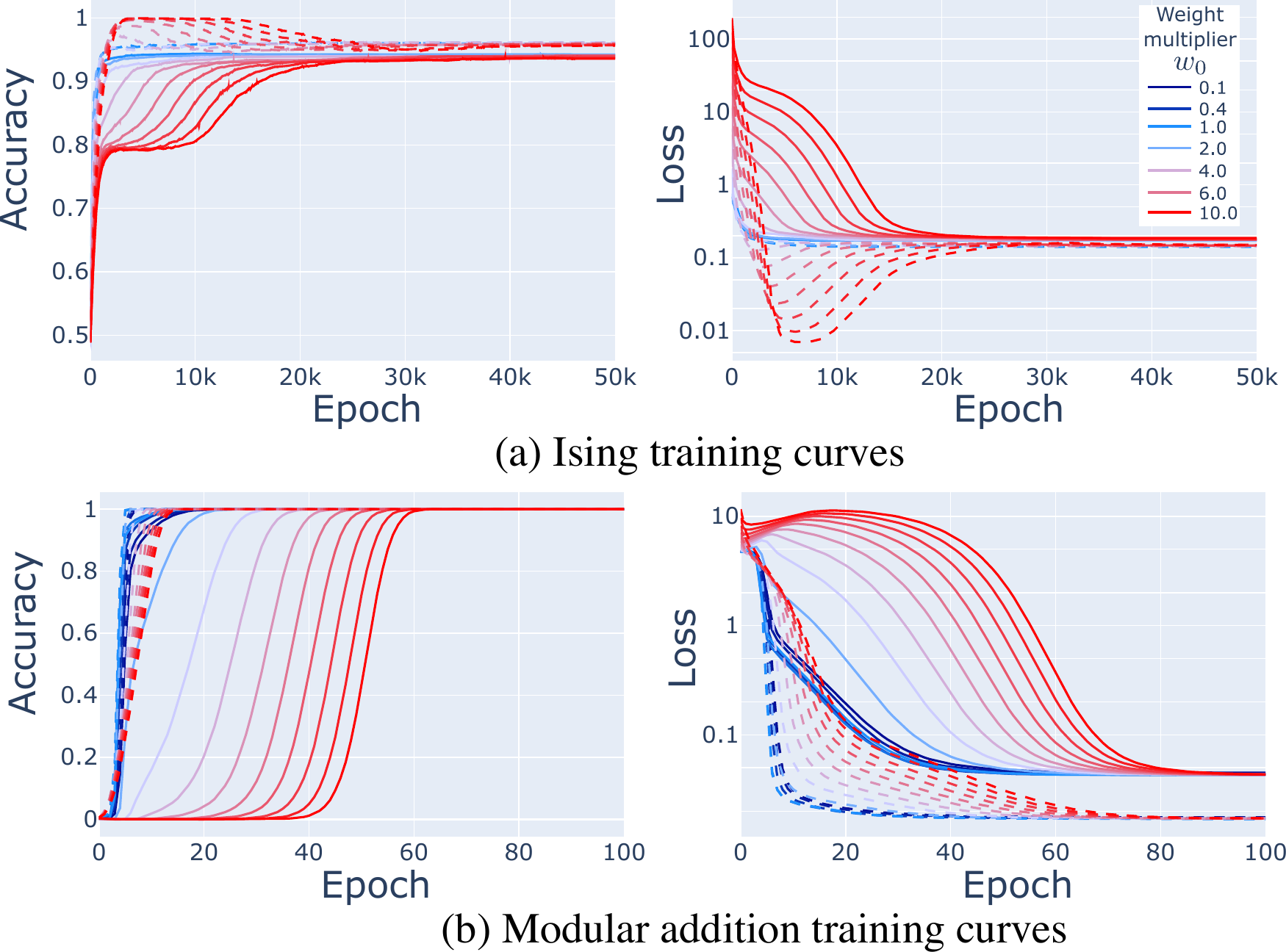}
    \caption{Ising and modular addition training curves, for varying weight multipliers at initialization (larger values are redder). The dashed lines indicate training accuracies/losses, and the solid lines test accuracies/losses.}
    \label{fig:traincurves}
\end{figure}

We use a pragmatic definition of the \textit{grokking time}:
\begin{definition}[Grokking time]
    The grokking time is the number of training epochs between when the model is within 5\% of its maximum test accuracy, $t_{\text{test}}$, and when the model is within $5\%$ of its maximum train accuracy, $t_{\text{train}}$:
    \begin{equation}
        t_{\text{grok}}\equiv t_{test}-t_{train}
    \end{equation}
\end{definition}
from which we define grokking as:
\begin{definition}[Grokking]
    A training run is said to grok if $t_{\text{grok}} > t_{\text{train}}$. Otherwise, we say that the model ``steadily learns".
\end{definition}
Following ~\cite{tegmark2023omnigrok}, we induce grokking by simply multiplying the initialization weights by a scale factor $w_0$, favouring overfitting during training. 
The resulting training curves for the Ising task are shown in Fig.~\ref{fig:traincurves}(a) and for the modular addition task in Fig.~\ref{fig:traincurves}(b). The curves are averaged over five and ten seeds (random initialization), respectively, and we summarize the distribution over individual seeds in Appendix C. Increasing the weight initialization scale $w_0$ smoothly interpolates between the grokking and steady learning regimes. According to our definition, grokking sets approximately at $w_0=3$ for the Ising task, and $w_0=0.4$ for the modular addition task.

\section{Grokked and learned models have the same features}

\label{sec:same_features}
\subsection{Learning Ising phases: energy and magnetization are encoded in the neuron activations}
\begin{figure}[t]
    \centering
    \includegraphics[width=1\linewidth]{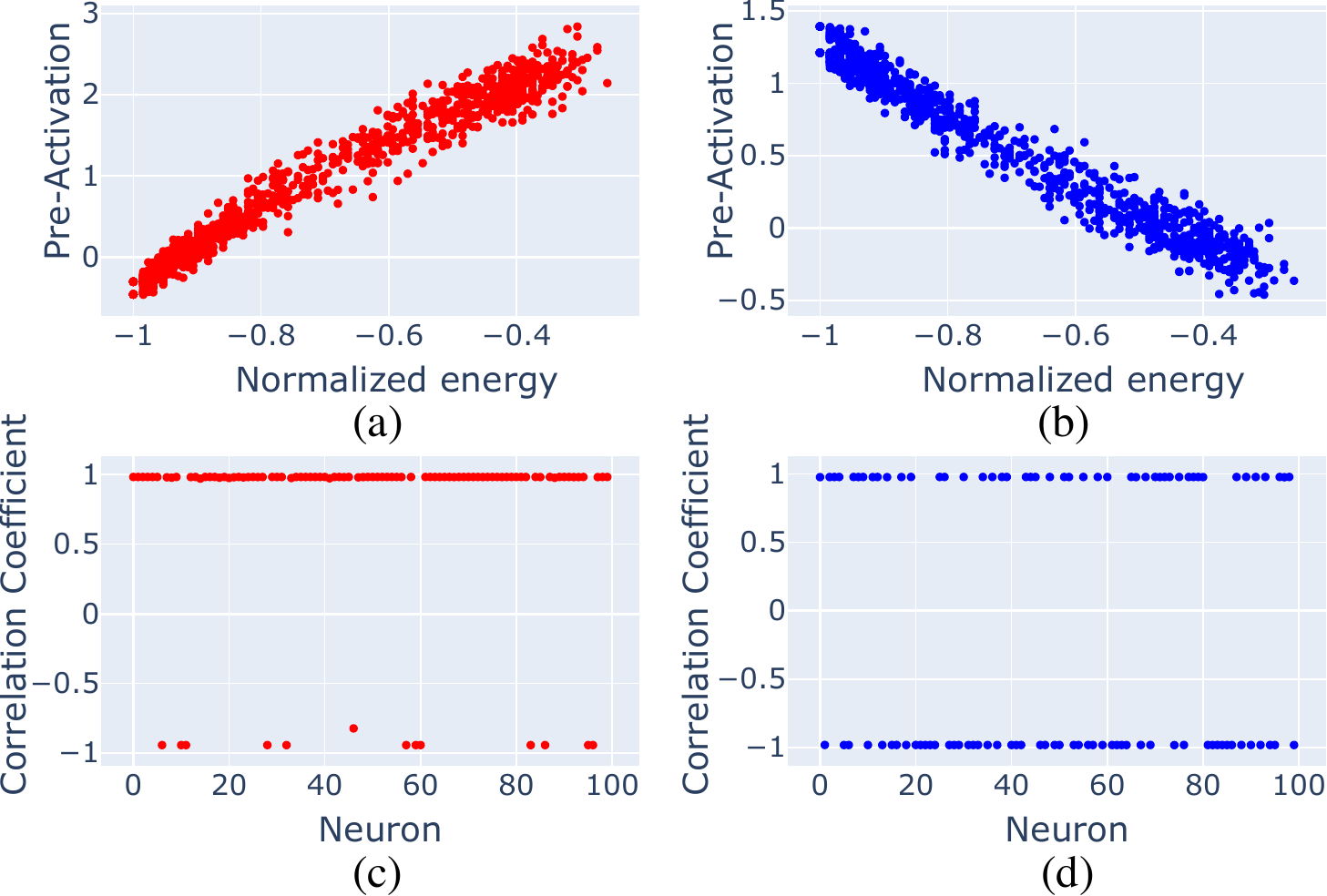}
    \caption{In the Ising task, the model learns the energy for both the grokking (a,c) and learning (b,d) regimes. In the top row, we show a scatter plot of the energy of input snapshots versus the resulting neuron pre-activations in the final layer. Data is shown for only most activated neuron. In the second row, we show that this high degree of correlation is typical of all neurons in the final layer.}
    \label{fig:single_neuron_ising}
\end{figure}

We now show that models reached via grokking and ``steady" learning learn the same dataset features. 
Previous studies have established that the key features learned by models trained on the Ising task are the energy (Eq.~\eqref{eq:Energy}) and the magnetization (Eq.~\eqref{eq:Magnetization})~\cite{hu_ising_int_2017, isingint_suchs_2018}.
In these earlier works, it was shown that models (trained via the conventional steady learning trajectory) learned a mix of both features.
To make a cleaner comparison between grokked and learned models, it is helpful to have a single dominant learned feature. We thus deliberately design our setup to favour learning the energy. To achieve this, we include training snapshots that are in the ordered phase but have anomalously low magnetization.
These snapshots are correctly classified by the energy but misclassified by the magnetization, and often appear as metastable states in local-update Monte Carlo simulations of the Ising model~\cite{Suchsland2018May}.
Moreover, we use a more powerful network, including a CNN component to detect the interfaces associated with calculating energy.\\

To understand which of the three features the model learns, we correlate the pre-activations for all neurons in the networks with the energy, the magnetization, and the absolute value of the magnetization. In Fig.~\ref{fig:single_neuron_ising}, we compare the result for models reached via grokking and steady learning trajectories. We find that, in both grokking and steady learning, \textit{every} neuron in the final layer is either perfectly correlated or anti-correlated with the energy of the input. In Appendix C, we show how this develops in each layer of the model. For every seed, in both learning and grokking, the energy has the highest correlation of the three measures with the majority of neuron activations in the second CNN and in the fully connected layer. We hence conclude that both grokking and steady learning models classify snapshots by learning the same feature - the energy.

\subsection{Learning modular addition: both models learn the Fourier coefficients}
\begin{figure}[t]
    \centering
\includegraphics[width=1\linewidth]{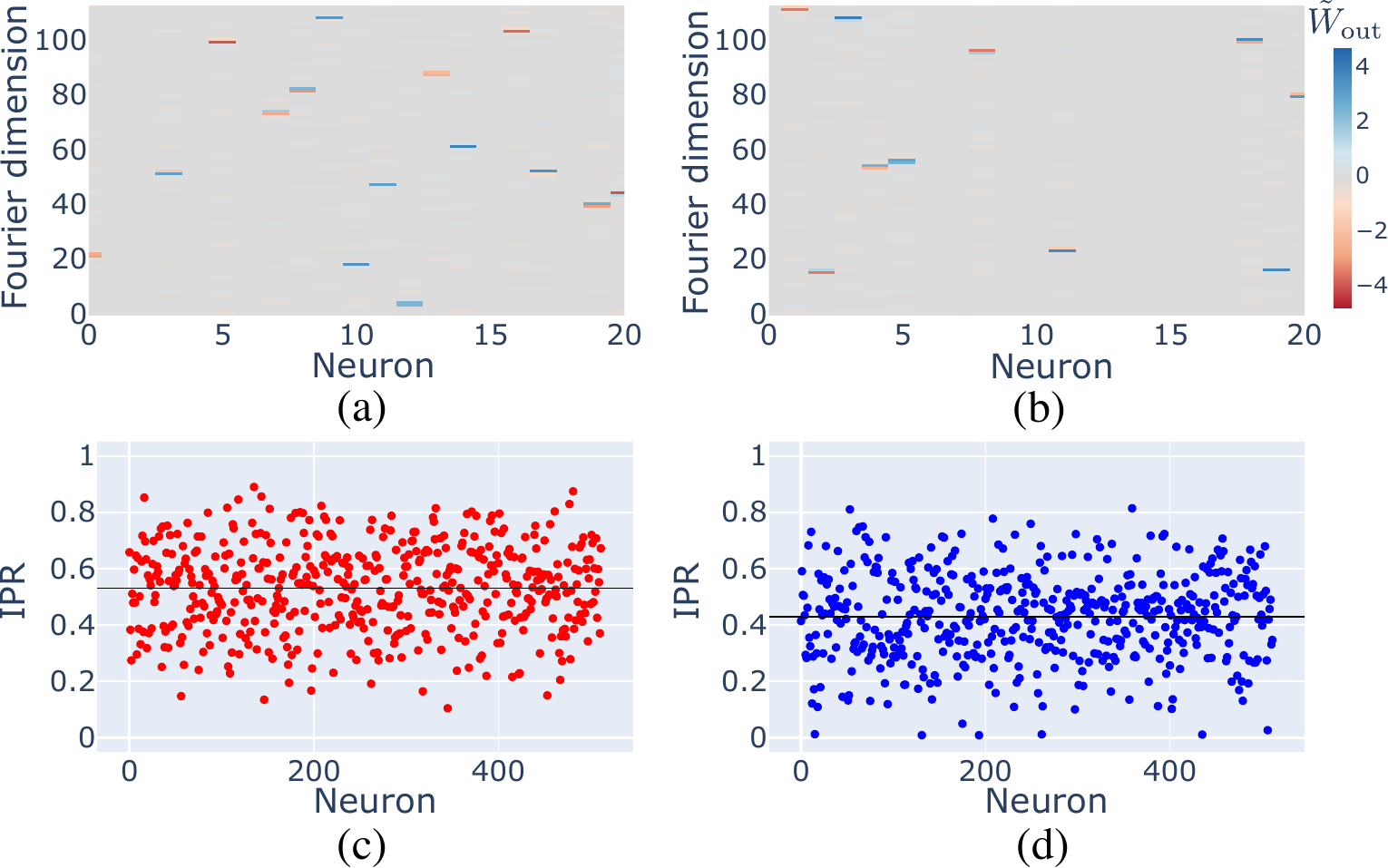}
    \caption{In the modular addition task, the model learns the Fourier modes of the input data. To see this, we Fourier transform the unembedding matrix and plot its coefficients, which correspond to different Fourier frequencies. 
    The heatmaps for (a) grokking and (b) learning suggest that each neuron predominantly activates on the sine and cosine of a single frequency. This is confirmed by calculating the IPR for all neurons, shown in (c) for grokking and (d) for learning.} 
    \label{fig:mod_fourier_intp}
\end{figure}
To better understand the task dependence of our comparison between grokking and learning, we also study the modular addition task.
A number of studies have convincingly shown that models trained on the modular addition task typically learn a Fourier representation of the problem \cite{power2022grokking,nanda2023progress,gromov2024_grokking}.
For a two layer MLP with square activation, \cite{gromov2024_grokking} showed that an ansatz of Fourier modes for the model weights solves the modular addition problem. Moreover, the inverse participation ratio (IPR) can be used as a measure of how well-localized the weights of the model are in the Fourier basis. This provides a quantitative comparison of the extent to which grokked and learned models learn the same features. \\
The distribution of neuron post-activations in the fully connected layer is shown in Fig.~\ref{fig:mod_fourier_intp}(a-b).
We see that each neuron learns the sine and cosine components of a single frequency. To quantify the localization to a given frequency in a given neuron $k$, we first Fourier transform the input (output) weight matrices in the embedding (unembedding) dimension, and then use Gromov's Inverse Participation Ratio measure~\cite{gromov2024_grokking} to quantify the localization to a given frequency:
\begin{equation}
    IPR\,(k)=\sum_{\nu=1}^{D}\frac{|\tilde W_{\nu k} |^{4}}{(\sum_{\nu}\tilde W_{\nu k}^2)^2},
\end{equation}
where $\nu$ is the index of the embedding (unembedding) dimension of the input (output) matrices to the hidden layer, and $D = 2P$ ($D=P)$ is its size. The results are shown in Fig.~\ref{fig:mod_fourier_intp}(c-d). First, we notice that the average IPR is around $1/2$, consistent with each neuron predominantly processing the sine and cosine components of a given frequency. Second, we see comparable mean IPRs between the grokking and the steady learning cases; 0.43 and 0.53, respectively. We therefore conclude that, as in the Ising task, grokked and steadily learned models learn the same dataset features.
\section{Learning models can be more compressible than grokking models}
\begin{figure}[t]
    \centering
    \includegraphics[width=\linewidth]{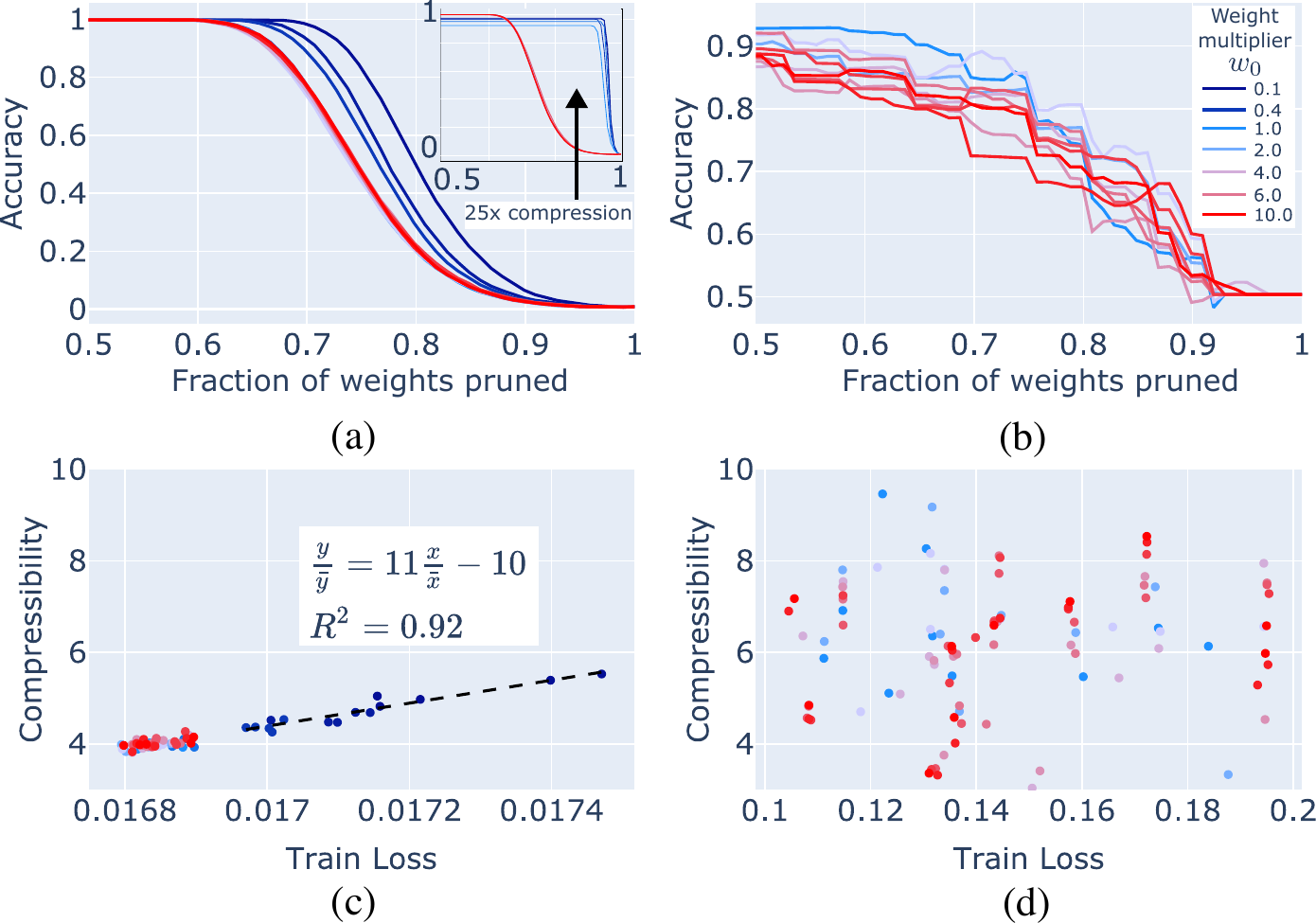}
    \caption{Pruning curves for (a) Ising and (b) modular addition. In the modular addition task, there is a linear trade-off between the end-loss and the compressibility $c$, shown in (c), while no such relationship for the Ising task, shown in (d). Data in the top row is averaged over seeds, while the bottom row shows all seeds.}
    \label{fig:pruning_both}
\end{figure}

Having established that both grokking and steady learning lead to similar features, we now show that there can nevertheless be significant differences in the efficiency with which the features are encoded. To see this, we use magnitude pruning on the end model in both tasks. Our key result in this section is the emergence of a linear trade-off between end model \emph{training} loss and compressibility in modular addition (Fig.~\ref{fig:pruning_both}). We call this parameter range the ``compressive regime". The compressive regime is a particularly striking example of a general feature: despite learning the same dataset features, grokked and steadily learned models can exhibit large differences in compressibility.  

We start by introducing our pruning scheme. In the main text, we use a global magnitude pruning scheme. In the appendix, we show the results of pruning each layer in the network individually. To account for differences in weight scales and layer architecture, we slightly modify naive global pruning. First, we rank all weights within each layer separately by their absolute values. Given a pruning fraction $p$, we set the smallest $p$ fraction of the weights in each layer to zero, and then evaluate the model accuracy on the test set.
To quantify model compressibility, we first integrate the accuracy $a(p)$ at each pruning fraction $p$ with respect to $p$ - i.e. we find the area bounded by the pruning curve and the x-axis in Fig.~\ref{fig:pruning_both}(a-b). We then define the compressibility $c$ as:
\begin{equation}
    a\equiv \int a(p)dp, \text{  }   c\equiv\frac{1}{1-a}.
\end{equation}
This is analogous to finding the size of the pruned model at a fixed accuracy, but averaged over the entire range of pruning.

We show the results of pruning across weight multipliers for modular addition in Fig.~\ref{fig:pruning_both}(a). In the steady learning regime $w_0\leq 0.4$, model compressibility improves systematically with reducing weight multiplier. In the Fig.~\ref{fig:pruning_both}(c), we see that once the model has transitioned into this parameter regime there emerges a striking linear trade-off between encoding efficiency and end model training loss. In Appendix D, we provide data for the relationship between the compressibility and a range of other network measures. In particular, we emphasize that the there is no correlation between compressibility and the \textit{test} loss. Surprisingly, we also find no relationship between the compressibility and the model degeneracy, as measured by the local learning coefficient~\cite{hoogland2024developmentallandscapeincontextlearning,bayesianphasellc2023,lau2024locallearningcoefficientsingularityaware}.

In the grokking regime of modular addition, we see that there is no significant dependence of model compressibility on $w_0$ - the red curves are on top of each other. Hence, although models trained in both grokking and steady learning learn the Fourier components of the input data, models trained in the compressive regime of steady learning are significantly more compressible. We emphasize that the existence of a compressive regime in steady learning is not a generic feature, but is parameter-dependent. In Appendix F, we show that this regime can be eliminated by increasing the batch size from 64 as used in the main text to 200. The compressive regime is thus a novel parameter regime of steady learning, rather than a generic feature. 

In the inset of Fig.~\ref{fig:pruning_both}(a), we show that we can achieve extremely high compressions for a higher weight decay ($3\times 10^{-4}$ vs $3\times 10^{-5}$). Pairing this with a low weight multiplier, we are able to \textit{cut 95\% of the weights} in the model for a 5\% drop in accuracy, a compressibility 25x that of the original model, and 5x the highest compression achieved in at the lower weight multiplier used in the rest of the main text. In Appendix E, we explore this regime. We show that in this extremely compressed regime, the mean IPR $\tilde{IPR}$ falls by close to an order of magnitude, indicating that the model's Fourier representation is breaking down. We also note that the model is itself close to breakdown - models with a 30\% larger weight decay no longer learn at all.

Finally, we show the model compressibility 
for the Ising task in Fig.~\ref{fig:pruning_both}(b,d). Here we see no clear relationship between compressibility and weight multiplier, and therefore no compressive phase. Furthermore, there is no meaningful trade-off between model compressibility and the end loss. This reinforces the view that the compressive phase found in the modular addition task is a distinct regime of training dynamics.

\begin{figure}[t]
    \centering
    \includegraphics[width=0.9\linewidth]{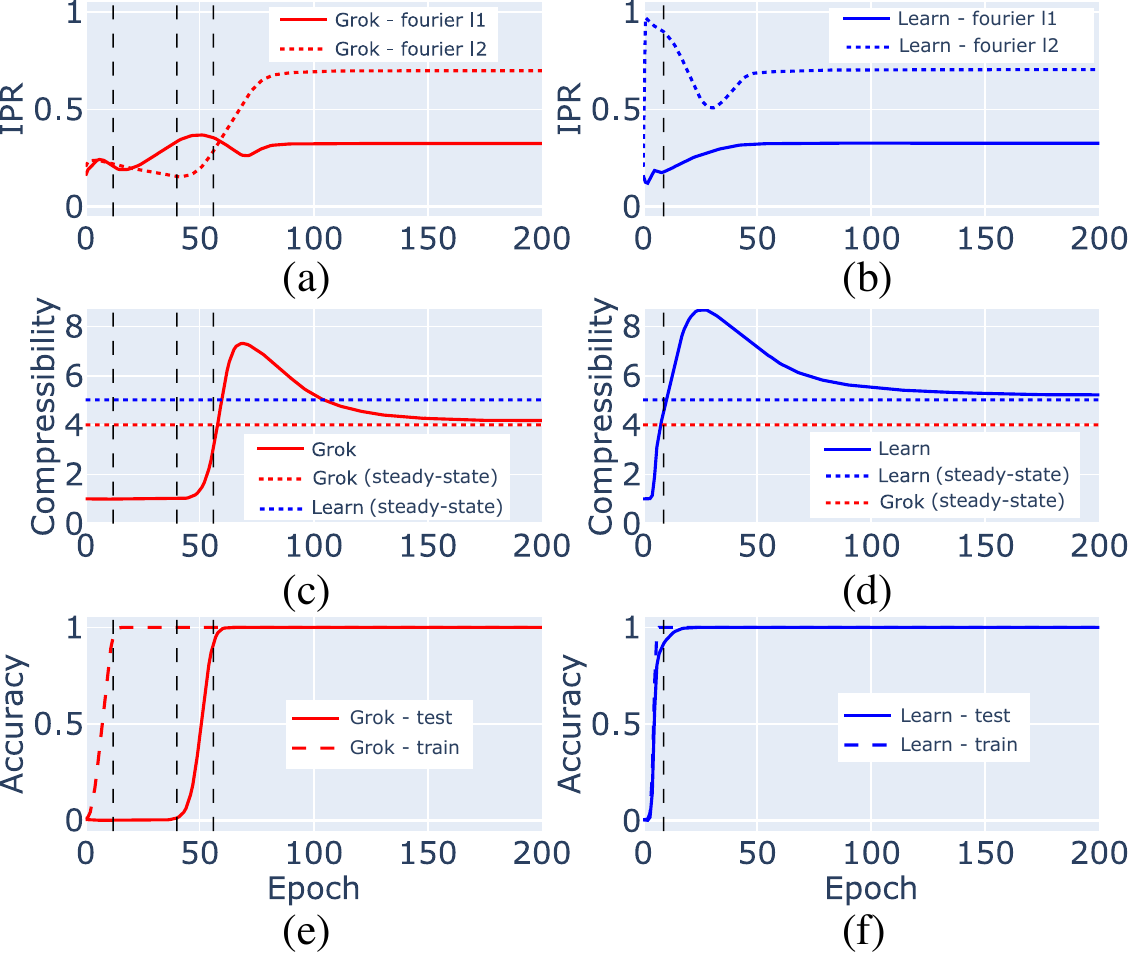}
    \caption{Dynamical measures comparing learning and grokking for modular addition. From left to right, the vertical dashed lines represent the training time, the end of the grokking plateau, and the test time. The model appears to be developing before and during the grokking transition, We also see a transient compressibility spike in the pruning dynamics.}
    \label{fig:mod_iprs_time}
\end{figure}

\section{Model development in grokking and learning}
Having established measures for both the dataset features learned and model compressibility, we now ask how they develop over the course of training.
We present two main results. 
First, that the development of models in the pre-grokking plateau is different in the Ising and modular addition tasks.
In the modular addition task, consistent with previous literature on ``progress measures" \cite{nanda2023progress, gromov2024_grokking}, we find that the model continues to develop during the plateau, despite the fact that the accuracy does not improve. For the Ising task, however, we find that the model does not develop in the grokking plateau. 
This suggests that model dynamics in grokking is task dependent.
We support this through the introduction of novel FIM-inspired measures for trajectory analysis in model space, which provide another perspective for identifying model development.

\subsection{Feature Development}
To measure the development of model features, we first define layer-wide summary averages for our interpretability measures. For the Ising model, we define a pre-activation weighted correlation coefficient $\tilde{r}$ for a layer $L$ as the sum of the (absolute value of) the correlation coefficients for each neuron $k$ in that layer, weighted by that neuron's share of the absolute pre-activation, $w_k$
\begin{equation}
    w_{k}\equiv \frac{\sum_{i\in \mathcal{D}}|z^i_{k}|}{\sum_{i\in\mathcal{D}}\sum_{k\in L}| z^i_k|},\,\,\text{   } 
    \tilde{r}\equiv\frac{1}{|L|}\sum_{k\in L}w_{k}|r_k|,
\end{equation}
where $z^{i}_{k,L}$ is the pre-activation of the $k$-th neuron in layer $L$ on the input vector $i$, $|\mathcal{D}|$ is the number of elements in the test dataset and $|L|$ is the number of neurons in the layer. For modular addition, we simply take the mean of the IPR across neurons $k$ in the hidden layer to define $\tilde{IPR}$ as:
\begin{equation}
    \tilde{IPR}\equiv\frac{1}{|L|}\sum_{k\in L}IPR(k),
\end{equation}

\begin{figure}[t]
    \centering
    \includegraphics[width=1\linewidth]{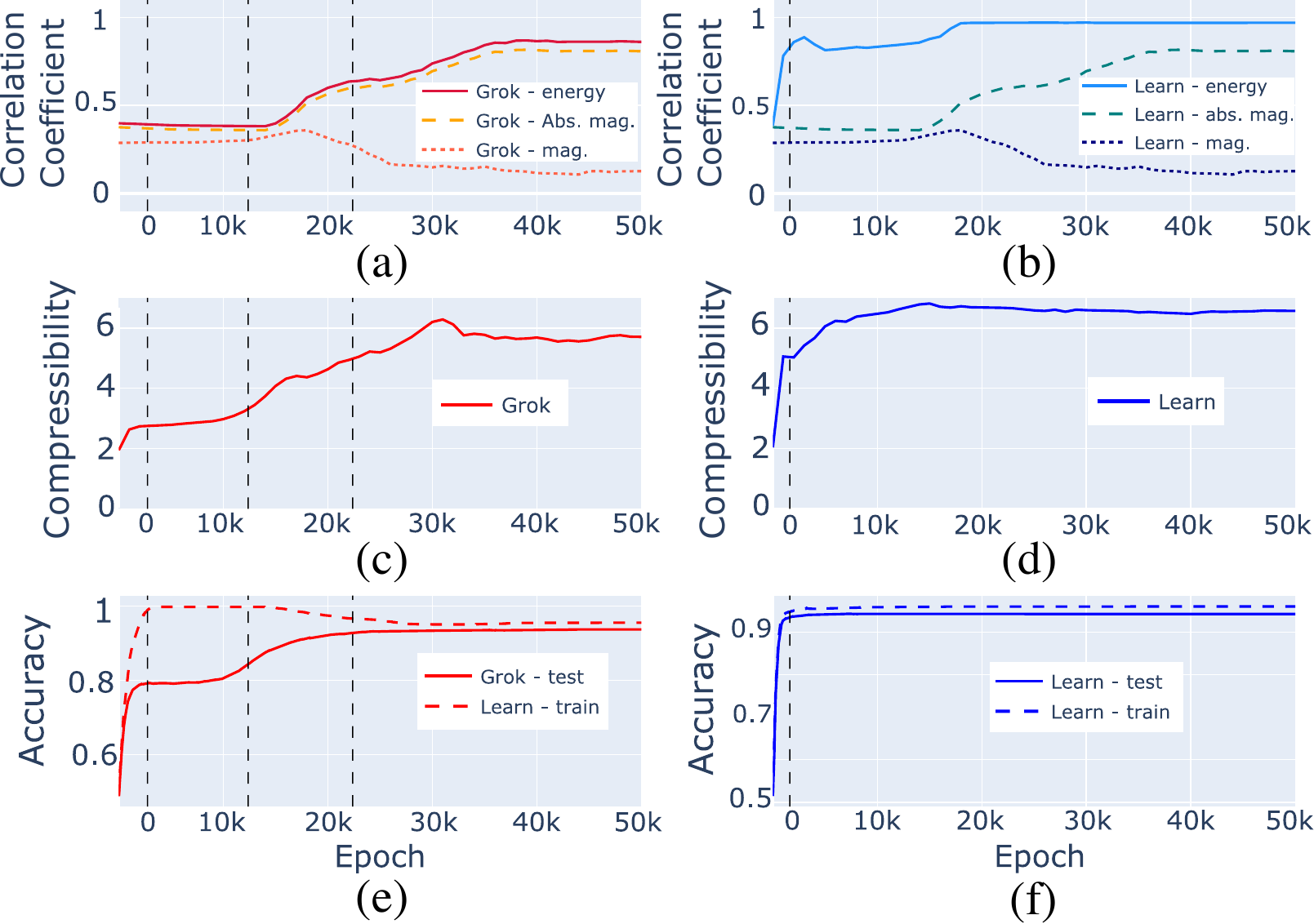}
    \caption{Dynamical measures comparing learning and grokking for the Ising task. Unlike modular addition, the model shows no sign of developing through the grokking plateau.}
    \label{fig:ising_dyn_corr}
\end{figure}

We summarize the behaviour of these measures during training for the modular addition task in Fig.~\ref{fig:mod_iprs_time} and for the Ising task in Fig.~\ref{fig:ising_dyn_corr}, and additional data is provided in Appendix G.
The two tasks have markedly different behaviour in the grokking plateau.
In the modular addition case, the model improves its localization in the Fourier basis throughout the plateau despite no improvement in model accuracy or loss. 
This is consistent with the ``Circuit formation" picture described for a transformer in Nanda et al.\cite{nanda2023progress} and for a two layer fully-connected network by Gromov \cite{gromov2024_grokking}. 
The model trained on the Ising task, however, does not improve its representation of the energy in the plateau - all three metrics are roughly constant across all layers in the plateau. Whether or not the model is developing features in the grokking plateau is hence task dependent. Interestingly, in both grokking and steady learning, the model  continues to improve its representation of the dominant feature well after the accuracy and loss are saturated.\\

\subsection{The dynamics of compressibility}
We also consider the dynamics of model compressibility. The results for the Ising task are given in the middle row of Fig.~\ref{fig:ising_dyn_corr}(c-d). In the grokking case, we see that the compressibility peaks a short time after the accuracy saturates and then declines slightly. The results for the modular addition task are given in Fig.~\ref{fig:mod_iprs_time}(c-d). Here we see a transient 50\% larger than the steady state compressibility for the steady learning trajectory, and 67\% larger in the grokking trajectory. Comparing to the model formation measures in Fig.~\ref{fig:mod_iprs_time}(a-b), we note that after the compressibility peak, the compressibility \textit{decreases} as the neuron IPR $\tilde r$ increases. We also note that we observed the transient peak in the compressibility in all our modular addition runs - even at larger batch sizes for which there is no compressive phase. This suggests that the transient behaviour in the compressibility is a distinct phenomenon from terminal model compression.

\subsection{Information-Geometric Trajectory Analysis}\label{sec:fim}
Once a model architecture has been set, the space of parameters defines the space of functions the model can represent, known as the model space.
This model space may have a non-trivial geometry, which can be measured by the Fisher Information Metric (FIM). We give a fuller account of this connection in Appendix F. The FIM can then be used to measure how the information geometry of a model changes in training.\\

\begin{figure}[t!]
    \centering
    \includegraphics[width=\linewidth]{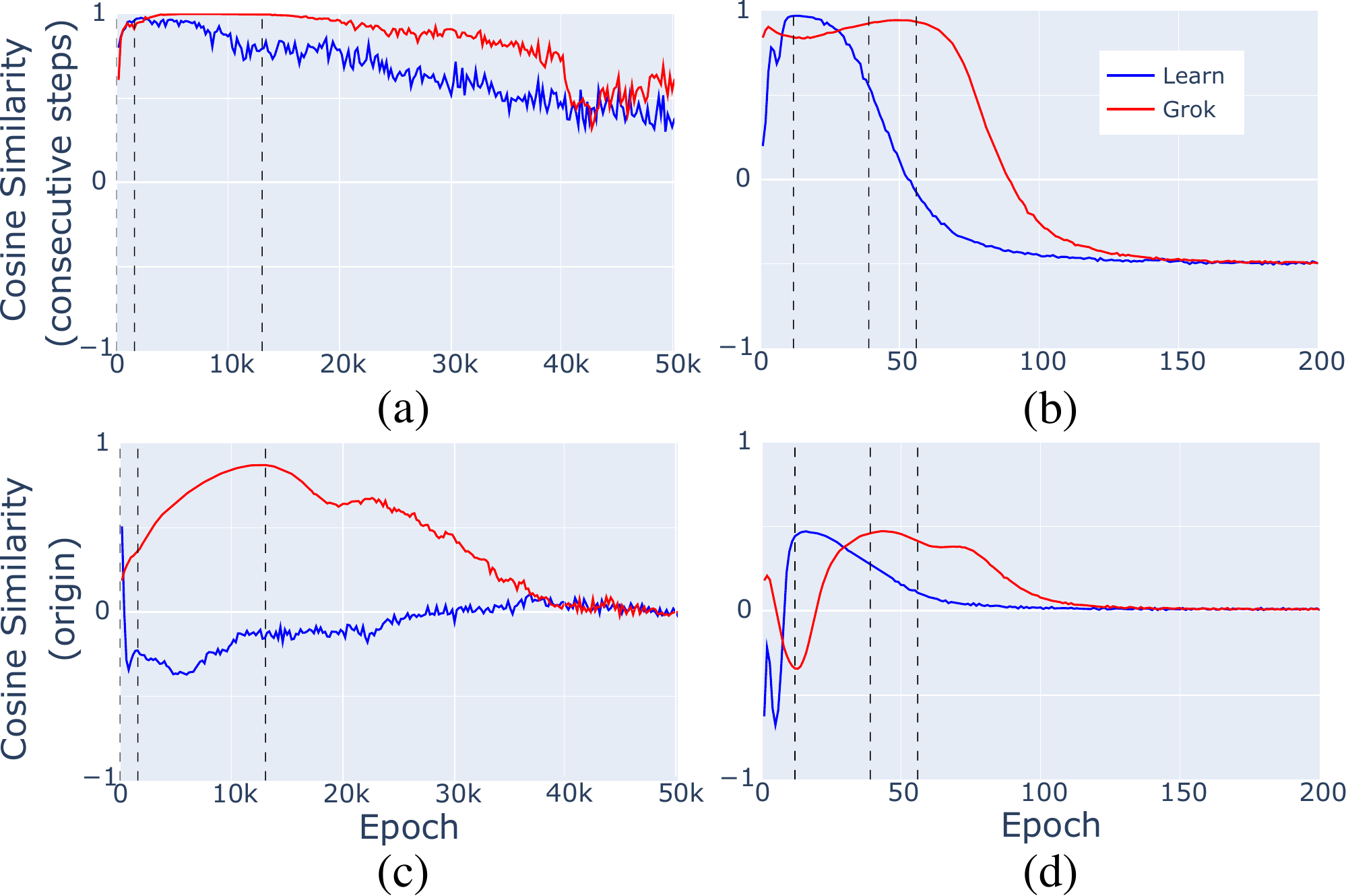}
    \caption{FIM-cosine similarities, $S_{C-FIM}(s,s')$, of the model trajectories through model space during training (averaged over 10 runs). The similarity between consecutive steps, $(s^e,s^{e+1})$, is shown in (a) for the Ising task and in (b) for modular addition. The similarity between each step and a trajectory pointing to the origin of model, $(s^e,s^{OT})$, is shown in (c) for the Ising task and in (d) for modular addition.}
    \label{fig:dynamics_cosine_full}
\end{figure}

We use this connection to introduce measures of model ``speed" and ``direction" in model space through training.
The measures are the information-geometric generalisation of step magnitude and cosine similarity between consecutive update steps along the trajectory.
Given a model space position $\theta^e$ at epoch $e$, the model space step is defined as $s^e := \theta^{e+f}-\theta^e$, where $f$ is the frequency between samples which is $200$ $(1)$ for the Ising (modular addition) task. 
From this the step magnitude and cosine similarity between two steps $s$ and $s'$ are given by
\begin{align}
    |s|_{FIM} & := \sqrt{s_i \cdot g^{FIM}_{ij} \cdot s_j} &\quad & \in [0,\infty)\;,\\ 
    S_{C-FIM}(s,s') & := \frac{s_i \cdot g^{FIM}_{ij} \cdot s'_j}{|s|_{FIM}|s'|_{FIM}} &\quad  & \in [-1,1]\;,
\end{align}
where repeated indices are summed. Focus is placed on the cosine similarity, where values close to one indicate steps that are approximately parallel, while values close to zero indicate orthogonal steps.
We track the step cosine similarities, averaged over the 10 seed runs, for both the Ising and modular addition tasks, and for both the grokking and steady learning. In particular, we compute the cosine similarity of consecutive steps, $s^e$ and $s^{e+1}$, and show the results in Fig.~\ref{fig:dynamics_cosine_full}(a-b).

In both tasks and both regimes, the cosine similarity initially increases, reaches a peak around the learning time, and then decreases to near zero. These results reflect the training dynamics: once the model finds a good trajectory, it continues updating in roughly the same direction until it reaches its minimum loss. After learning, the model updates are random fluctuations around the loss minimum, and consecutive steps become uncorrelated.
The grokking regime shows peaks later in the training, and for the Ising task the peak is noticeably higher than in the steady learning regime. In fact, the peak cosine similarity is nearly one, indicating that the model evolves along a straight line in the high-dimensional parameter space. 

Previous work suggests that the dynamics before grokking is dominated by a decay of extraneous parameters~\cite{nanda2023progress}, a hypothesis corroborated by the exponential fitting of the total weight norm in~\cite{tegmark2023omnigrok}.
During uniform weight decay, the trajectory of a model points towards the origin of model space. This is a particularly special path, and may explain the unusual directness of the grokking trajectory for the Ising task.

To understand whether this is actually the case, we compute the cosine similarity between each step and the negative of the model position at the start of that step (which points to the origin).
The values of this cosine similarity have a particularly natural interpretation: when they are positive the FIM-weighted average weight is increasing, and for negative values it is decreasing. In the limits of $S_{C-FIM} \approx \pm 1$, uniform weight decay/growth is the dominant feature.
These results, for both the Ising and modular addition tasks, and for both the grokking and steading learning regimes, are shown in Fig.~\ref{fig:dynamics_cosine_full}(c-d). At late times, all cases exhibit a cosine similarity of zero, indicating neither weight decay or growth once the terminal model is reached.

For the Ising task, the grokking regime initially follows a path close to the origin trajectory with the cosine similarity approaching one. The opposite occurs in the learning case: the negative similarity score corresponds to network parameters marginally increasing.
These results provide ulterior evidence that, in the Ising task, the dominant effect before the grokking transition is weight decay. Furthermore, such a trajectory also explains why feature development does not occur until the grokking transition. In the loss plateau prior to the transition, the dynamics are almost entirely weight-decay driven, and there appears to be no ``hidden'' model formation.

For the modular addition task, the cosine similarity to the origin trajectory is somewhat different. At the start of training, both grokking and steady learning regimes exhibit an early regime where the model seems to be moving slightly away from the origin, which later gives way to a regime of slightly origin-oriented updates approaching the learning transition. However, the cosine similarity is significantly lower than in the Ising task, indicating a trajectory that is less dominated by weight decay. 
This corroborates the fact that for modular addition, the grokking trajectory also develops features before the grokking transition, and thus the dynamics cannot be simply a uniform decay of weights.

Overall, these novel information-geometric inspired measures provide a new means of analysing machine learning training dynamics by considering the geometry of the model space implicitly. By tracking the direction of the trajectory in model space, we have provided evidence that grokking dynamics is dominated by weight decay for the Ising task, but not for modular addition. Besides capturing the directions of model updates, the FIM can also be used to track their magnitude. An analysis of update step magnitude is left to the appendix, but we note that it displays unique gradient changes at the start and end points of the grokking plateau, and hence suggests a novel method of identifying grokking.

\section{Conclusion}
To compare the features learned in grokking and learning as closely as possible, we studied toy models with clearly measurable features.
This immediately raises the question of which of our findings generalise to larger models. In general, our work suggests three main questions for further work.

First, \textit{are there properties of grokking that are practically useful}? There are two ways in which this could occur: a grokked model may learn different features or be more compressible than one trained through ``steady" learning. In our case, the features learned were the same, and the steady learning regime had higher compression. If both of these conclusions are generally true, this suggests that grokking should be eliminated whenever possible. In our tasks, we accomplished this by tuning the weight scale at initialization. However, it is unknown if there are tasks which can \textit{only} be grokked, or, vice versa, can \textit{only} be steadily learned.

Second, \textit{what is the nature of the compressive regime, and can it be used to compress practical models}? Fortunately, we discovered a compressive regime in ordinary learning and not in grokking. The very large compression achieved raises the interesting possibility of using such a training scheme to obtain more compressible models in general.
In addition, the compressive regime allows us to tune the size of the effective model by changing the initial weight multiplier. 
Since models are known to compress features by superposition \cite{elhage2022superposition}, it would be particularly interesting to investigate in further work whether models in this regime display significantly higher superposition than models trained outside of it.

Finally, \textit{can information geometry be used to help scale interpretability}? Our interpretability measures allowed us to follow model development through training. Although this provided progress measures in which the model develops smoothly in the modular addition task, it does not do so for the Ising task, where generalisation still emerges suddenly. Crucially, these measures had to be constructed by hand for each task. By contrast, ouinformation geometric measures were general, and identified signatures of the absence of model development in the Ising task as well as the non-trivial pre-grokking dynamics of the modular addition task. It would be interesting to further explore how information geometric measures can be used for model interpretability.

\section*{Acknowledgments}
We gratefully acknowledge extremely helpful discussions with Marc S. Klinger. Marc brought together the initial collaboration, and contributed important insights on information geometry. 
The work of B.B. was supported by the US National Science Foundation under Grant No. DMR-1945058.
A.G.S, D.S.B, \& E.H acknowledge support from Pierre Andurand over the course of this research. G.D.T. acknowledges the support from the EPiQS Program of the Gordon and Betty Moore Foundation.
This work made use of the Illinois Campus Cluster, a computing resource that is operated by the Illinois Campus Cluster Program (ICCP) in conjunction with the National Center for Supercomputing Applications (NCSA) and which is supported by funds from the University of Illinois at Urbana-Champaign. This research was supported in part by the Illinois Computes project which is supported by the University of Illinois Urbana-Champaign and the University of Illinois System. 
This research also utilised Queen Mary's Apocrita HPC facility \cite{apocrita}, supported by QMUL Research-IT.

\section*{Data Availability}
The code and datasets used in this work is available here: \url{https://github.com/xand-stapleton/grokking_vs_learning}. 


\nocite{*}
\bibliographystyle{plainnat} 
\bibliography{references} 


\appendix

\onecolumn
\section{Model parameters and seed distribution}
In this work, we control grokking solely by the modification of the weight multiplier. For convenience, \cref{tab:model-params} provides all other model parameters for the data generated in the main text. 
\begin{table}[h]
\caption{Model Architecture and Training Parameters}
\label{tab:model-params}
\centering
\begin{tabular}{lcc}
\toprule
\textbf{Parameter} & \textbf{Ising CNN} &\textbf{Modular addition MLP} \\
\midrule
Weight decay & 0.1 & $3\times 10^{-5}$\\
Learning rate & $10^{-4}$& $5\times 10^{-3}$\\
Epochs & 100,000 & 1000\\
Loss & Cross Entropy & Cross Entropy \\
Optimizer & Adam & Adam\\
Activation & ReLU & ReLU\\
Modulation (P) & -- & 113\\
Train fraction & -- & 70\% \\
Train dataset size & 300 & 8938\\
Batch Size & 300 & 64 \\
Test dataset size & 1000 & 3831\\
Fully connected layer size & 100 & 512\\
Convolutional channels & 2,4 & --\\
Convolutional stride & 2,2 & --\\
Kernel size & 2,2 & --\\
\bottomrule
\end{tabular}
\end{table}

\begin{figure}[h!]
    \includegraphics[width=1\linewidth]{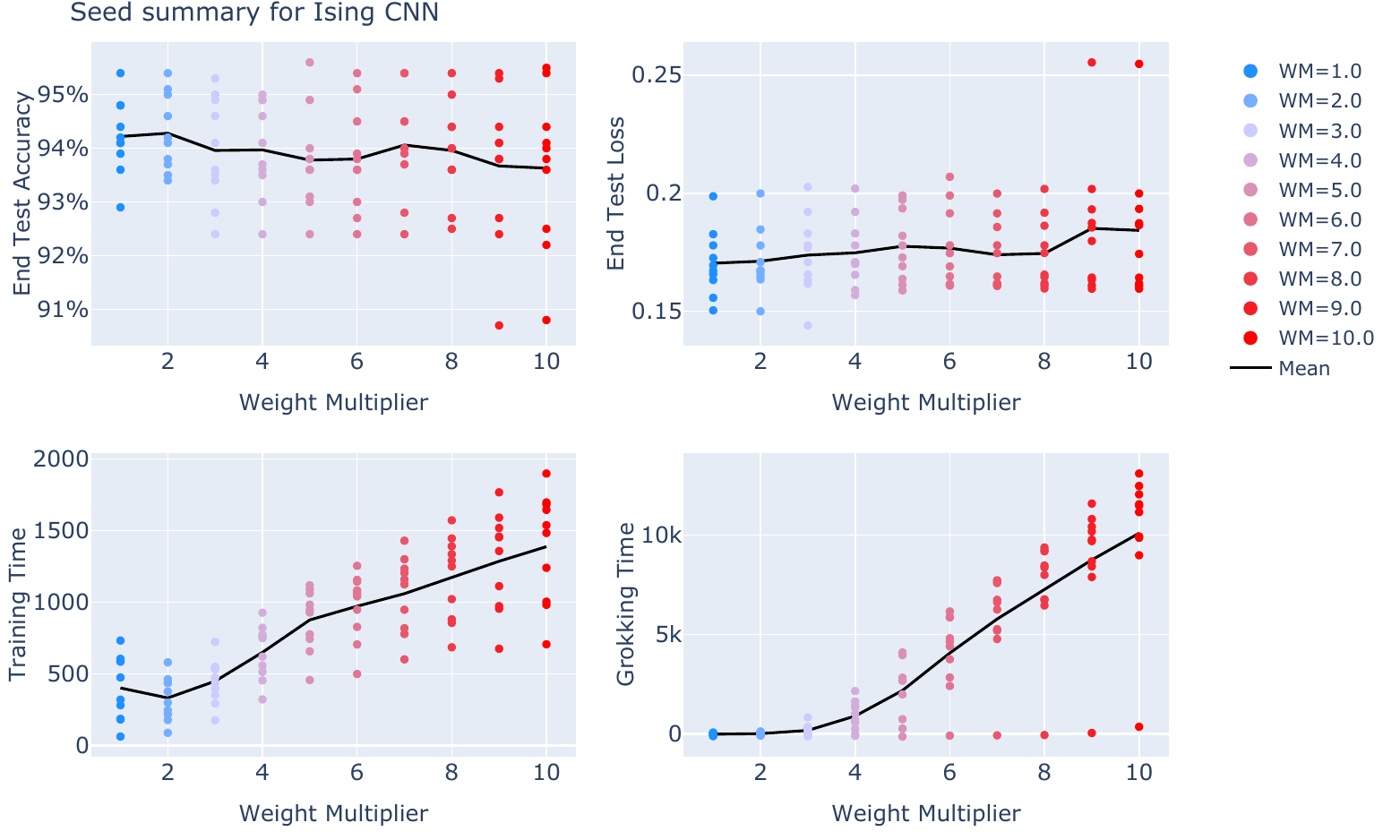}
    \caption{A summary of the outcomes of the runs for the 10 different seeds used for each weight multiplier. We see that all bar one of the runs groks for the higher weight multiplier regime of the plot, and that for the majority of seeds the final test accuracies and losses for grokking and learning are within a few percent of each other.}
    \label{fig:ising_seed_summary}
\end{figure}

 We also provide summary data for the individual runs. In \cref{fig:ising_seed_summary} we provide the distribution over individual seeds for Ising runs. We see that final model accuracy and loss are consistent as we tune from grokking into learning. We also see a slight increase in seed variability as we tune deeper into the grokking regime, with one seed persistently failing to grok. 

\begin{figure}[h!]
    \includegraphics[width=1\linewidth]{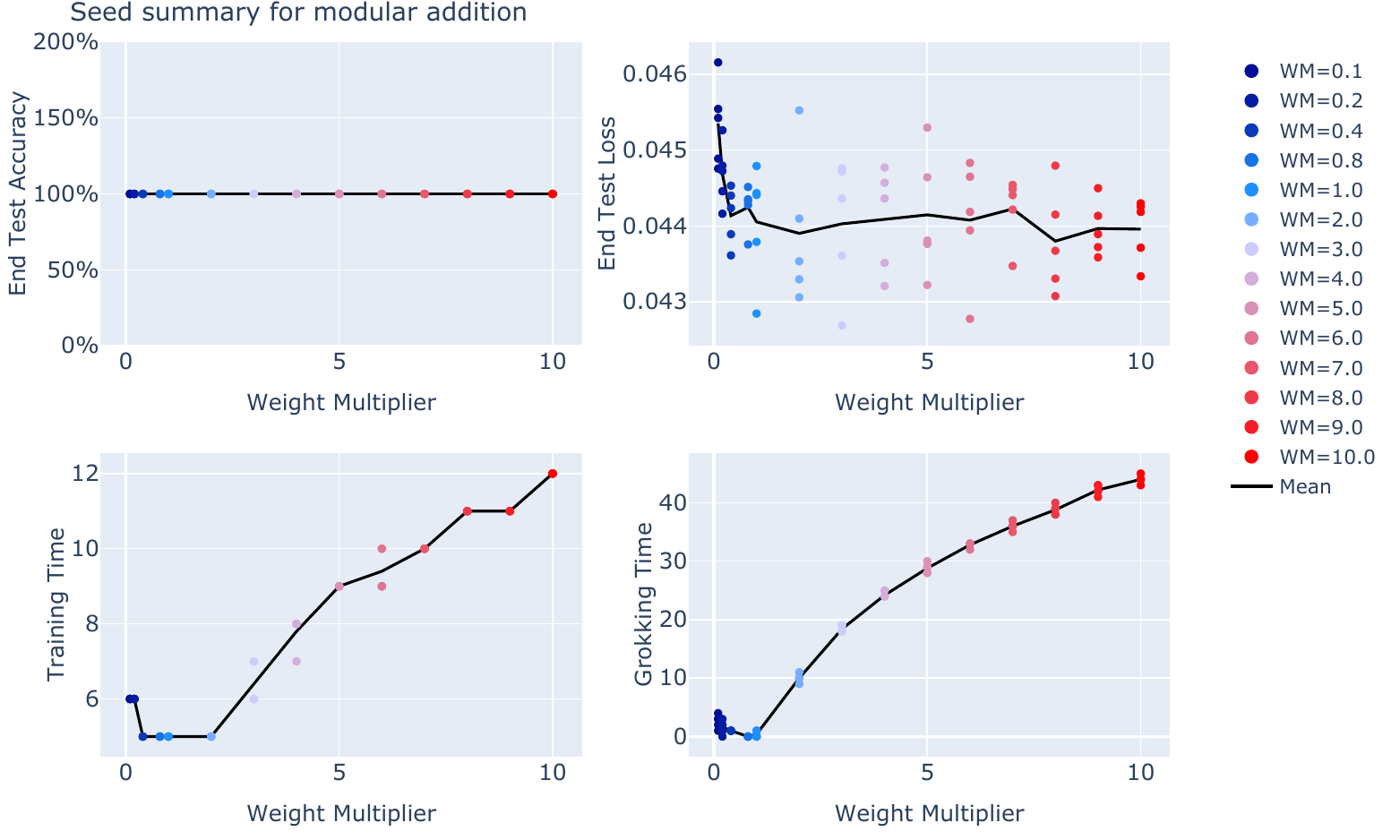}
    \caption{A summary of the outcomes of the runs for the 5 different seeds used for each weight multiplier in modular addition.}
    \label{fig:mod_seed_summary}
\end{figure}

 For modular addition, we provide the summary data over seeds in \cref{fig:mod_seed_summary}. We see that the modular addition task is significantly less seed dependent than the Ising task. All seeds reach 100\% accuracy, and the end losses are within 5\% of each other, whereas for the Ising tasks, end accuracies are within 5\% and  end losses can vary by more than 50\%. We also note that, interestingly, both modular addition addition and the Ising task have an increase in their training time as they transition from the grokking to the learning regime.

\section{The Ising classification problem}
This section gives a more formal definition of the Ising model used in the main text.
Generically, the Ising model is defined on a graph $G$ (with a vertex set $V$ and an edge set $E$), where each node is associated with a discrete variable $\sigma_x \in \{-1, 1\}$, called a spin. The space of spin configurations (snapshots) is given by all possible combinations $\Omega = \{-1,1\}^V$. The Ising model can be defined as a probability distribution over this space of snapshots $\Omega$. More precisely, a snapshot $\{\sigma_x\}_{x \in V}$ iw drawn from the Boltzmann distribution:
\begin{equation}
\label{eq:Boltzman_eq}
p(\{\sigma_x\}) = \frac{e^{- E(\{\sigma_i\})/T}}{Z(T)},
\end{equation}
where $T$ is the temperature of the system, $Z(T)$ is a normalization factor (the partition function), and $E(\{\sigma_x\})$ is the energy (cost) function:
\begin{equation}
\label{eq:Energy_Ising}
    E(\{\sigma_x\})=-J\sum_{(i,j)\in E}\sigma_i\sigma_j.
\end{equation}
Here the sum is over edges of the graph, and $J>0$ is the interaction strength between spins, which in the main text is set to one for simplicity.
The phase transition is determined by the competition between the interaction $J$, which favours ordering by aligning the spins to minimize the energy in Eq.~\ref{eq:Energy_Ising}, and the temperature $T$, which enlarges the width of the probability distribution in Eq.~\ref{eq:Boltzman_eq}, and thus favours disorder by exploring spin configurations with random spin alignments.

In the main text, we consider the case in which the graph $G$ is the square two-dimensional ($2D$) lattice of linear size $L=16$ ($|V| = L^2$). 
In the limit of large linear size $(L\rightarrow \infty)$, the model presents a phase transition at $T_c= \frac{2J}{\log{(1+\sqrt{2})}}$, separating an ordered phase at $T<T_c$ and a disordered phase at $T>T_c$. 

Formally, these two phases are uniquely characterized by the averaged magnetization density (the local order parameter):
\begin{equation} \langle M \rangle (T) := \lim_{h\rightarrow 0}\lim_{L\rightarrow \infty} \langle \frac{1}{L^2}\sum_{i\in V}\sigma_i \rangle_h , \end{equation}
where $\langle \cdot \rangle_h$ indicates the average over the Boltzmann probability distribution with an external field, $p_h({\sigma_x}) = \frac{e^{- E_h({\sigma_i})/T}}{Z(T,h)}$, with $E_h({\sigma_x}) = E({\sigma_x}) - h\sum_{i\in V} \sigma_i$, and $Z(T,h)$ a normalization factor. The term $- h\sum_{i\in V} \sigma_i$ adds an energy penalty for having $\sigma_x = -1$, breaking the energy degeneracy $E({\sigma_x}) = E({-\sigma_x})$.
In the ordered phase, $\langle M \rangle$ is positive, while in the disordered phase, it is zero in the limit $L \rightarrow \infty$.
In particular, for the two-dimensional case:
\begin{equation} \langle M \rangle (T) = \begin{cases} \left ( 1- \frac{1}{\sinh^4{2J/T}} \right )^{1/8}, & T<T_c \\ \quad 0, & T>T_c. \end{cases} \end{equation}

From a concrete point of view, spin configurations drawn from Eq.~\eqref{eq:Boltzman_eq} for a finite $L$, which can be seen simply as binary images, look very different for $T$ smaller or larger than $T_c$. For $T=0$, the Boltzmann probability distribution is concentrated with equal weight in only two configurations: all spins $+1$ or all spins $-1$, and the absolute value of the magnetization, $|\sum_{x\in V} \sigma_x|$, equals $|V|$. As the temperature increases within the ordered phase, thermal fluctuations induce small clusters of spins with a sign opposing the bulk magnetization. Nevertheless, the magnetization remains extensive in in absolute value. In the opposite limit of $T\rightarrow \infty$, deep in the disordered phase, $p(\{\sigma_x\}) = 1/2^{|V|}$, and all configurations are equally probable. Consequently, typical configurations consist of random spins, and the absolute value of the magnetization is small compared to $|V|$. These qualitatively different spin arrangements make detecting the phase of a finite-size snapshot similar to the problem of image recognition.

The snapshot configurations used to train our convolutional neural network are generated using local-update Monte Carlo simulations. Specifically, we use the Glauber dynamics algorithm~\cite{glauber1963}, which we describe below. 

To sample a spin configuration from the Boltzmann probability distribution at fixed $T$, one starts with a configuration at infinite temperature, $T\rightarrow \infty$. In other words, one samples a spin configuration $\{\sigma_x\}_{x\in V}$ from the uniform probability distribution, $p(\{\sigma_x\}) = 1/2^{|V|}$. Next, one updates individual spins to obtain a typical configuration drawn from Eq.~\eqref{eq:Boltzman_eq} and approach the target average energy cost function. Each update is called a Monte Carlo step.

For concreteness, we consider the case in which the graph $G$ is a two-dimensional grid of linear size $L$, and each node is specified by two-dimensional coordinates $1\le x,y \le L$. We take periodic boundary conditions, e.g., $\sigma_{L+1,y} = \sigma_{1,y}$. The procedure is as follows:
\begin{enumerate} 
\item We sample a spin configuration (snapshot), ${\sigma_{x,y}}$, from the uniform probability distribution and calculate its energy, $E_0({\sigma_{x,y}})$. 
\item We randomly choose a node $(x,y)$ and compute the energy, $E_t$, of a new configuration in which the spin at that node is flipped: $\sigma_{x,y} \rightarrow -\sigma_{x,y}$. 
\item We choose to flip the spin and update the spin configuration with probability $p_\text{flip} = 1/(1+e^{\Delta E/T})$, where $\Delta E = E_t-E_0$ is the energy difference between the two spin configurations. If the spin-flip is accepted, we set $E_0 = E_t$.
\item We repeat steps 2 and 3, updating the spin configurations, for a total of $N$ Monte Carlo steps. We take $N \gg L^2$ to ensure at least partial equilibration before saving the resulting snapshot.
\end{enumerate}

To generate the training data for our classification task, we repeat this procedure 1000 times for 50 temperatures in the interval $[T_c - 1, T_c + 1]$ while fixing $J=1$. We then index the (finite-size) snapshots by their temperature, and their ultimate phase label (ordered or disordered) is based on the phase of an infinite system with that temperature. This procedure is guaranteed to generate snapshots of the Ising model in equilibrium if it is run for infinite time. In equilibrium, snapshots are completely characterized by their magnetization, and the classification task is almost trivial. 

\begin{figure}[h!]
    \centering
    \includegraphics[width=0.9\linewidth]{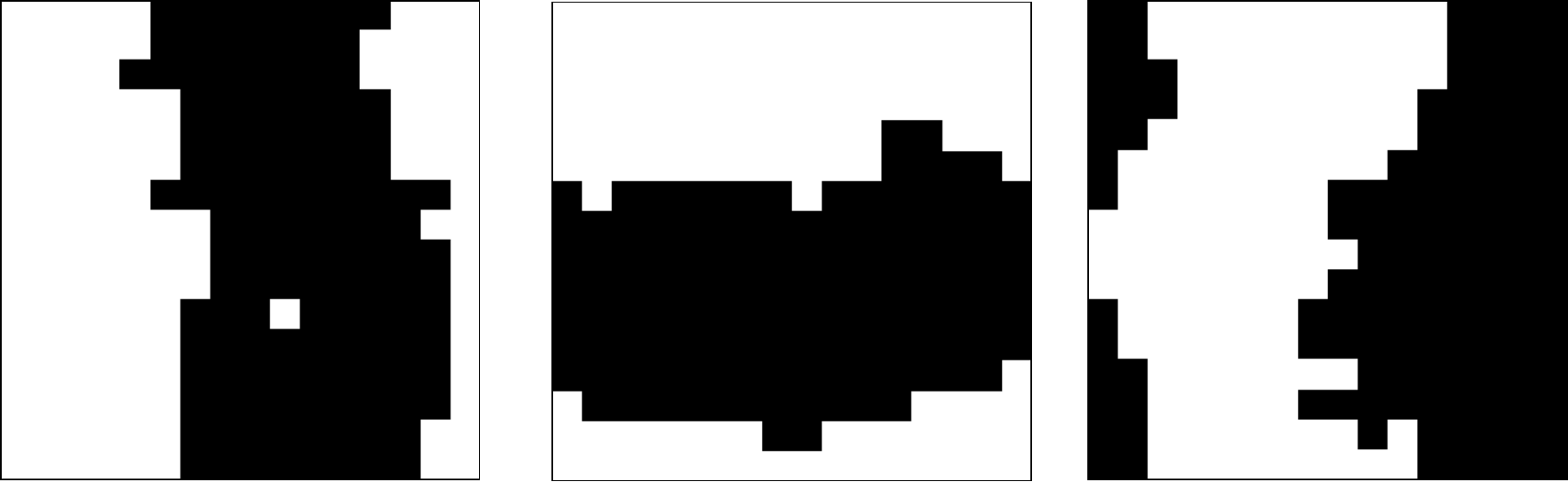}
    \caption{Three extended domain wall configurations of the Ising model. These snapshots are correctly classified as ordered by the energy, but are misclassified if the network only learns its magnetization. Such configurations are common in local update Monte Carlo simulations.}
    \label{fig:edw}
\end{figure}

We therefore choose to run the simulation for a large but fixed number of steps $N$ so that a few snapshots of the training data are not fully equilibrated. In particular, local-update Monte Carlo simulations like ours often get ``stuck'' in metastable configurations before reaching equilibrium. By fixing a stopping point for our simulations, we necessarily include some metastable snapshots in our data. More specifically, we include so-called extended domain wall configurations, which are ordered snapshots that have an anomalously low magnetization~\cite{Suchsland2018May}. These snapshots are correctly classified as ordered by looking at their energy, but are misclassified if our network only learns to discern magnetization. A few examples of these extended domain wall configurations are shown in fig.~\eqref{fig:edw}. By including such snapshots in the training data, we allow our network to learn both the magnetization and the energy.

Another important difference between the theoretical Ising model and our snapshots is the finite size of our grid. 
The main effect of finite system size is that the critical point at $T_c$ extends to a critical region surrounding $T_c$. In an infinite system, the correlation length~\footnote{the correlation length in the Ising model roughly corresponds to the scale on which spins are correlated with each other in a snapshot} between spins diverges precisely at $T_c$. In our finite lattice, however, as soon as the correlation length is larger than the system size ($L=16$), the system behaves critically. In training our model, this finite-size effect actually provides a useful test of whether the model overfits.  Roughly $10 \%$ of our training snapshots are in this ``critical region'', and therefore cannot be classified using any physical quantities. If the model's accuracy on the training data is near $100 \%$, then, it means that it must be memorizing the phase labels of these ambiguous points. On the other hand, after the model has learned, the test accuracy drops to around $95 \%$, which is the best it can do using physical features of the snapshots.

\section{Feature development in the Ising CNN}
Here we provide the by-layer breakdown, across all seeds, of the features learned by the model trained on the Ising task for the grokking and learning cases. We note first that features become better defined as we step through the network. In the first CNN layer, there appears to be no information about the energy or absolute magnetization. In the second CNN layer this information emerges, and then is processed by the hidden layer, where the absolute value of the correlation to the energy in each neuron is around 0.99 and to the absolute magnetization around 0.9. We note in passing that the fully connected layer can be considered to be acting as a non-linear probe on the second convolutional layer \cite{bengioprobes}, which supports the view that the information about the energy is already formed at this point in the network.\\

\begin{figure}[h!]
    \includegraphics[width=1\linewidth]{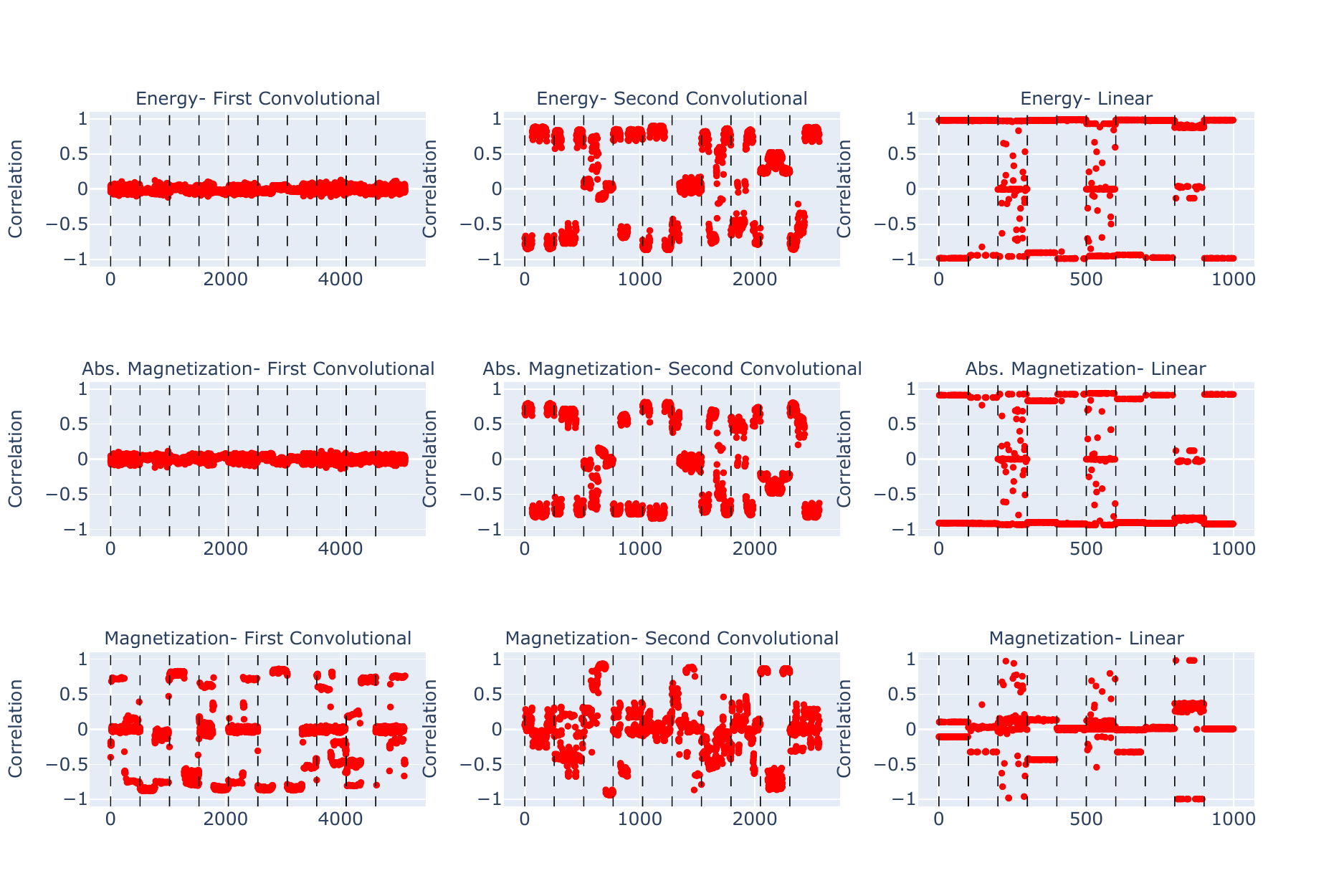}
    \caption{A summary of the correlation coefficients for the different measures across seeds and all layers for the grokking case in Ising. Vertical dashed lines separate different seeds. We can see that although the dominant behaviour is still to learn the magnetization, the grokking case is more noisy than learning - in some seeds learning the magnetization.}
    \label{fig:ising_wm10_intrp}
\end{figure}

We also note that the representation is more noisy in the grokking than in the learning regime. In the learning regime \cref{fig:ising_wm10_intrp}, every seed has almost all of it's fully connected layer neurons close to perfectly correlated or anti-correlated with the energy. In the grokking regime \cref{fig:ising_wm10_intrp}, three of the seeds have significant deviation from perfect correlation with the energy, and instead correlate to the \textit{raw} value of the magnetization. 


\begin{figure}[h!]
    \includegraphics[width=1\linewidth]{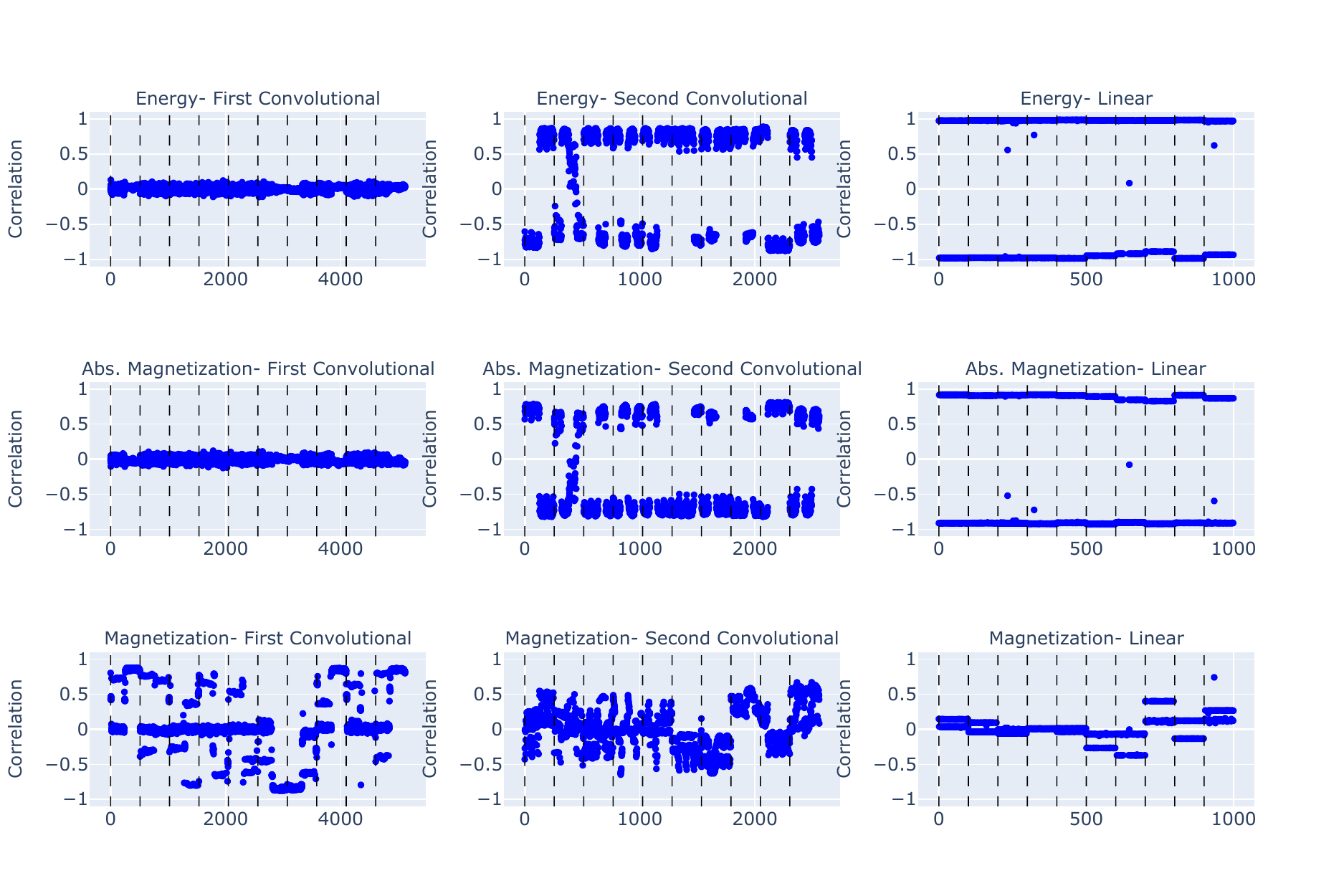}
    \caption{A summary of the correlation coefficients for the different measures across seeds and all layers for the learning case in Ising. Vertical dashed lines separate different seeds.}
    \label{fig:ising_wm1_intrp}
\end{figure}

\section{Compressibility}
\begin{figure}[h!]
    \centering
    \includegraphics[width=\linewidth]{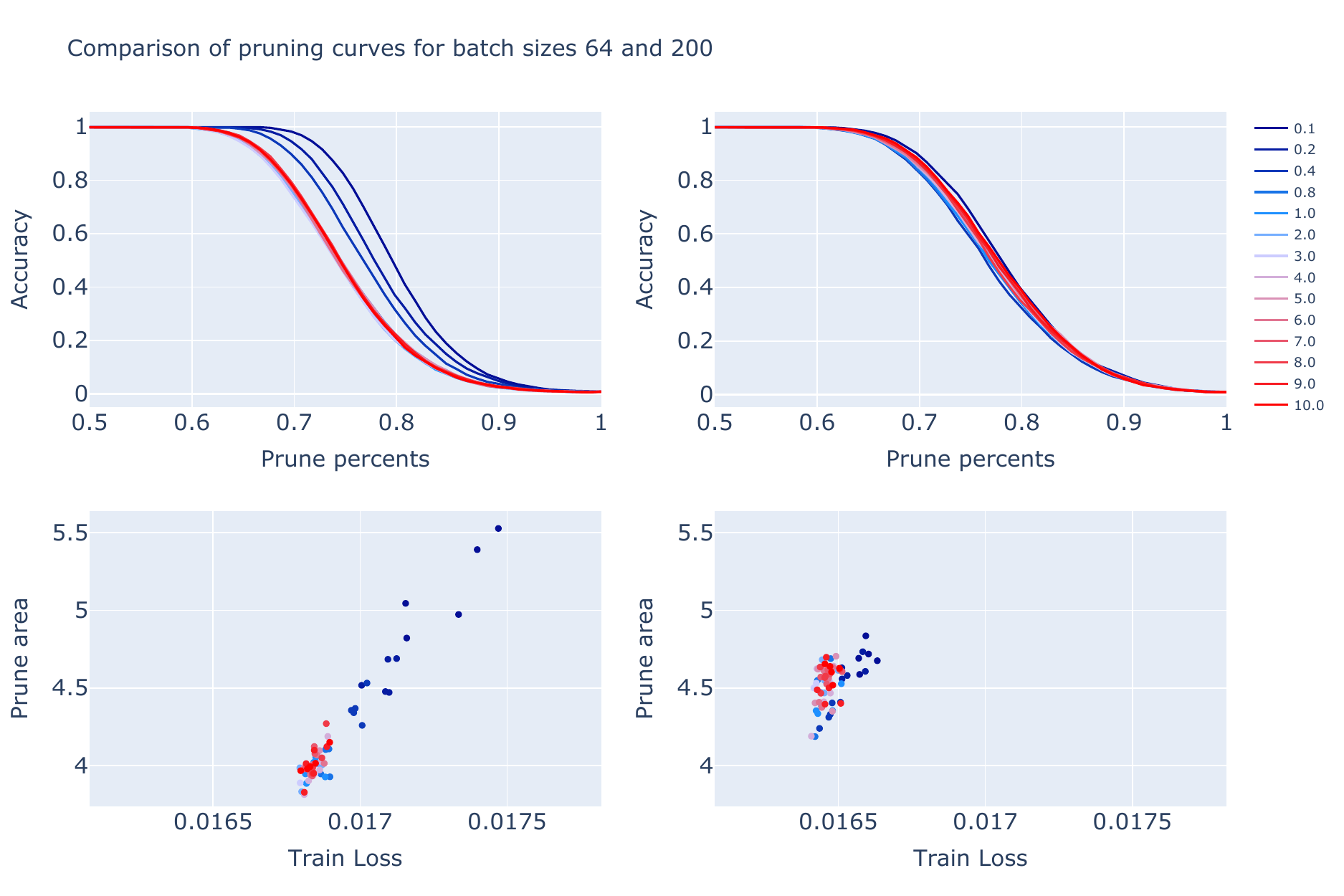}
    \caption{Pruning curves for batch size 64 (left) as in the main text, and 200 (right). All other parameters are as in \cref{tab:model-params}. We see that changing the batch size eliminates the compressive regime. At batch size 200 there is no significant increase in compressibility at lower weight multiplier.}
    \label{fig:113_64_prune}
\end{figure}

We present additional results for compressibility. In the modular addition task, we identified a ``compressive regime" which is characterized by a linear trade-off between end model loss and compressibility. Whether or not this regime exists depends on a wider range of model parameters. Although we have not done an exhaustive parameter search for the compressive regime, we have found that it is increasing the batch size to 200 from 64 in the main text eliminates the compressive effect, as shown in \cref{fig:113_64_prune}.

\begin{figure}[h!]
    \centering
    \includegraphics[width=\linewidth]{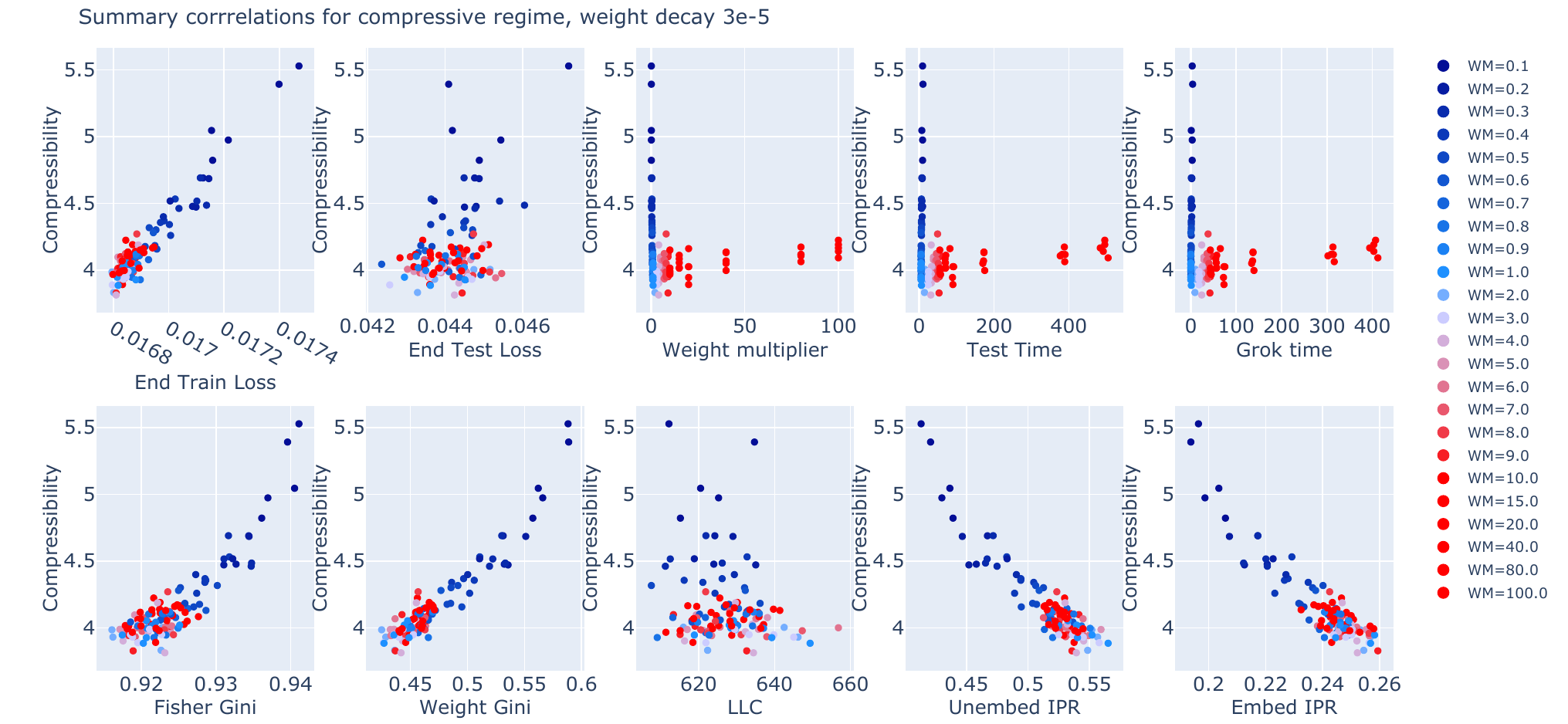}
    \caption{Correlations between model compressibility and other model properties for the parameters considered in the main text. We note that the although the train loss trades off linearly with the compressibility, there is no relationship with the magnetization. In the compressive regime, we also see a reduction in localization in the Fourier basis, as measured by the IPR.}
    \label{fig:113_64_corr}
\end{figure}

\begin{figure}[h!]
    \centering
    \includegraphics[width=\linewidth]{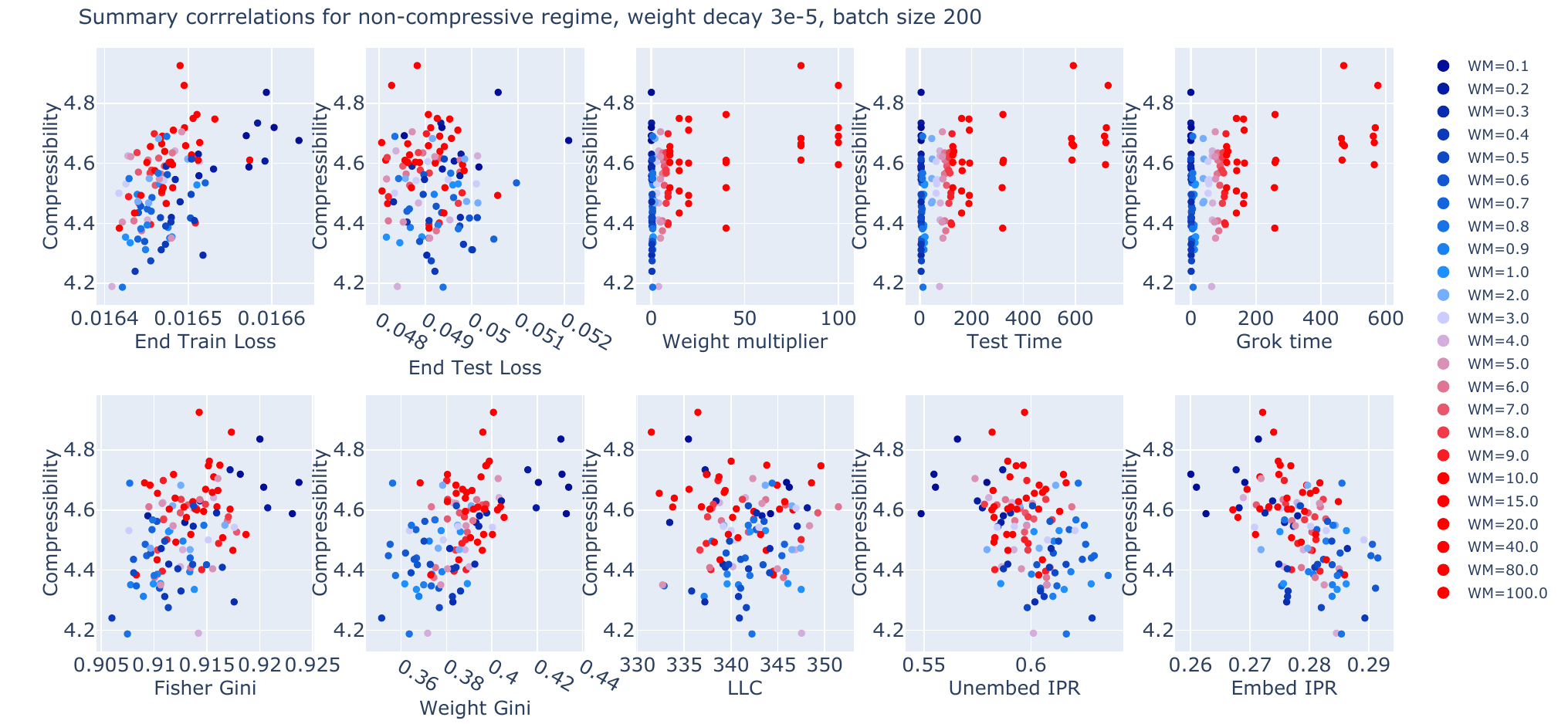}
    \caption{Correlations between model compressibility and other model properties for batch size 200, and remaining parameters as in the main text. Note that here, there is no relationship between grokking and the IPRs.}
    \label{fig:113_200_corr}
\end{figure}

To better understand the nature of the compressive regime, we also provide data for the correlation between model compressibility and a range of other measures in \cref{fig:113_64_corr}. We also provide equivalent data for the non-compressive, batch size 200, runs in \cref{fig:113_200_corr}. We first note that, despite the linear relationship with the \textit{train} loss in the compressive regime, there is no relationship with the \textit{test} loss in either the compressive or non-compressive regime. \\

Second the compressive regime is associated with a significant reduction in the localization of neurons to a particular frequency. We see in the rightmost plots of the bottom row that for both input and output weight matrices to the hidden layer, the compressive regime leads to an improvement in compressibility through a decrease in Fourier basis localization. In the extremely compressed regime that we discuss in \cref{sec:high_compression}, this effect becomes very dramatic. The low weight multiplier runs are associated to an almost complete breakdown of Fourier basis localization (\cref{fig:113_64_corr}). It is important to point out that the reduction in the Fourier basis localization is a property of the compressive regime and not a difference between grokking and learning generally. In \cref{fig:113_200_corr} we provide data for the same measures, but at the higher batch size. Here, there is no significant difference between the IPRs of grokking and learning. \\

Third, unsurprisingly, more compressible models have higher weight Gini coefficients. In addition to the Gini coefficients of the weight magntidudes, we also measure the Gini coefficient of the diagonal entry corresponding to a given weight in the Fisher Information Metric as discussed in \cref{app:fim}. The Gini coefficients of the Fisher diagonals are substantially higher than the Gini coefficients of the raw weights. This should be unsurprising, since the Fisher diagonals should give a more refined measure of the importance of a given weight. Given this observation, is it puzzling why pruning on the basis of the Fisher diagonals gives worse performance than magnitude pruning, as we shown in \cref{sec:fisher_pruning}. Resolving this tension could help us to understand how the FIM can be used in ablations and other in other interpretability contexts.\\

Finally, we also compute the ``local learning coefficient" (LLC) for all runs, using code from \cite{ninaLLC}. The LLC is intended to be a measure of a model degeneracy. Intuitively, models with higher degeneracy should be expected to be more compressible. Surprisingly, there have not been many studies that investigate the relationship between the LLC and pruning in a controlled way. Since the compressive regime provides a way to tune the model compressibility we are able to do so in this context. Interestingly, we find no relationship between the LLC and the model's compressibility. A deeper understanding of this observation could open the way towards developing the LLC as a tool for model compression.

\begin{figure}[h]
    \centering
    \includegraphics[width=\textwidth]{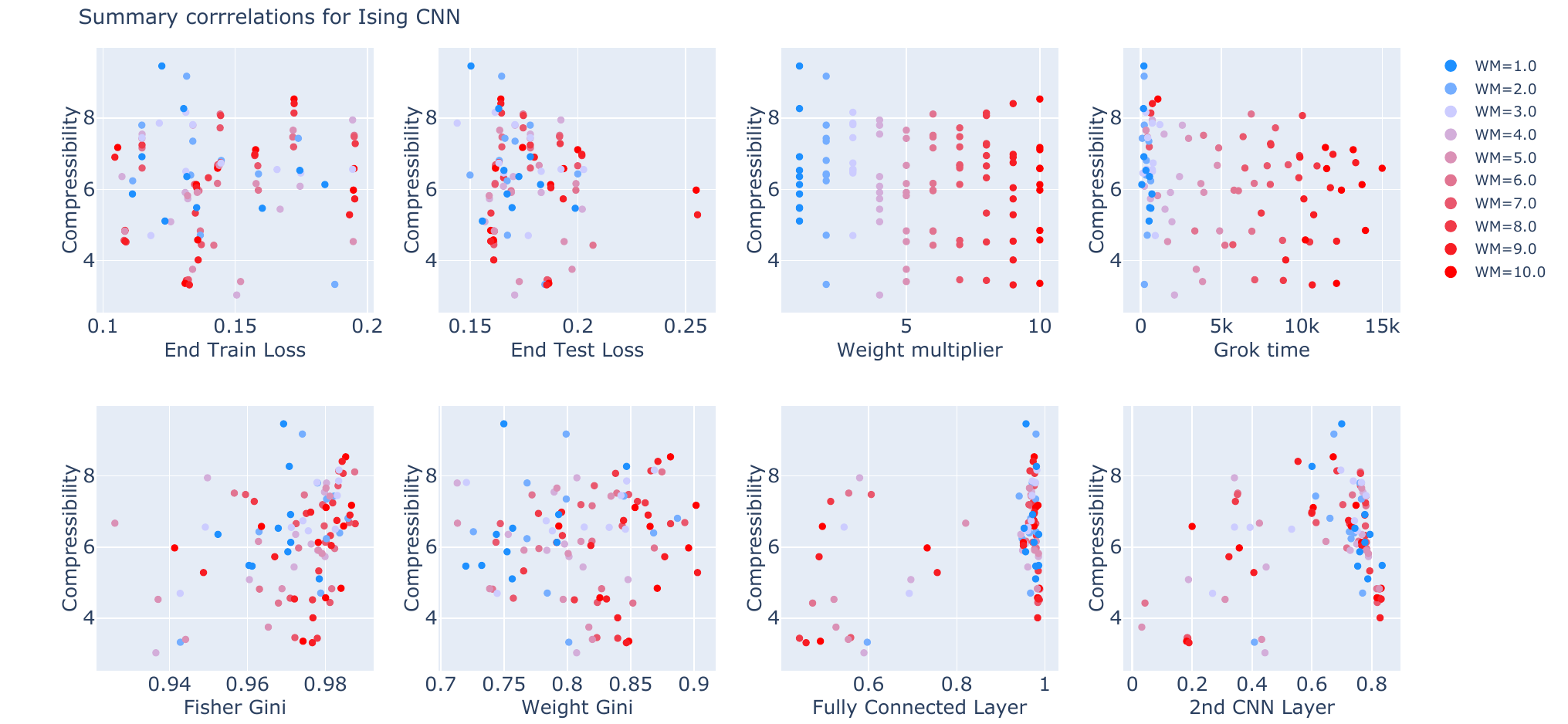}
    \caption{In the Ising case, there is no compressive phase, and so there is no trade-off between the end-loss and the model compression accessible by tuning the weight multiplier.}
    \label{fig:ising_pruning_corr}
\end{figure}

\section{25x compression in modular addition}
\label{sec:high_compression}

\begin{figure}[t]
    \centering
    \includegraphics[width=0.99\linewidth]{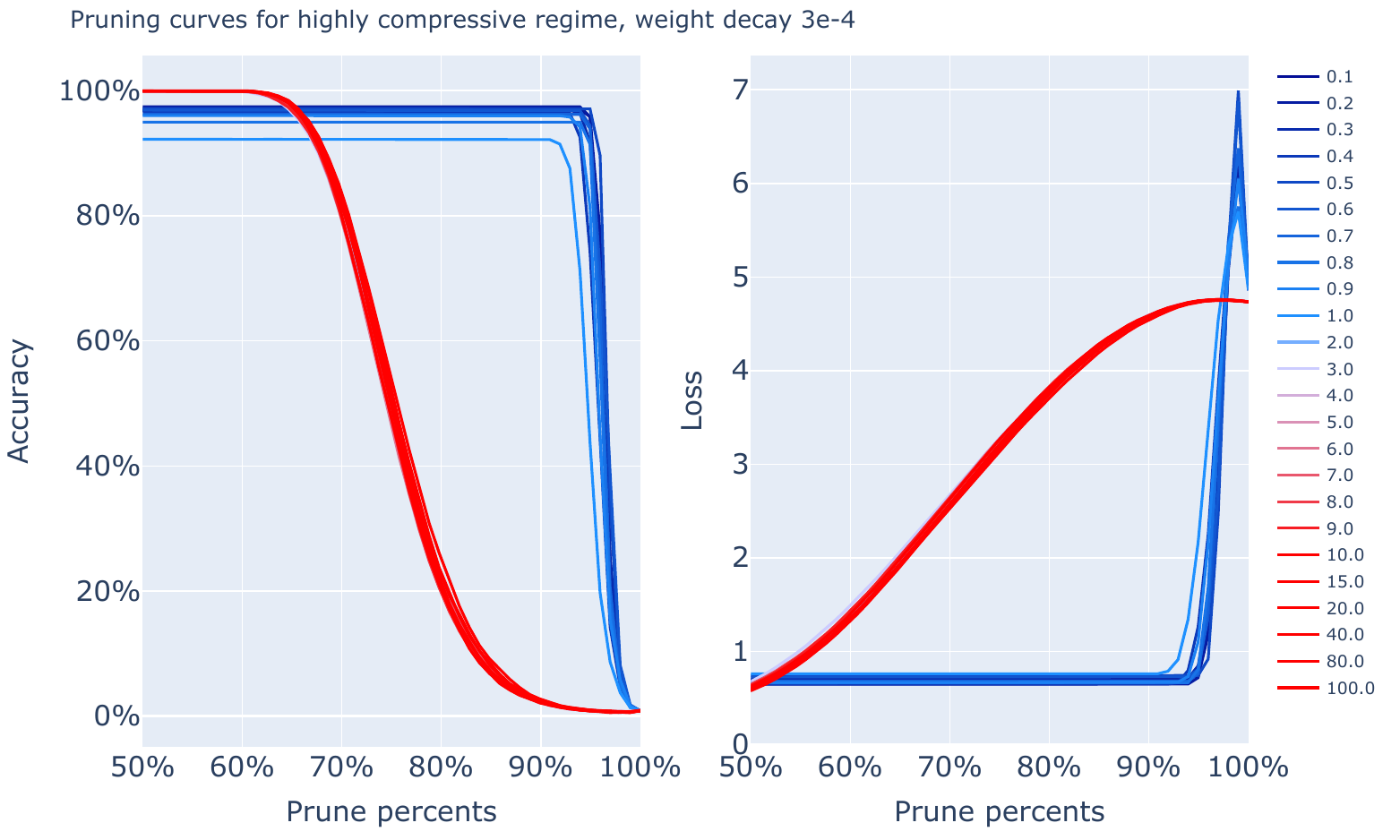}
    \caption{Pruning curves for the highly compressed curves in the top left of the modular addition pruning areas curve. There is a very sharp transition to very highly compressed models between weight multipliers $w_0=2$ and $w_0=1$.}
    \label{fig:mod_pruning_compressed}
\end{figure}

Conducting a rough parameter search, we found a range with extremely compressed models for weight decay $3\times 10^{-4}$, an order of magnitude larger than that considered in the main text. In \cref{fig:mod_pruning_compressed}, we show the pruning curves for the same parameters as in the main text, but at the higher weight decay. We see a very large improvement in the pruning, that occurs suddenly as we transition into the compressive regime. Model compressibility increases suddenly from around 32\% of the weights needed to maintain an accuracy of 95\% at $w_0=2$ to only 5\% of the weights needed to maintain the same accuracy at $w_0=1$. This corresponds to a compressibility of \textbf{25x the base model, and 5x the compression achieved in the main text.}

\begin{figure}[t]
    \centering
    \includegraphics[width=0.99\linewidth]{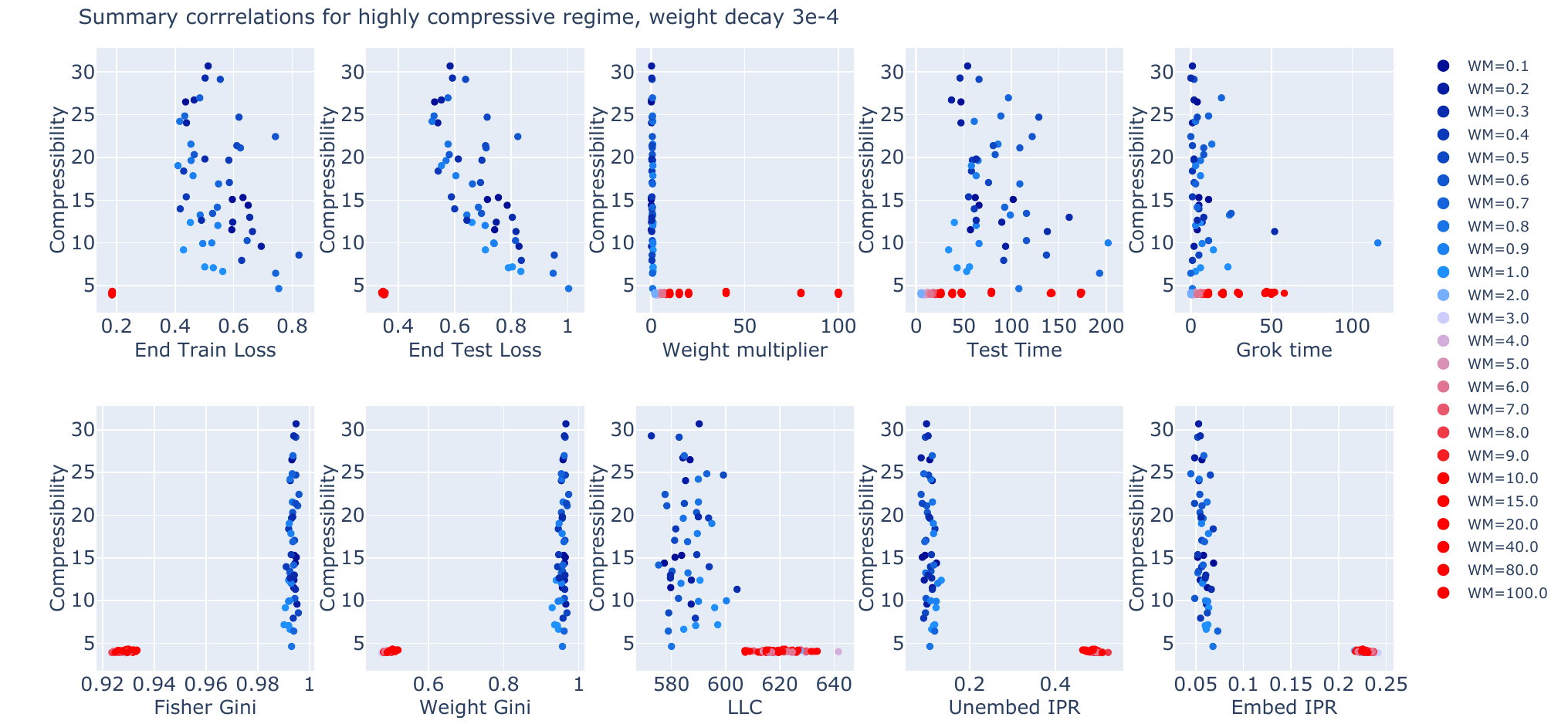}
    \caption{Correlations between the compressibility and measures of training dynamics and model features in the highly compressive regime (weight decay $3\times 10^{-4}$.}
    \label{fig:mod_compressed_corr}
\end{figure}

To better understand this highly compressed regime, we again provide data for the same measures as considered in the previous section in \cref{fig:mod_compressed_corr}. We see a very stark difference between the compressive regime and the other weight multipliers. Most strikingly, we see in the two rightmost sub-figures of the bottom row that the Fourier basis localization has almost completely broken down. Interestingly, in this regime, model train and test loss are negatively correlated to the compressibility.

\clearpage
\section{Fisher Information Geometry}
\label{app:fim}
This section provides a whistle-stop tour of the basics of information geometry. For a more complete treatment, see for example \cite{amari2016information}.

Given an architecture function, which for us we take to be a neural network, $f_{NN}(x_{\text{input}})$, with parameters $\theta_i$, the space of neural networks with this architecture is spanned by the allowed values of the parameters.
Each weight and bias make take any real value such that this model space is homeomorphic to $\mathbb{R}^N$ where there're $N$ parameters.
Each set of parameter values defines a point on the model space, and as the parameters change smoothly\footnote{Practically the updates are discrete and hence the path is not smooth, but we can consider the infinitesimal update limit for this theoretical illustration.} through training a path on the model space is traced out, forming a trajectory.

To define a notion of distances on the model space we require a metric. 
Where the task we consider is a classification problem (as in the tasks considered in this work), the trained function output is a normalised probability distribution (often enforced by a final softmax activation).
For this we first assume the model space is a manifold, and then consider the tangent space, traditionally spanned by a a basis $\{\frac{\partial}{\partial \theta_i}\}$ but may also suitable by spanned by the basis of score vectors\footnote{Note the classification nature of the task is essential here, as non-negativity of the outputs is required for logarithms to be well defined over the real numbers.} $\ell_i = \frac{\partial}{\partial \theta_i}\ln(f_{NN}(x|\theta))$ which is the basis we choose $\{\ell_i\}$. 
The Fisher metric is then defined on this tangent space basis as:
\begin{equation}
\begin{split}
    g^{FIM}_{ij} := & \ \mathbb{E}_x (\ell_i\ell_j)\;,\\
    = & \ \mathbb{E}_x \bigg( \frac{\partial \ln(f_{NN}(x|\theta))}{\partial \theta_i}  \frac{\partial \ln(f_{NN}(x|\theta))}{\partial \theta_j} \bigg)\;.
\end{split}
\end{equation}
The expectation is theoretically an integral over the entire data space $x$, however practically we compute a discrete expectation as an average over samples from the training data. 
The metric changes depending on the position in model space since $f_{NN}$ is a non-linear function of $\theta$, and this is what defines the non-flat geometry of the space.
The discrete derivatives are computed using \texttt{pytorch} functionality, via the \texttt{nngeometry} package \cite{george_nngeometry}.

The Fisher information metric provides a measure of the similarity between two models, well illustrated by the connection to the KL-divergence, $D_{KL}$, which measures the relative entropy between probability distributions $p(x|\theta)$
\begin{equation}
    D_{KL}(\theta' | \theta) = \mathbb{E}_x \bigg( \ln \bigg(\frac{p(x|\theta')}{p(x|\theta)} \bigg)\bigg)\;,
\end{equation}
which has a unique global minimum of 0 when $\theta' = \theta$, and in the neighborhood of this minimum the KL-divergence expands as
\begin{equation}
    D_{KL}(\theta + \delta\theta | \theta) = \frac{1}{2}g^{FIM}_{ij}(\theta)\delta\theta_i\delta\theta_j + \mathcal{O}(\delta\theta^3)\;.
\end{equation}
Therefore the FIM acts as an infinitesimal measure of distance in the model space.

In the Ising task, the neural network architecture used consists of 6798 parameters, whilst for the modular addition task the architecture has 174193 parameters.
This leads to real symmetric non-degenerate FIM matrices of respective sizes $g^{FIM}_{6798 \times 6798}$, and $g^{FIM}_{174193 \times 174193}$.
However, since the number of parameters is large, one can use the assumption that these matrices are approximately diagonal \cite{amari_fimdiagapprox} to ease computational cost; thus we use a vector of entries $g^{FIM}_{ii}$ to represent the FIM diagonal for each of the parameters $\theta_i$.

During the training the network parameters (i.e. the positions in model space) are recorded at regular intervals. Equivalently the FIM diagonal is also computed at these same intervals.
For the Ising task these were at intervals of 200 epochs for the first 100000 epochs, and for the modular addition task at intervals of 1 epoch for the first 500 epochs; after each of these periods the behaviour between the steady learning and grokking regimes were deemed equivalent.

\begin{figure}[!t]
    \centering
    \begin{subfigure}{0.45\textwidth}
        \centering
        \includegraphics[width=0.98\textwidth]{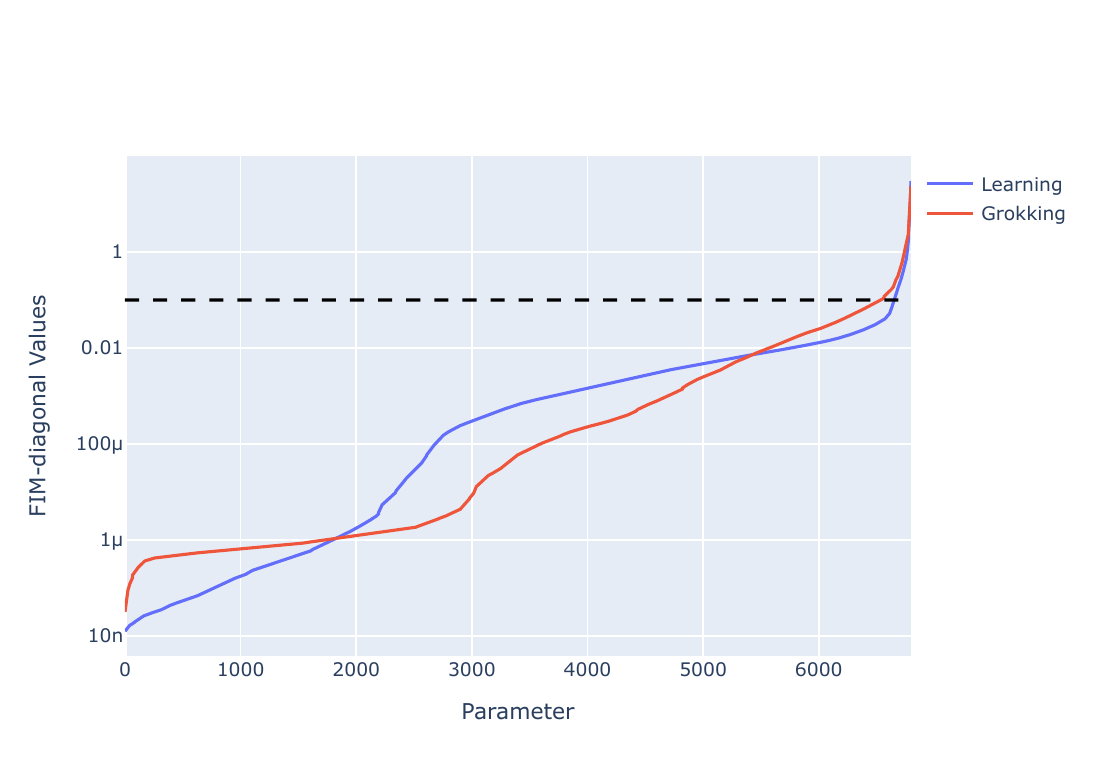}
        \caption{Ising}
    \end{subfigure} 
    \begin{subfigure}{0.45\textwidth}
        \centering
        \includegraphics[width=0.98\textwidth]{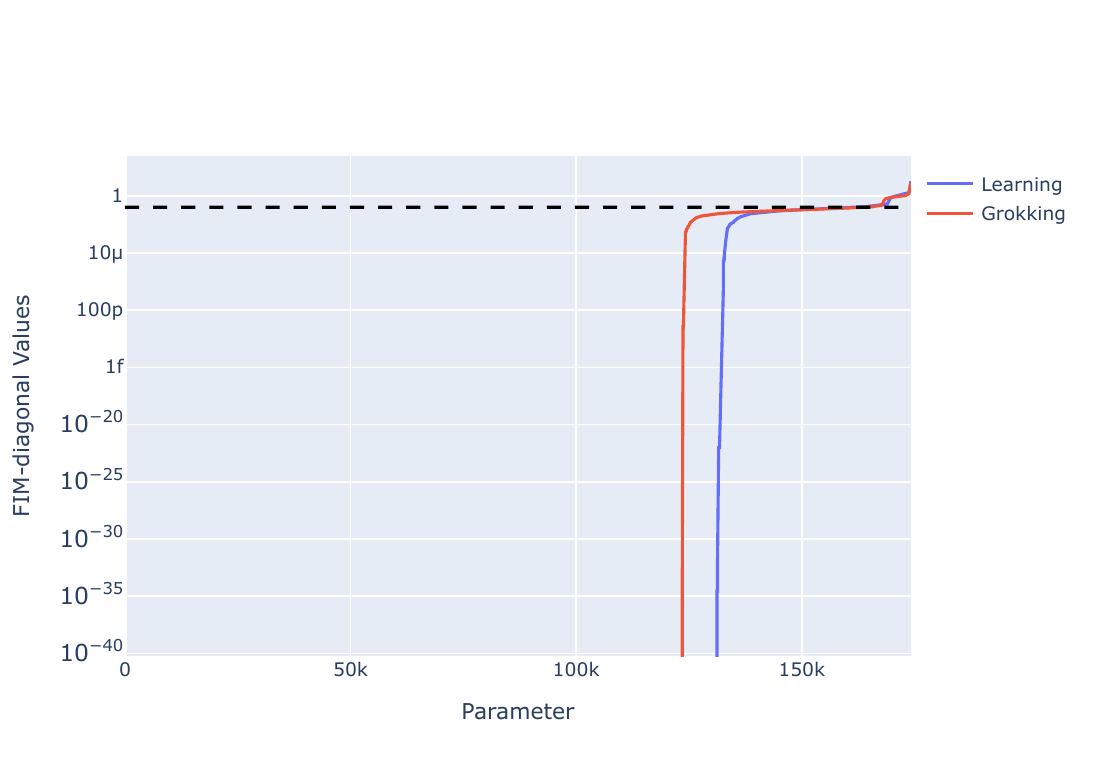}
        \caption{ModAdd}
    \end{subfigure} 
    \caption{FIM diagonal values for the neural network model after training, sorted increasing values, and averaged over the 10 seed runs. Plots showing both the learning and grokking regimes; for both the Ising (a), and ModAdd (b) tasks. The dashed line represent the 0.1 value set as the threshold for the subspace splitting, positioned at the spectrum mass gaps.}
    \label{fig:dynamics_fimdiags}
\end{figure}

For the neural network models trained on the tasks considered in this work, the final FIM diagonal spectrums after training, averaged over the 10 seed runs, can be computed and plotted (in importance increasing order for each parameter); shown in \cref{fig:dynamics_fimdiags}.
These spectrums show gaps in all cases, which are approximately at the same positions, and the dashed line marks the 0.1 value selected as the threshold for the stiff $\times$ sloppy splitting of the model space.

As described by the works \cite{Berman:2023rqb}, one may consider the FIM diagonal as a measure of the relative importance of parameters to the model.
\subsection{Fisher pruning} 
\label{sec:fisher_pruning}
As an alternative to magnitude pruning, this section applies the Fisher pruning
(BRG) scheme \cite{Berman:2022mak, Berman:2022uov, Berman:2023rqb,
howard2024bayesianrgflowneural} to the models studied throughout the note.

The geometry of model space is often assumed to be flat, such that a
Euclidean metric\footnote{Where the Euclidean metric is, for practical purposes,
the identity matrix.} can be used to compute distances. However one can compute
a more natural information geometry metric which directly depends on the
architecture: the Fisher Information Metric (FIM). The FIM, $g^{FIM}_{ij}$, as
introduced in appendix \ref{app:fim}, enables the intrinsic geometry of the
model space to be probed.

Fisher pruning is an information-theoretic network pruning scheme that
generalises the principles of the exact renormalization group (ERG) studied in
high-energy physics to arbitrarily parameterized probability distributions,
including those of NNs. Pruning is performed in parameter space with respect to
an information-theoretic distinguishability scale set by the Fisher information
metric. Fisher pruning removes neural network parameters which covary weakly
with the output of the model\footnote{This technique is also known as BRG
\cite{howard2024bayesianrgflowneural}.}.
After specifying an architecture for a neural network, one may treat the
parameters as a natural set of coordinates for the space of models (associated
with the given architecture). Initializing the network involves sampling a
random point in model space. During training, the parameters change smoothly,
and the position in model space updates tracing out a trajectory.

In the works \cite{Berman:2022mak, Berman:2022uov, Berman:2023rqb} the
eigenspectrum of the FIM is argued to provide a parameter importance measure,
allowing the model function to be analysed and pruned in detail. The eigenvalues
give this measure of relative importance for linear combinations of parameters
(defined by the eigenvectors) to the model. In the case where the metric is
approximately diagonal, valid for high-dimensional model spaces
\cite{amari_fimdiagapprox}, the eigenvalues are equal to the diagonal entries
and these give a relative importance of each parameter to the model.

Intuitively one may believe magnitude pruning to be sufficient to determine the
compressibility of a model \cite{lecun1989optimal}; this is typically not the
case\footnote{There do exist some exceptions; for example, magnitude-pruning a
very simple single-layer perceptron network, with monotonically increasing
activation functions, trained subject to the maximisation of output vectors
would be sufficient.}. In practice, deep networks are typically highly
non-linear (which is at least partly responsible for their great ability to
express complex functions). Moreover, modern networks
have built-in non-linearities in the form of activation functions. In the case
that the activation function $\sigma(x)$ does not monotonically increase, i.e.
there exists an $x$ such that $\sigma(x + \delta) \leq \sigma(x)$, for $\delta >
0$, magnitude pruning is poorly motivated. An example of a commonly used
activation function possessing this feature is the GELU activation. A novel
pruning technique, immune to such restrictions, is Fisher pruning scheme
\cite{Berman:2022uov, Berman:2023rqb}. Fisher pruning is performed
hierarchically: parameters are pruned such that those with the smallest
associated diagonal Fisher information metric components are removed first.
The advantages of the Fisher pruning scheme are most manifest when pruning the
model holistically, i.e. each layer simultaneously. 

\begin{figure}[h!]
    \centering
    \begin{subfigure}{0.45\textwidth}
        \centering
        \includegraphics[width=0.98\textwidth]{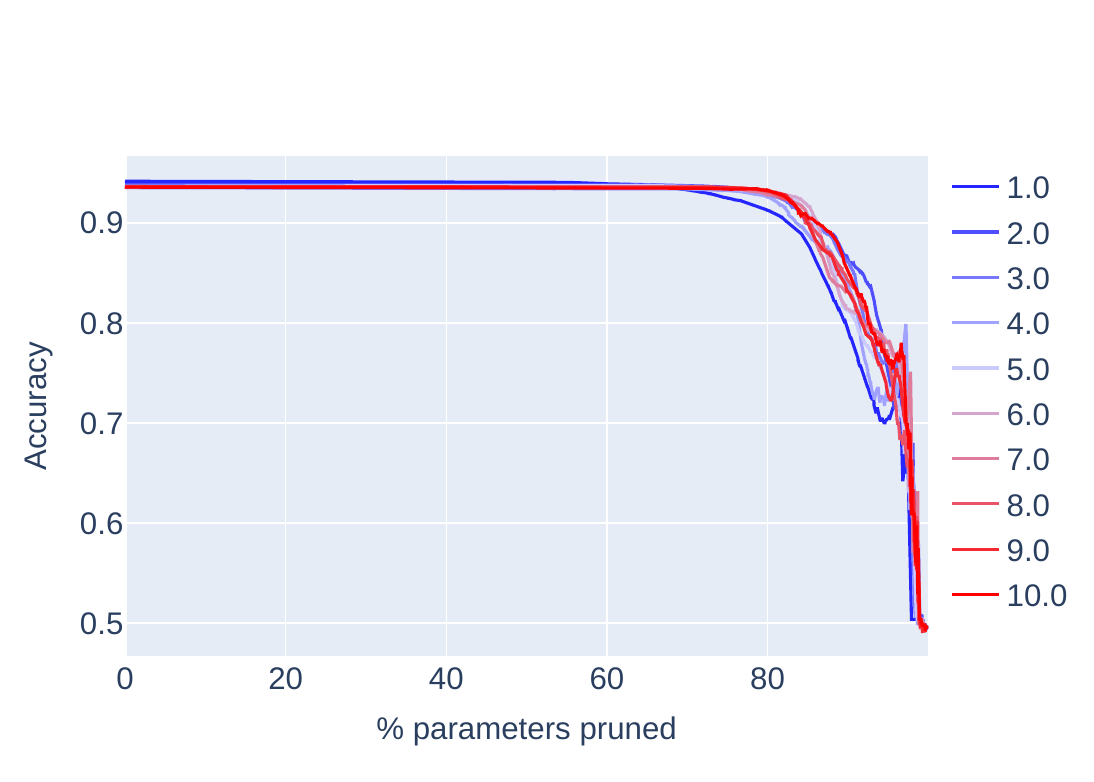}
        \caption{Ising -- Fisher pruning (accuracy)}\label{fig:fisher_pruned_ising_whole_acc}
    \end{subfigure} 
    \begin{subfigure}{0.45\textwidth}
        \centering
        \includegraphics[width=0.98\textwidth]{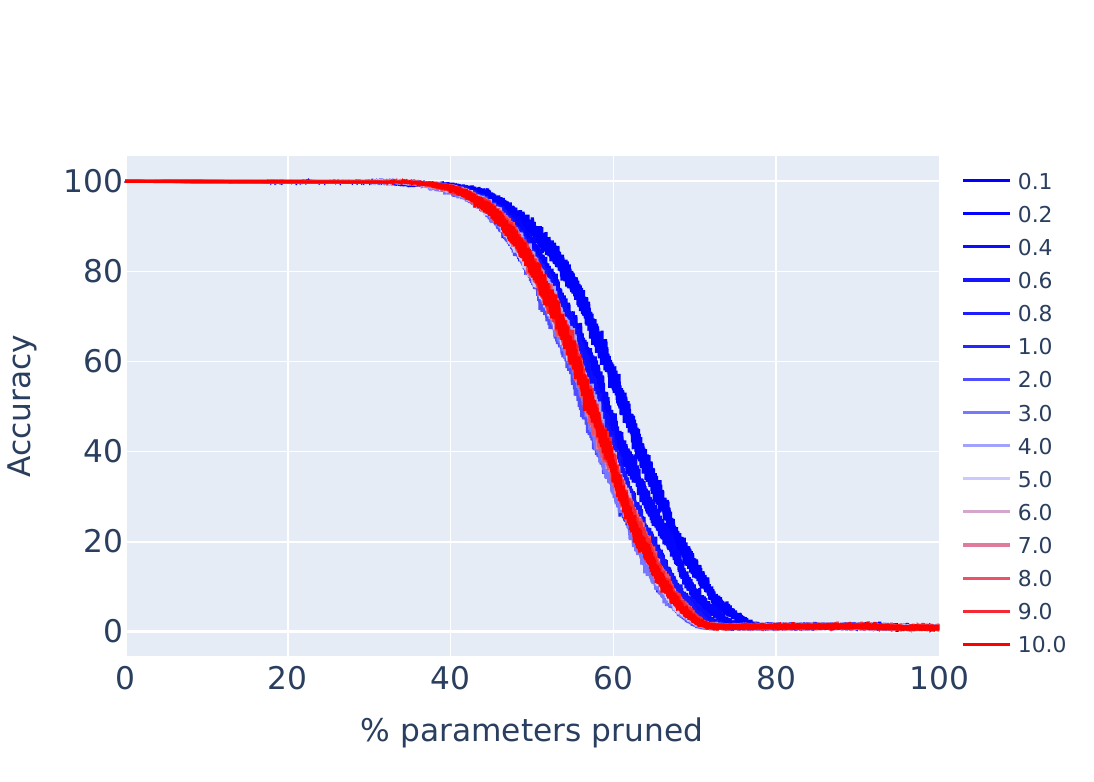}
        \caption{ModAdd - Fisher pruning (accuracy)}\label{fig:fisher_pruned_modadd_whole_acc}
    \end{subfigure}\\
     \begin{subfigure}{0.45\textwidth}
        \centering
        \includegraphics[width=0.98\textwidth]{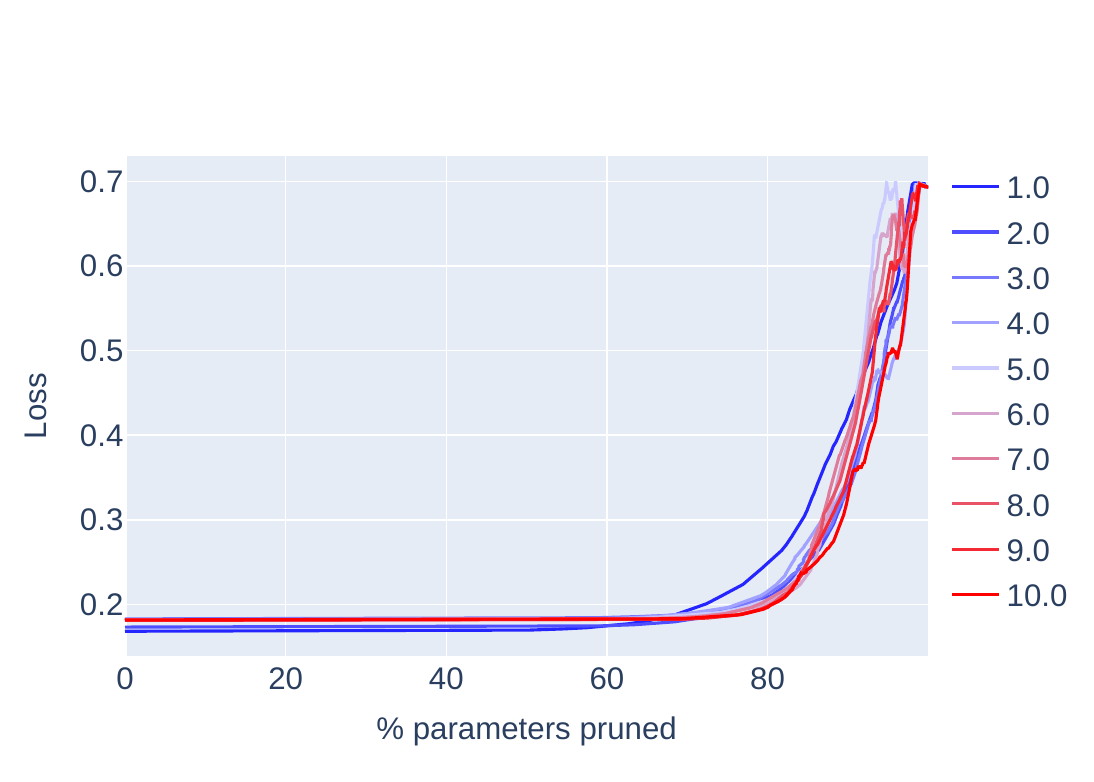}
        \caption{Ising -- Fisher pruning (loss)}\label{fig:fisher_pruned_ising_whole_loss}
    \end{subfigure} 
    \begin{subfigure}{0.45\textwidth}
        \centering
        \includegraphics[width=0.98\textwidth]{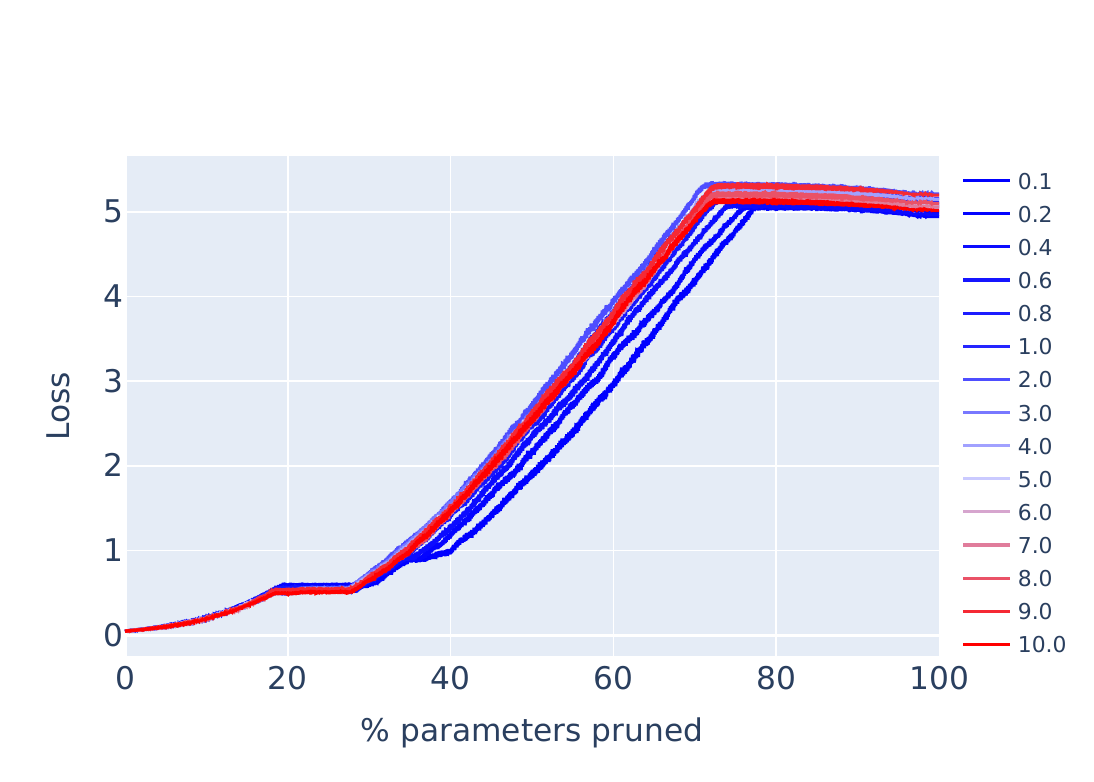}
        \caption{ModAdd - Fisher pruning (loss)}\label{fig:fisher_pruned_modadd_whole_loss}
    \end{subfigure}
    \caption{Accuracies and losses as a function of \% parameters pruned for the grokked and steady-learning regimes (Ising and modular arithmetic networks).}
    \label{fig:fisher_prune_full}
\end{figure}

In \cref{fig:fisher_prune_full}, we present loss and accuracy plots as a
function of number of parameters pruned for both the Ising and modular
arithmetic problems. In agreement with the magnitude pruning analysis, we
conclude hyperparameters conducive to grokking do not consistently result in a
pruning advantage. The Ising network presents an interesting property: there
does not exists a clear pruning advantage in the grokked regime when compared
the steady-learning phase (for a batch size of 200). While one might argue that
the chosen weight multiplier (10) to represent ``grokking" provides a pruning
advantage for the archetypical value selected for steady learning (0.1), it is
evident that other weight multiplier values resulting in learning produce the
opposite effect. We thus conclude that there is not a definitive compressibility
behaviour that can be attributed to grokking in the Ising model. 

The modular arithmetic network (\cref{fig:fisher_prune_full}) tells a similar story,
hyperparameter choices resulting in grokking do not produce a more prunable
model than those which do
not under the Fisher pruning regime (unlike the magnitude pruning regime).
This analysis clearly shows that a grokked model is not necessarily a more
compressed model. Compressibility is attributed to the weight multiplier and not the presence of grokking.

\begin{figure}[h!]
    \centering
    \begin{subfigure}{0.45\textwidth}
        \includegraphics[width=\linewidth]{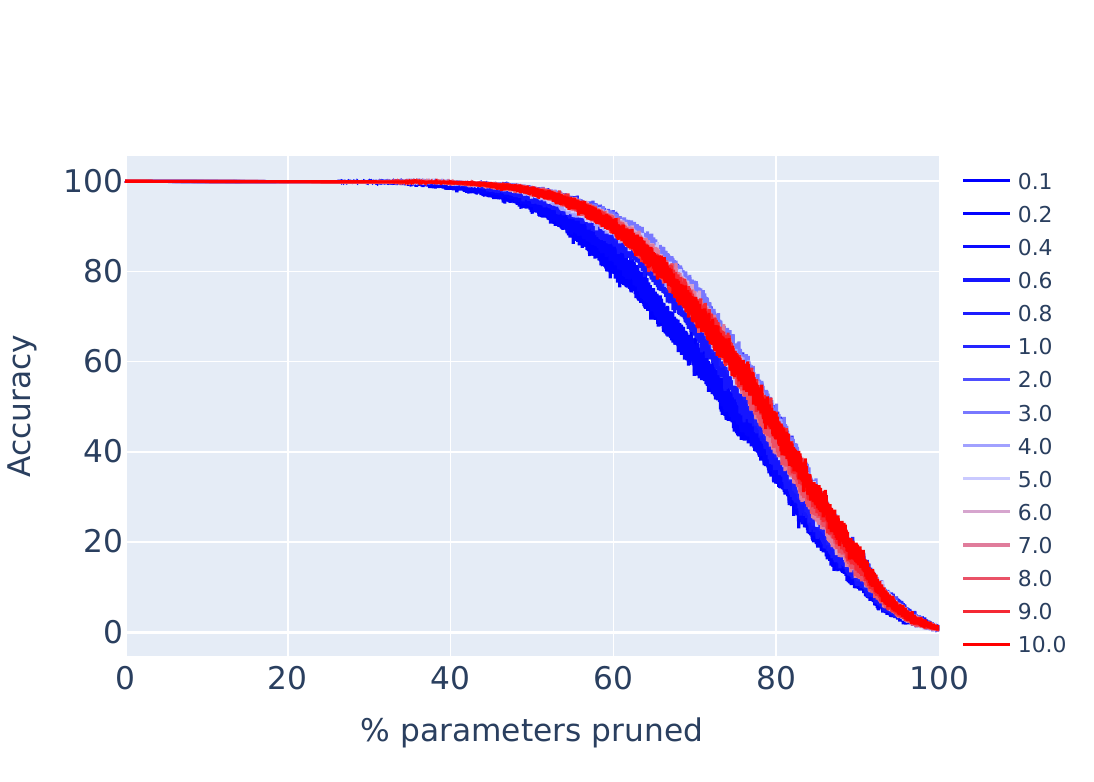}
        \caption{Acc. Layer 1}
    \end{subfigure}
    \hfill
    \begin{subfigure}{0.45\textwidth}
        \includegraphics[width=\linewidth]{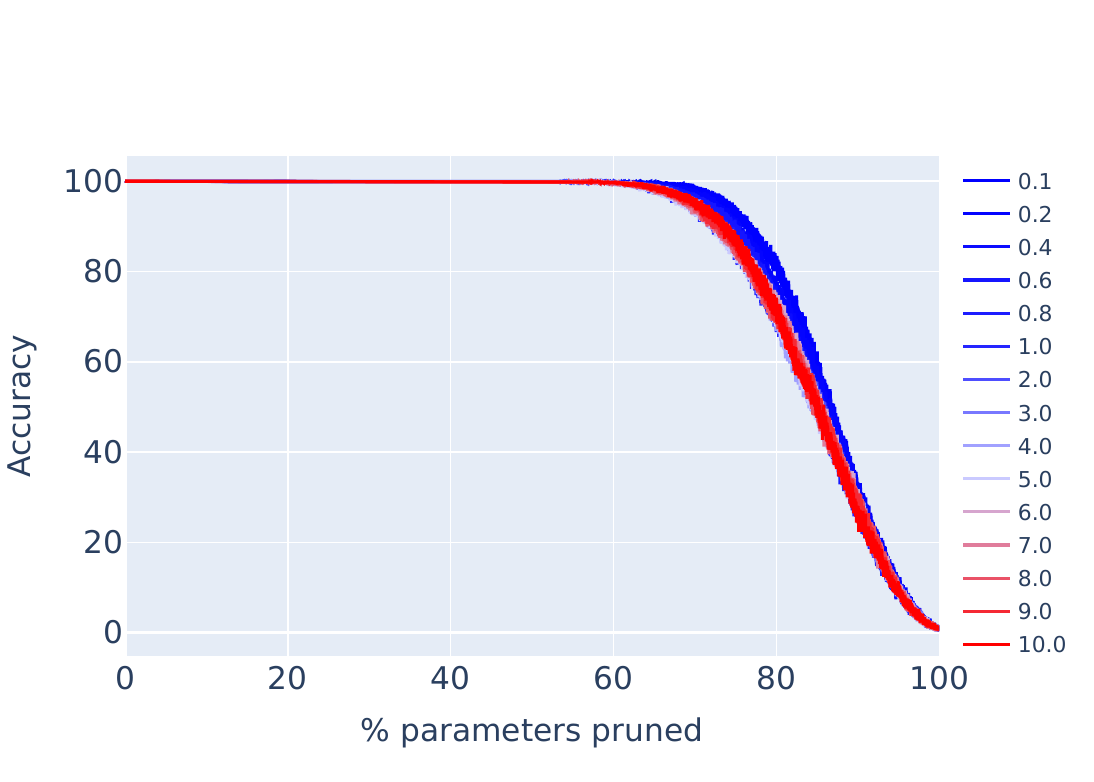}
        \caption{Acc. Layer 2}
    \end{subfigure}

    \begin{subfigure}{0.45\textwidth}
        \includegraphics[width=\linewidth]{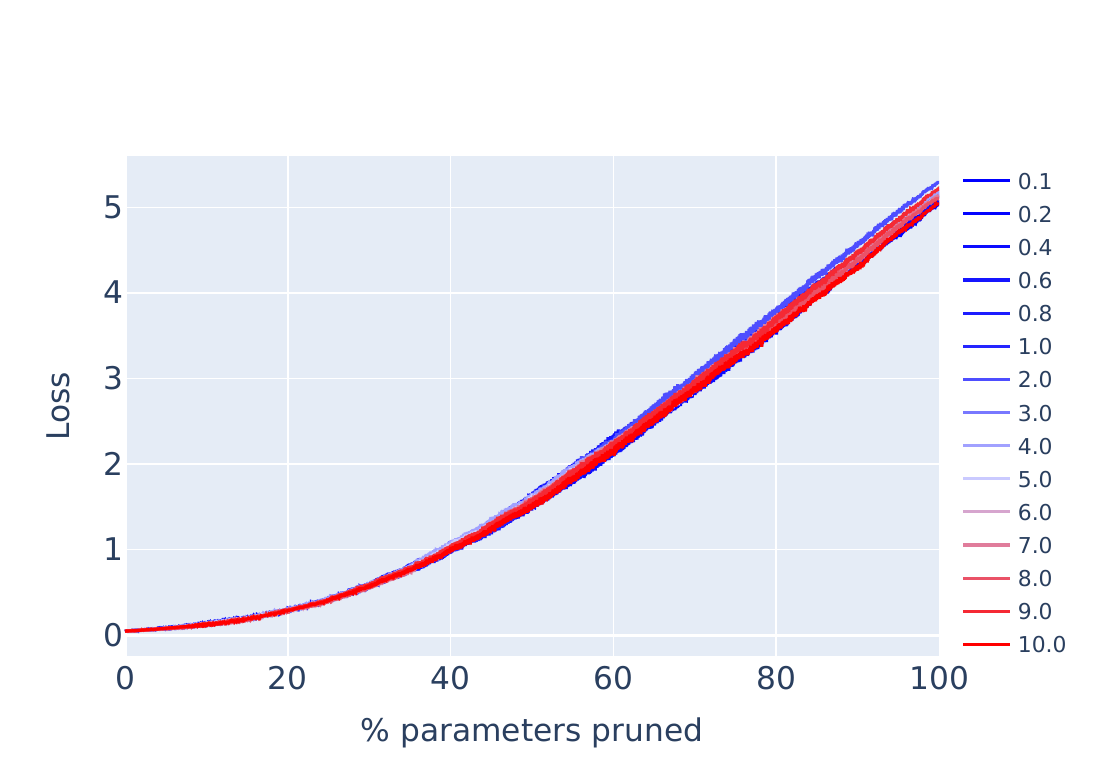}
        \caption{Loss Layer 1}
    \end{subfigure}
    \hfill
    \begin{subfigure}{0.45\textwidth}
        \includegraphics[width=\linewidth]{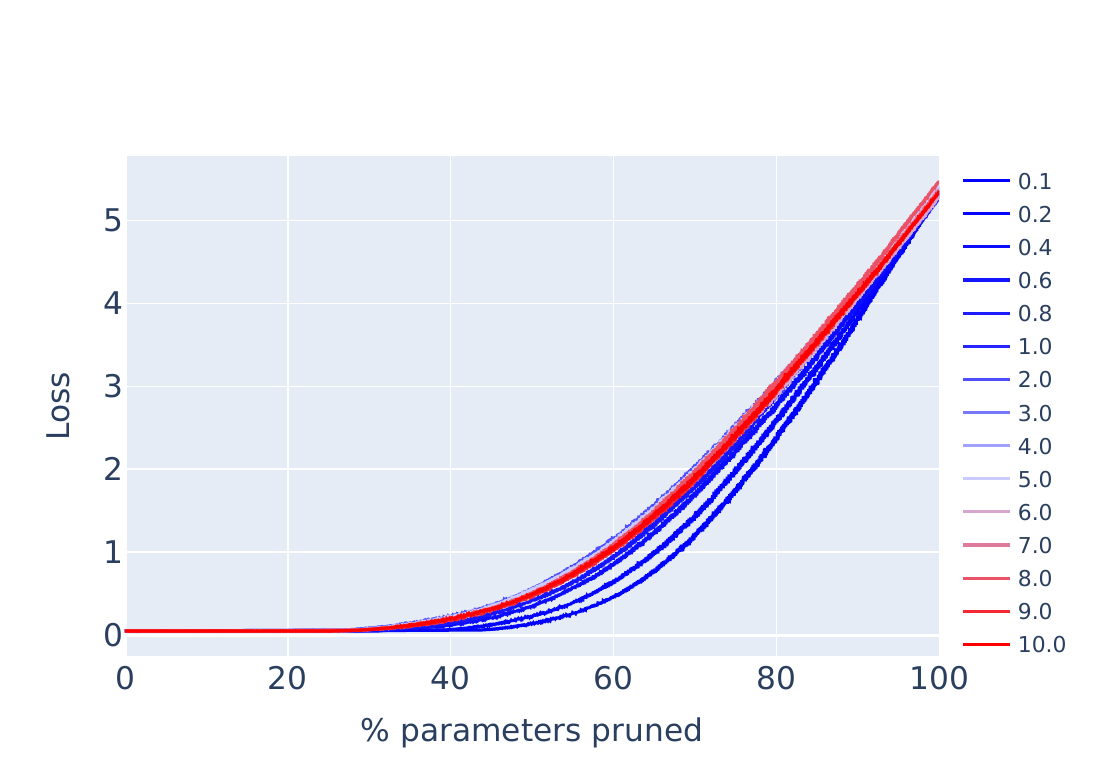}
        \caption{Loss Layer 2}
    \end{subfigure}
    \caption{Losses and accuracies of the modular addition network output as a function of parameters Fisher-pruned}
    \label{fig:fisher_pruning_ma_per_layer}
\end{figure}

\begin{figure}[h!]
    \centering
    \begin{subfigure}{0.45\textwidth}
        \includegraphics[width=\linewidth]{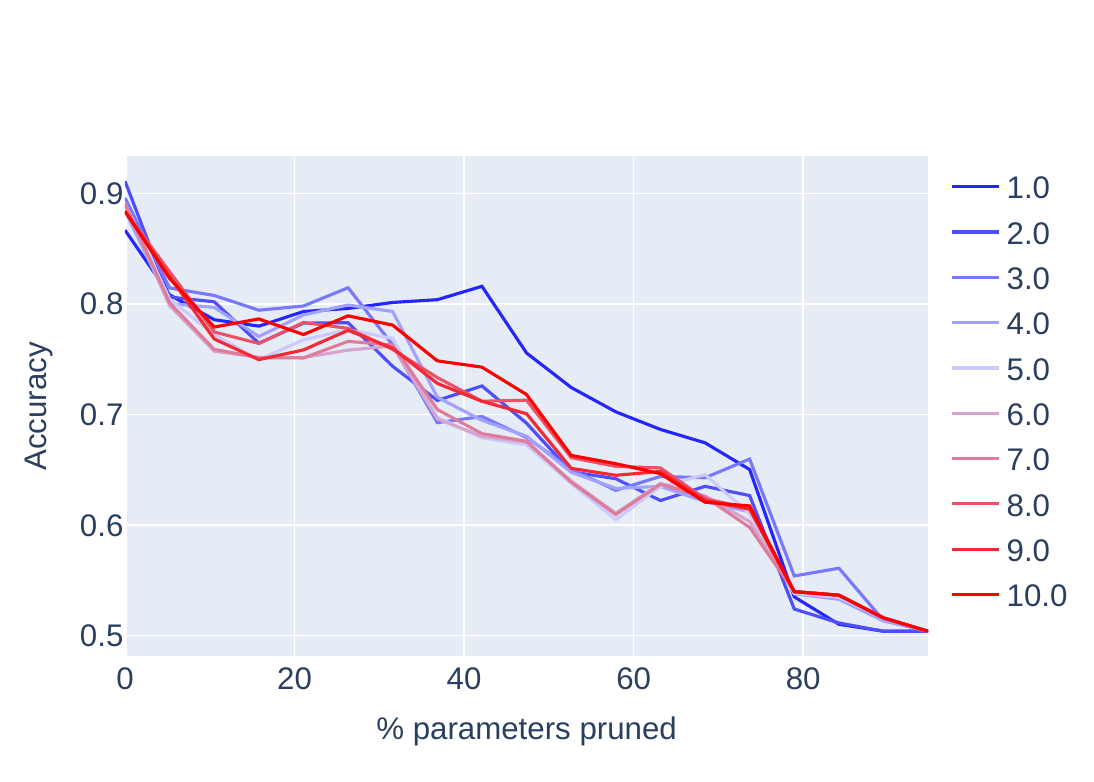}
        \caption{Acc. Layer 1}
    \end{subfigure}
    \hfill
    \begin{subfigure}{0.45\textwidth}
        \includegraphics[width=\linewidth]{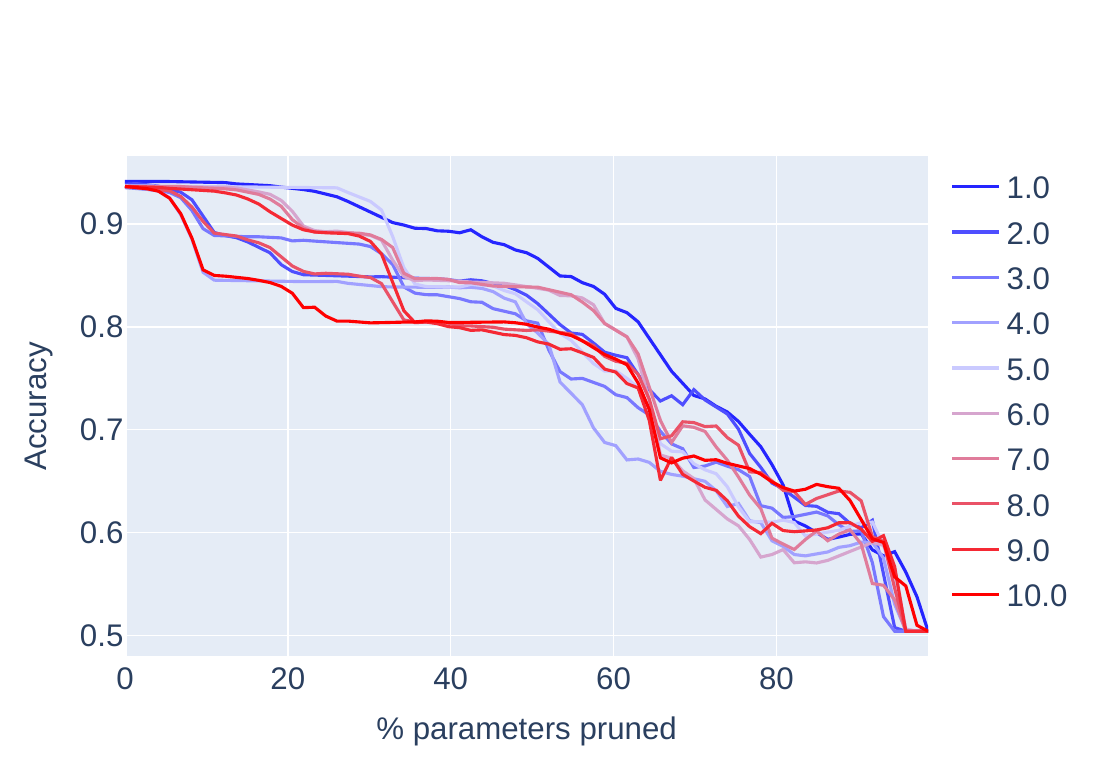}
        \caption{Acc. Layer 2}
    \end{subfigure}
    \begin{subfigure}{0.45\textwidth}
        \includegraphics[width=\linewidth]{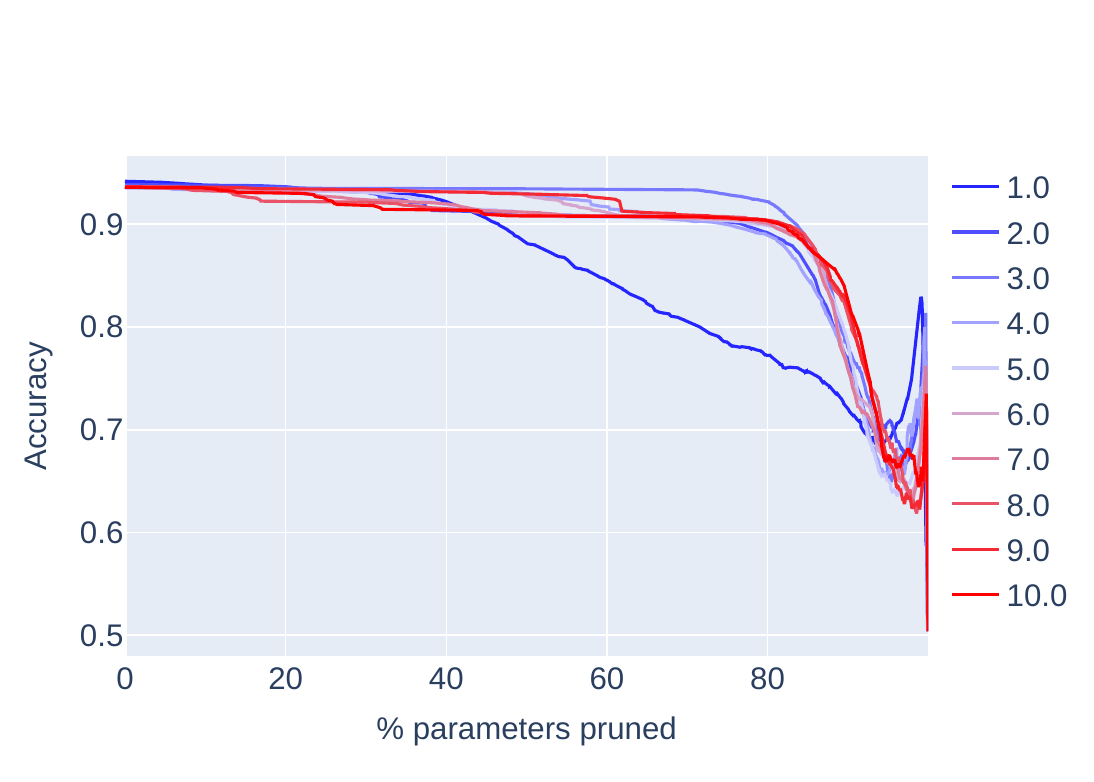}
        \caption{Acc. Layer 3}
    \end{subfigure}
    \hfill
    \begin{subfigure}{0.45\textwidth}
        \includegraphics[width=\linewidth]{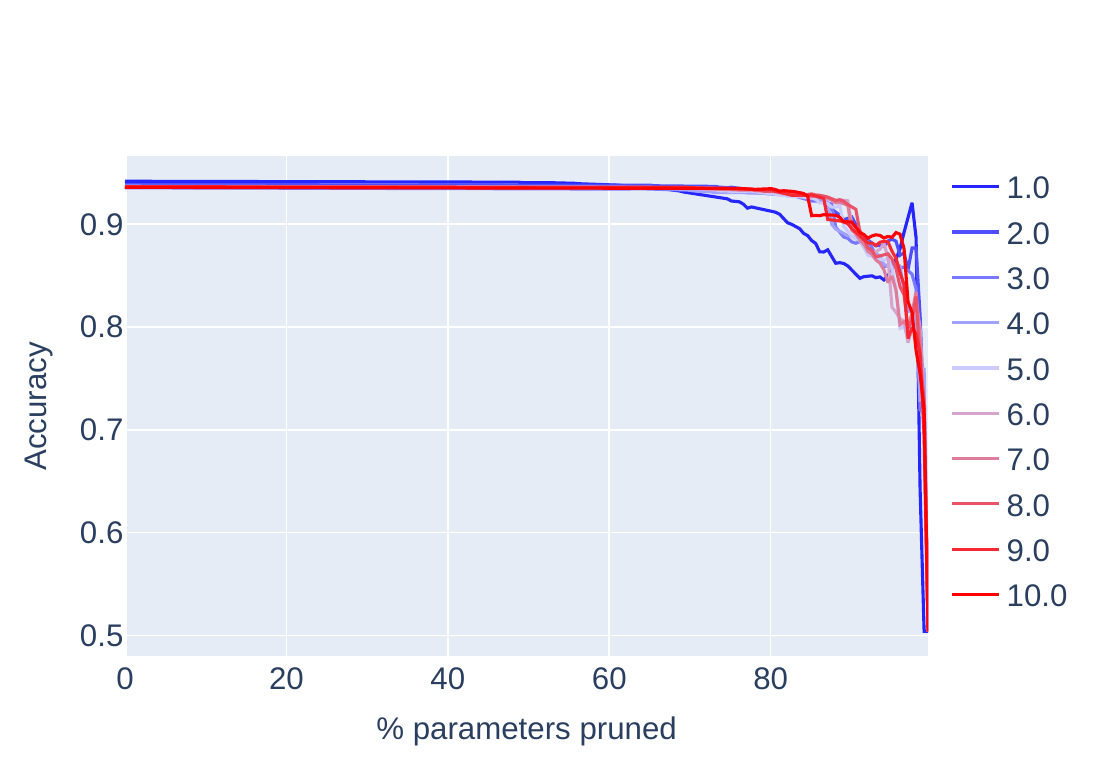}
        \caption{Acc. Layer 4}
    \end{subfigure}
    \hfill
    \caption{Accuracies of Ising network output as a function of parameters Fisher-pruned}
    \label{fig:fisher_prune_per_layer_acc}
    \end{figure}

\begin{figure}[h!]
    \centering
    \begin{subfigure}{0.45\textwidth}
        \includegraphics[width=\linewidth]{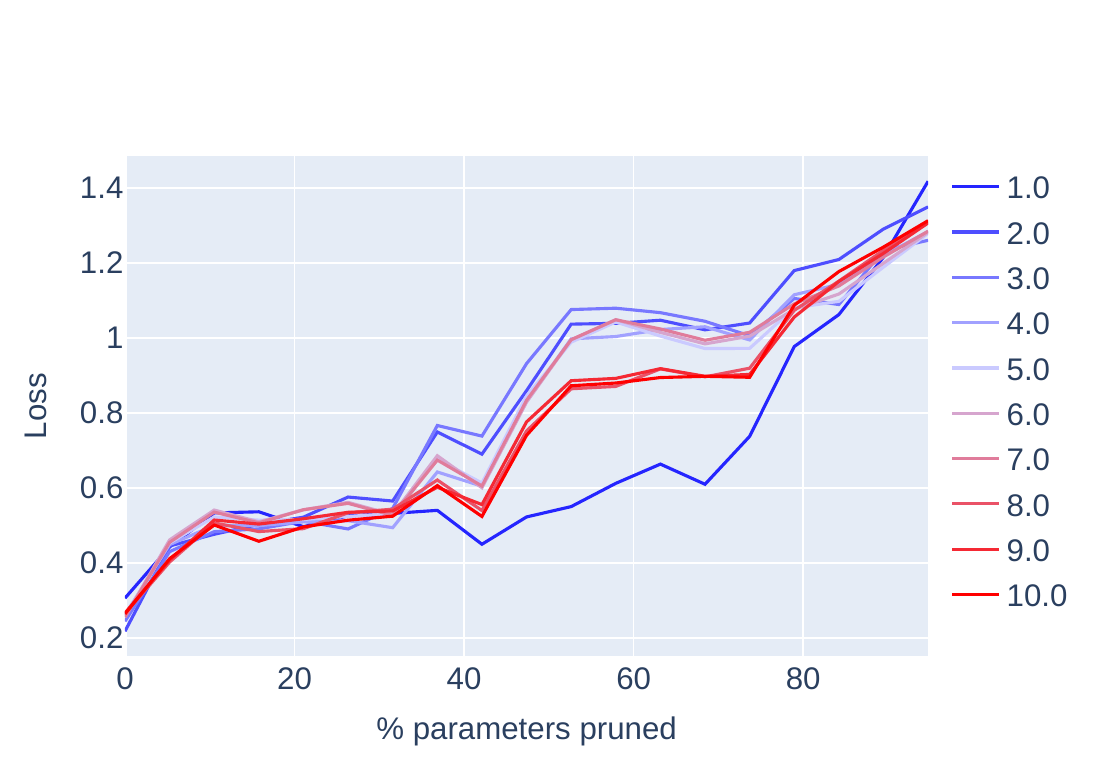}
        \caption{Loss Layer 1}
    \end{subfigure}
    \begin{subfigure}{0.45\textwidth}
        \includegraphics[width=\linewidth]{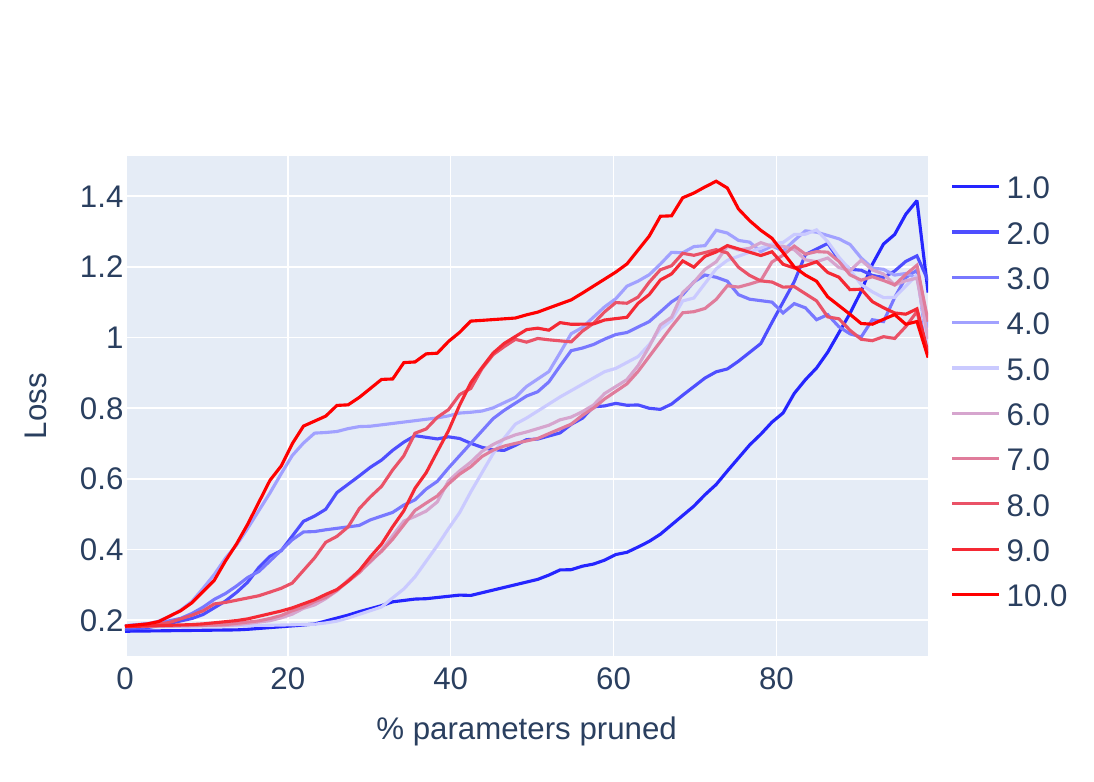}
        \caption{Loss Layer 2}
    \end{subfigure}
    \begin{subfigure}{0.45\textwidth}
        \includegraphics[width=\linewidth]{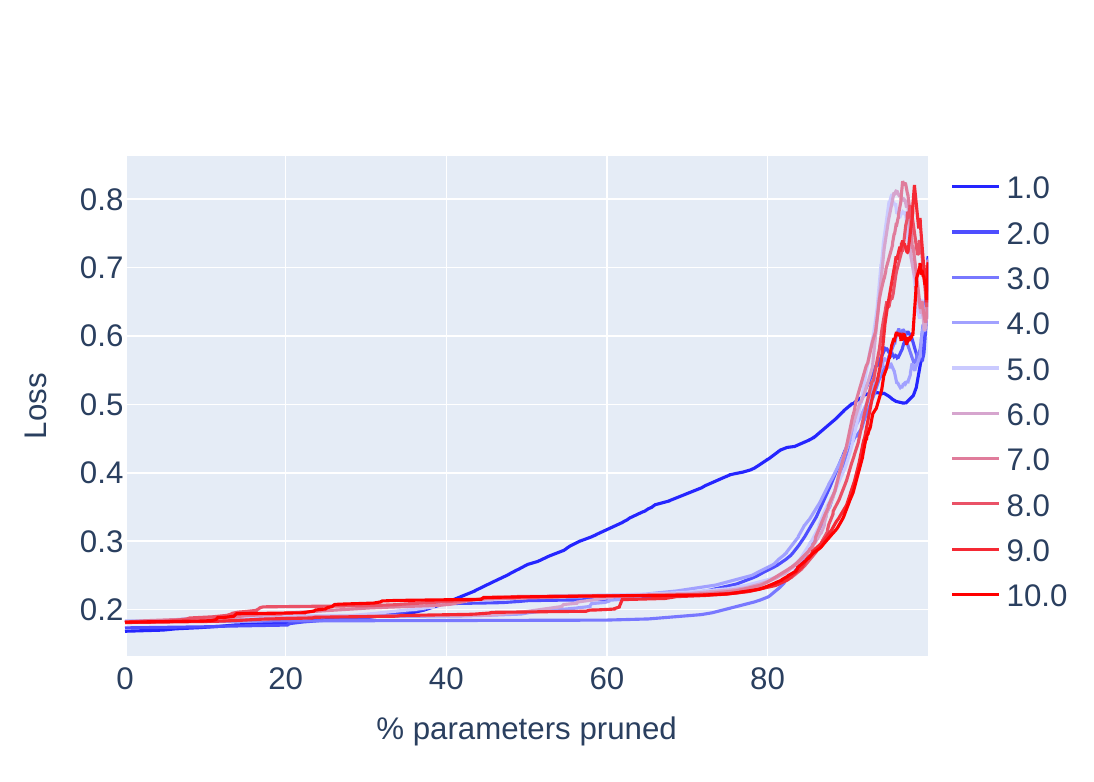}
        \caption{Loss Layer 3}
    \end{subfigure}
    \begin{subfigure}{0.45\textwidth}
        \includegraphics[width=\linewidth]{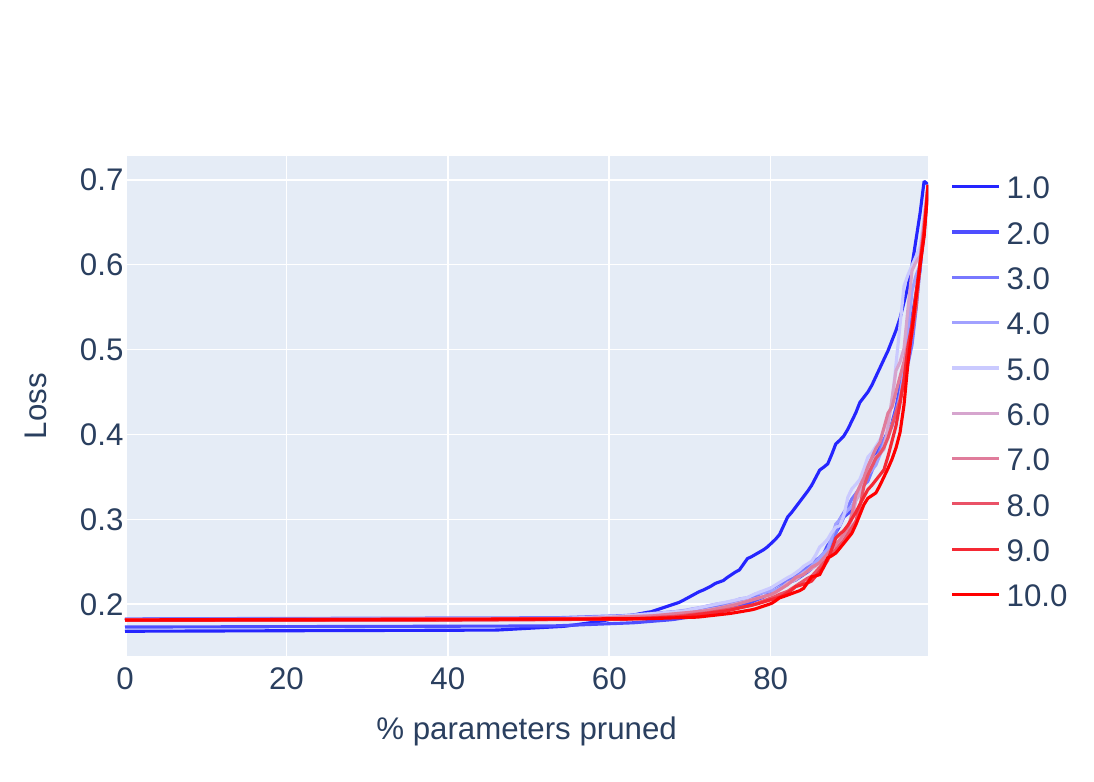}
        \caption{Loss Layer 4}
    \end{subfigure}
    \hfill
    \caption{Losses of Ising network output as a function of the number of parameters Fisher-pruned}
    \label{fig:per_layer_fisher_prune_ising}
\end{figure}

A natural question one may ask is: ``how does Fisher pruning act per-layer?" In order
to facilitate such an analysis, we propose freezing splitting the model
parameters from the whole model into two non-overlapping sets $\theta_a \sqcup
\tilde \theta_a$, where $a$ is the index of the layer to be examined. For each
model, we compute the Fisher metric $\mathcal I_{ij}(\theta)$ and systematically
remove $\theta^a_k$ in increasing order of their associated $\mathcal I_{ij}$
value. A per-layer breakdown of the Fisher pruning regime for the modular
addition problem is provided in figure \ref{fig:fisher_pruning_ma_per_layer},
whilst an analogous plots for the Ising problem are presented in figures
\ref{fig:fisher_prune_per_layer_acc} and \ref{fig:per_layer_fisher_prune_ising}.
%

\clearpage
\subsection{Other Magnitude Pruning Schemes}
\label{sub:mod_add_fisher}
In the main text, we chose to prune our networks using a particular magnitude pruning scheme in which the same percentage of weights are pruned in each layer. In Appendix F.1, we discuss the results of several pruning schemes that are based on the Fisher metric instead of the (Euclidean) magnitude of the weights. Here, we show several other magnitude pruning schemes for comparison to the Fisher-based approach. In particular, we focus on global unstructured pruning (both with and without rescaling each weights by layer norms) and layer-resolved pruning. 

\begin{figure}[htbp]
\centering
    \includegraphics[width=0.95\linewidth]{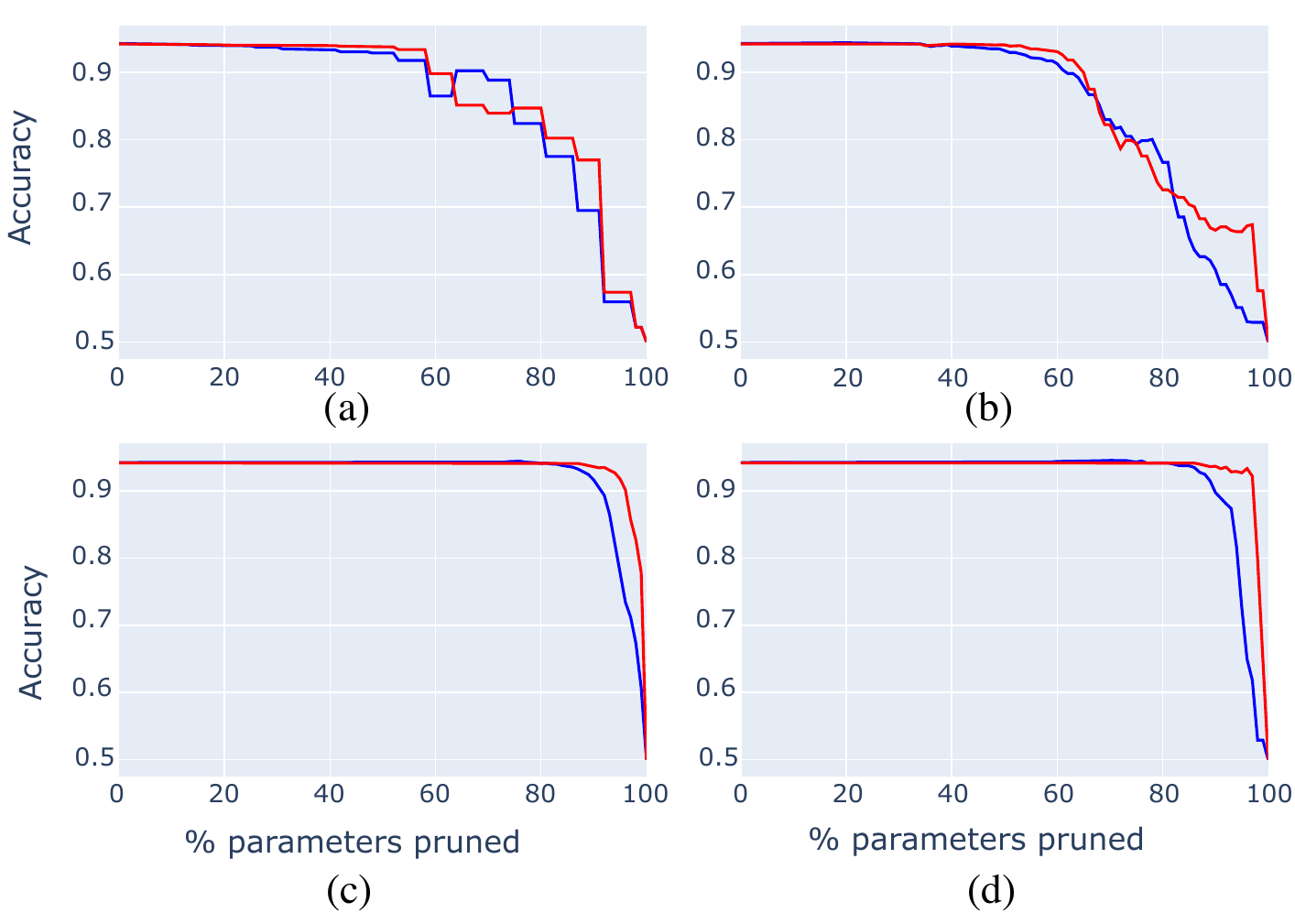}
    \caption{Different global magnitude pruning schemes for the two tasks, averaged over 8 seeds and using two weight multipliers. The results of pruning the Ising network are shown on the top row, while the analogous results for modular addition are on the bottom row. (a,d) Parallel pruning in all layers (as in main text) (b,e) Global unstructured pruning, and (c, f) Global unstructured pruning with each weight rescaled by the norm of all weights in its layer.}
    \label{fig:global_mag_pruning}
\end{figure}

In fig.~\ref{fig:global_mag_pruning}, we show the results of the different global pruning schemes for both tasks. The pruning scheme in the main text (parallel pruning in each layer) is shown in the first column, naive global unstructured pruning is shown in the second column, and global unstructured pruning done after rescaling all weights by their respective layer norms is shown in the third column. Using naive global unstructured pruning, it appears that a larger fraction of the network can be pruned at no loss in accuracy compared to the other strategies. However, this is an artefact of the large number of insignificant weights in the last fully connected layer.

\begin{figure}[t!]
\centering
    \includegraphics[width=0.9\linewidth]{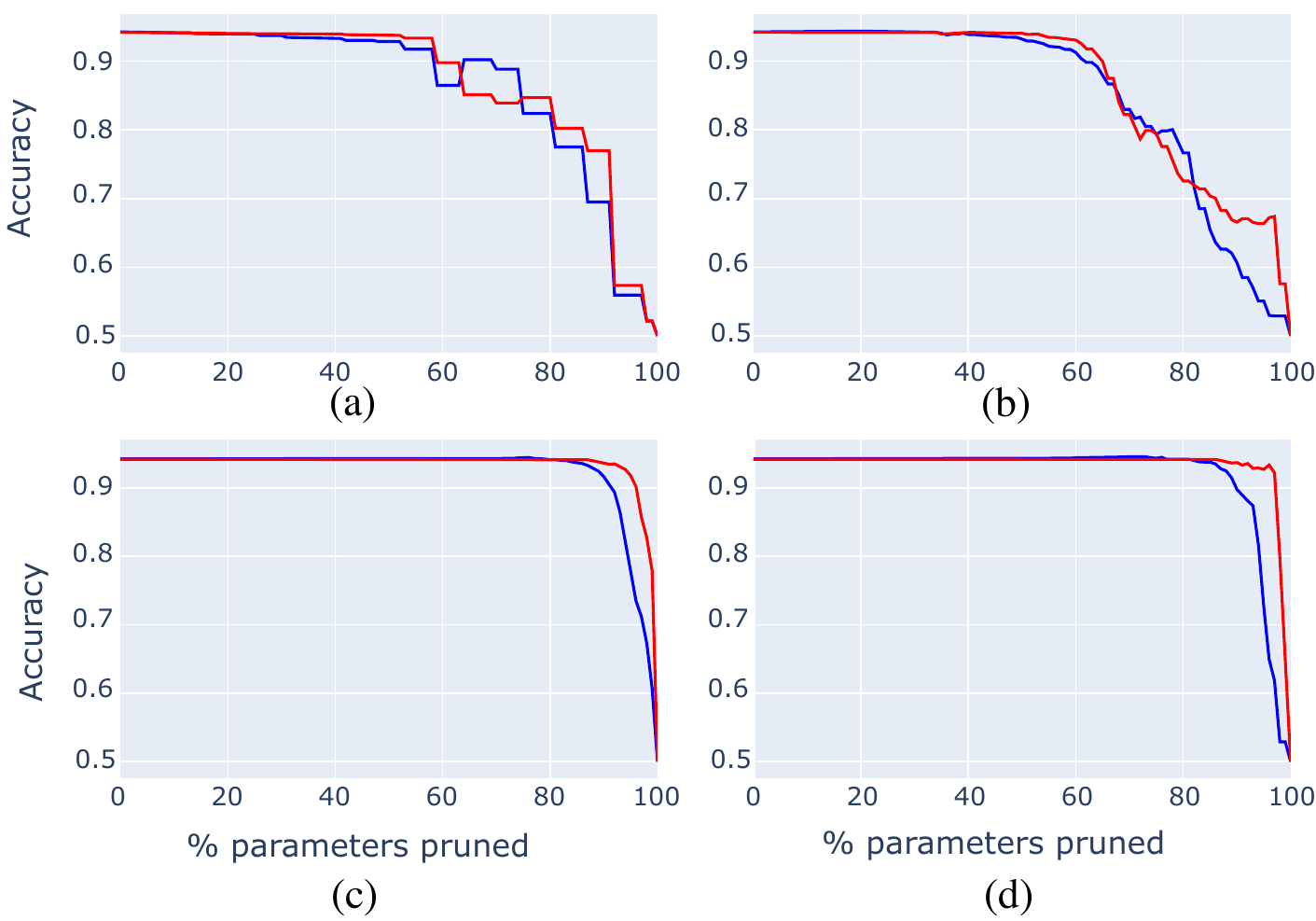}
    \caption{Layer-resolved magnitude pruning of the weights in the Ising network. Results of pruning only (a) The first CNN, (b) The second CNN, (c) The first fully connected layer, and (d) The second fully connected layer.}
    \label{fig:mag_ising_layer_pruning}
\end{figure}

\begin{figure}[h!]
\centering
    \includegraphics[width=0.9\linewidth]{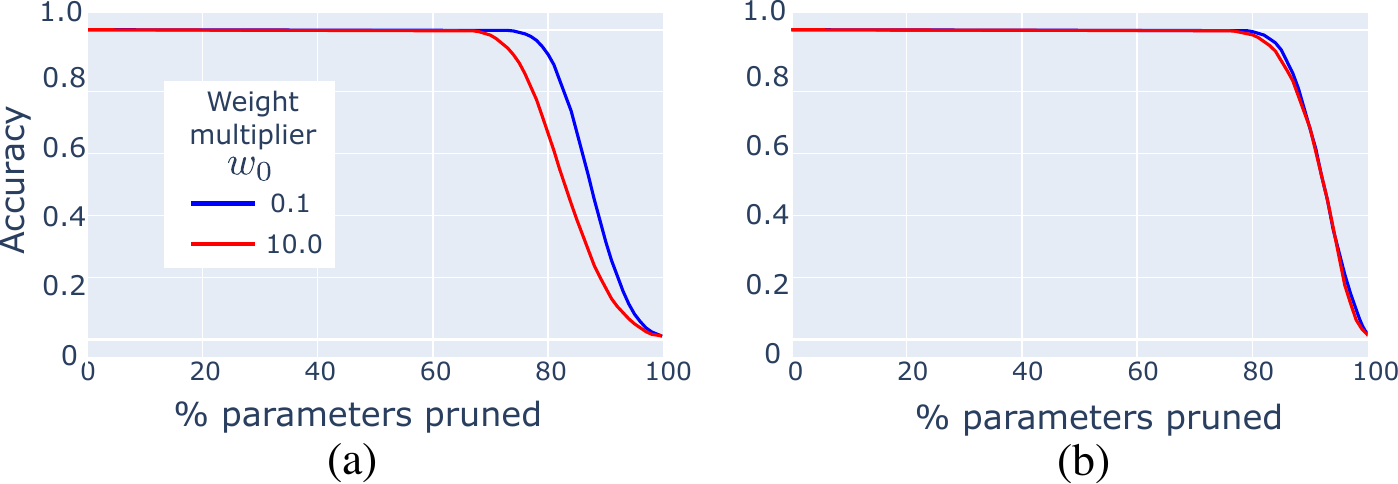}
    \caption{Layer-resolved magnitude pruning of the weights in the modular addition network. Results of pruning only (a) The first fully connected layer and (b) the second fully connected layer.}
    \label{fig:mag_modadd_layer_pruning}
\end{figure}

Since unstructured pruning removes the smallest weights independently of where they are in the network, it results in a behaviour very similar to simply pruning weights in the last layer, where the majority of the smallest weights reside. The network is relatively resilient to this kind of pruning, while pruning its initial convolutional layers, which have significantly fewer parameters, is more drastic. These observations can be verified by comparing fig.~\ref{fig:global_mag_pruning} to the results of pruning one layer at a time in the Ising network, shown in fig.~\ref{fig:mag_ising_layer_pruning}. We show the same layer-resolved pruning scheme for modular addition in fig.~\ref{fig:mag_modadd_layer_pruning}. 

\clearpage
\section{Dynamics}
\subsection{Feature development through interpretability}
\begin{figure}[htbp]
    \includegraphics[width=1\linewidth]{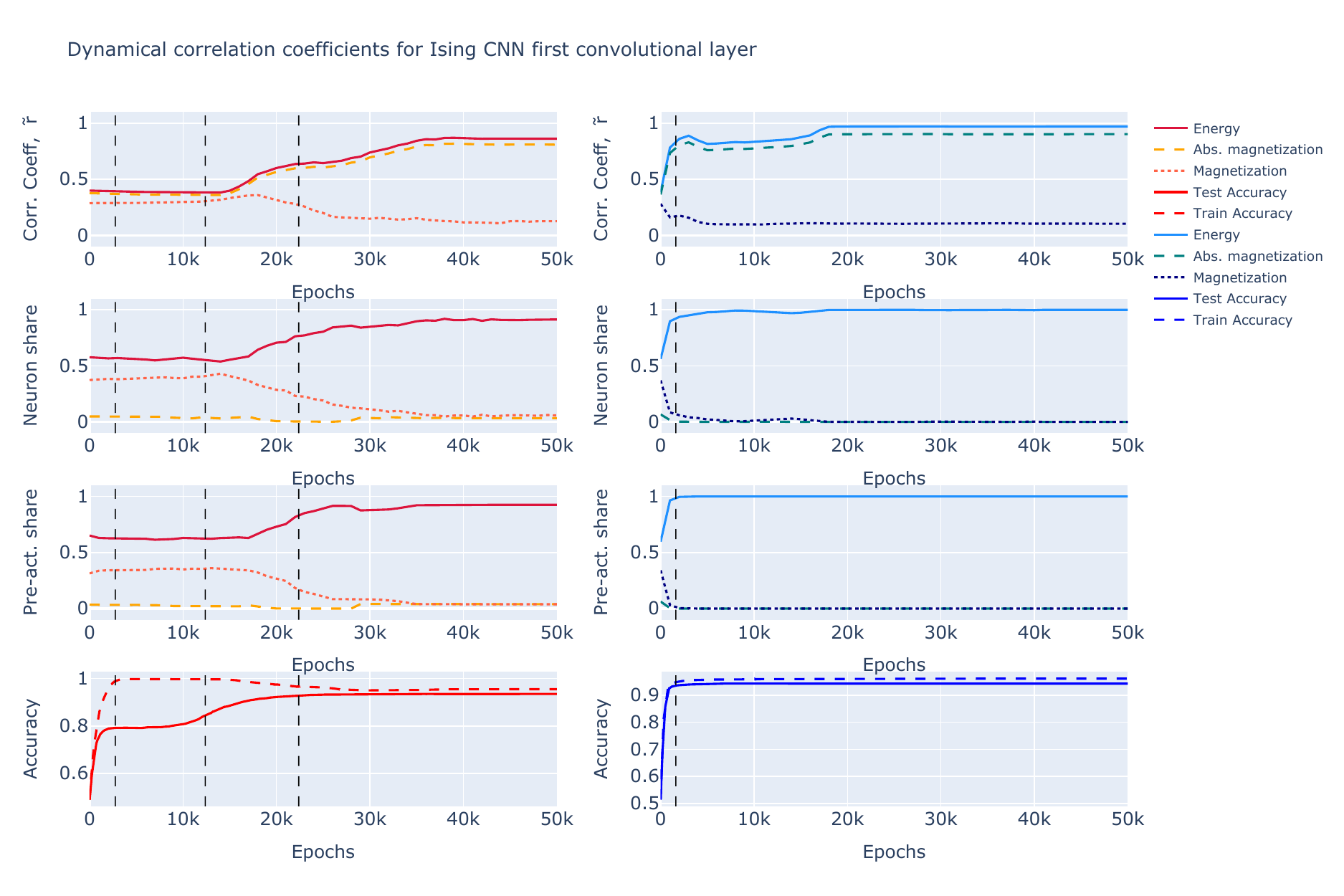}
    \caption{A summary of the correlation coefficients for the different measures across seeds and the first convolutional layer for the grokking case in Ising.}
    \label{fig:ising_dyn_corr_1cnn}
\end{figure}

\begin{figure}[htbp]
    \includegraphics[width=1\linewidth]{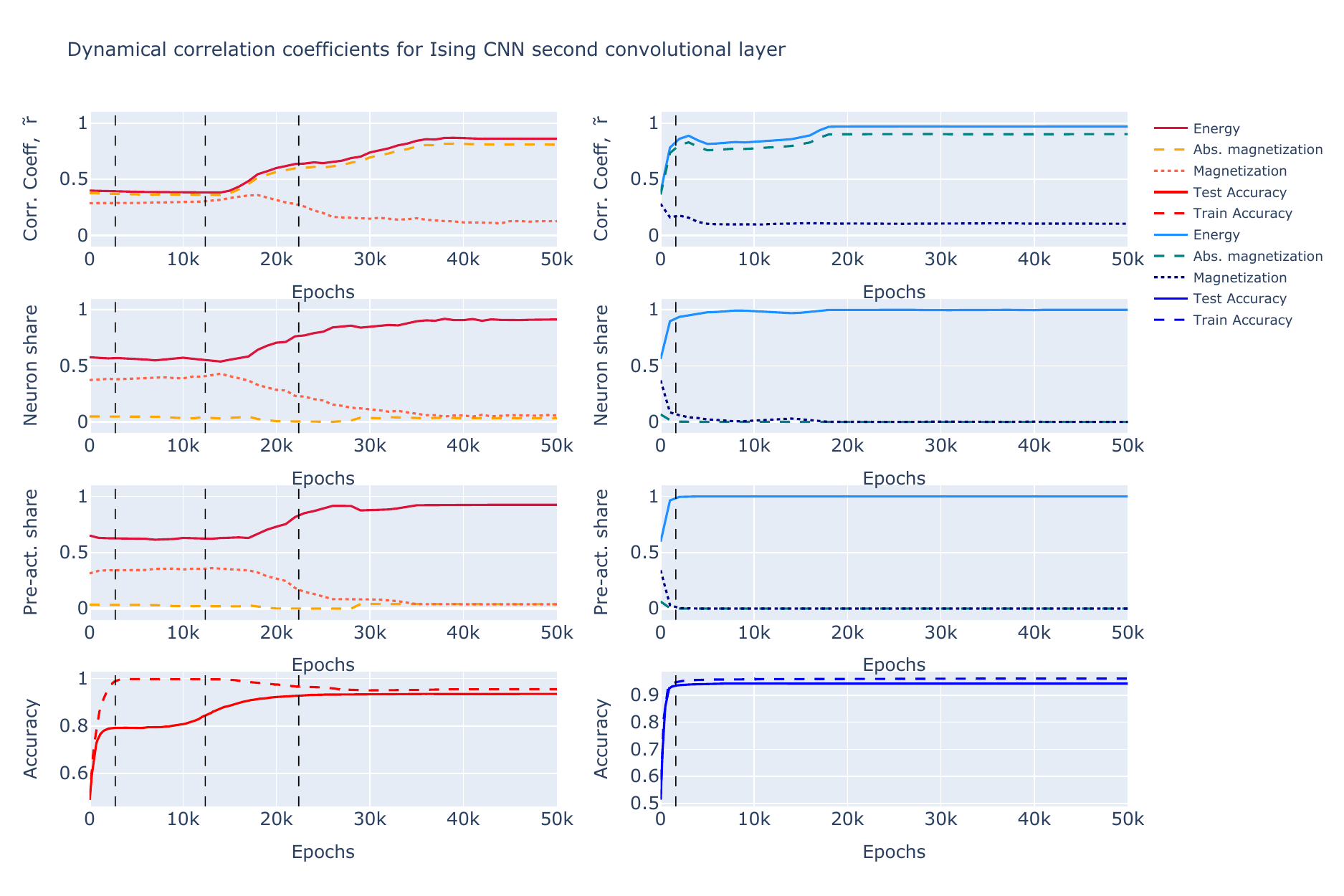}
    \caption{A summary of the correlation coefficients for the different measures across seeds and the second convolutional layer for the grokking case in Ising.}
    \label{fig:ising_dyn_corr_2cnn}
\end{figure}

\begin{figure}[htbp]
    \includegraphics[width=1\linewidth]{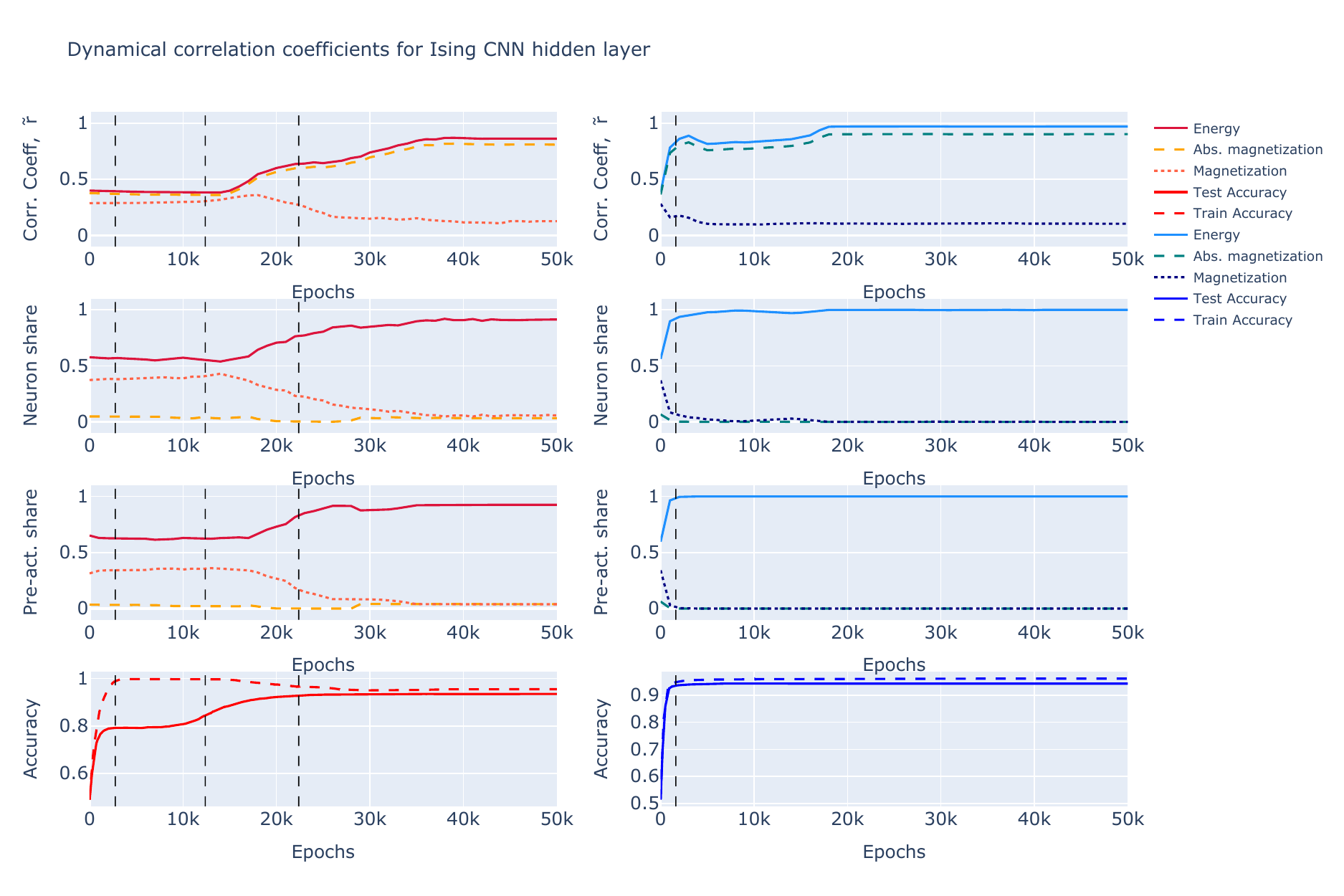}
    \caption{A summary of the correlation coefficients for the different measures across seeds and the fully connected layer for the grokking case in Ising.}
    \label{fig:ising_dyn_corr_fc}
\end{figure}

We provide here additional measures for feature development in the Ising task, across layers. In addition to the pre-activation weighted volume that we considered in the main text, we also consider the fraction of neurons which have their highest correlation coefficient with the each measure, and the share of the total pre-activation accounted for by each measure. We provide these measures for the first CNN layer in \cref{fig:ising_dyn_corr_1cnn}, for the second CNN layer in \cref{fig:ising_dyn_corr_2cnn}, and for the last layer in \cref{fig:ising_dyn_corr_fc}.

\subsection{Comparing Fisher and Euclidean Model Space Geometry}
One may wonder how the novel FIM-inspired dynamical measures would act in the
limit that the model space is assumed Euclidean, such that the FIM is not
computed or used.

Across the presented plots related to these measures each were also regenerated
using a Euclidean version of the measure\footnote{In fact, this paper's
respective codebase allows full optimised functionality for this should one
prefer to use it.}. Overall, the observed behaviour was similar, with the FIM
versions of the measures showing the analysed features to be more pronounced.
Motivating their use as improvements on more traditional assumptions, and better
theoretically motivated for the analysis of model space.

\begin{figure}[h!]
    \centering
    \begin{subfigure}{0.45\textwidth}
        \centering
        \includegraphics[trim=0 4 0 0, clip, width=0.98\textwidth]{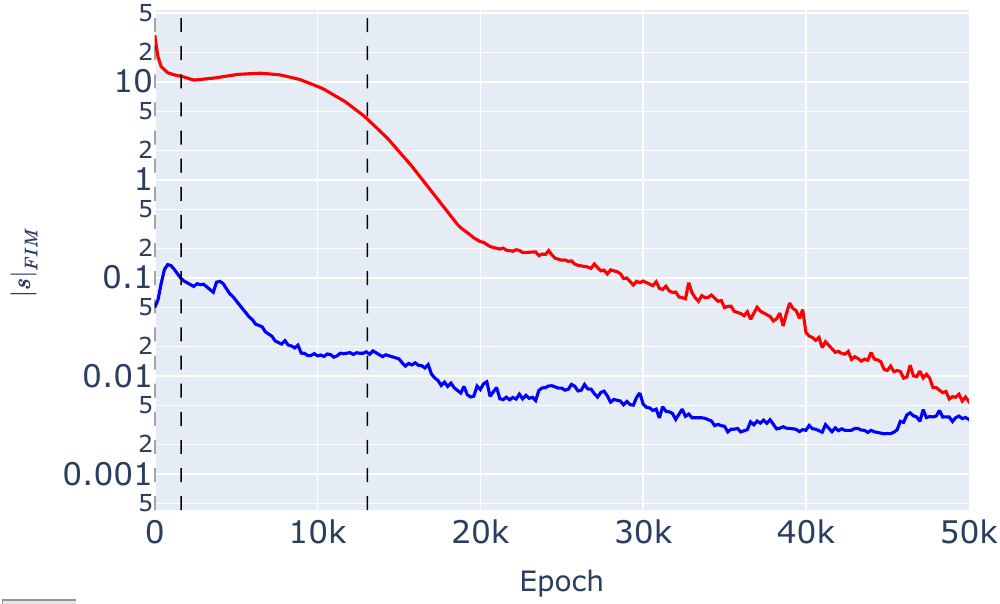}
        \caption{FIM Measure}
    \end{subfigure} 
    \begin{subfigure}{0.45\textwidth}
        \centering
        \includegraphics[trim=0 1 0 0, clip, width=0.98\textwidth]{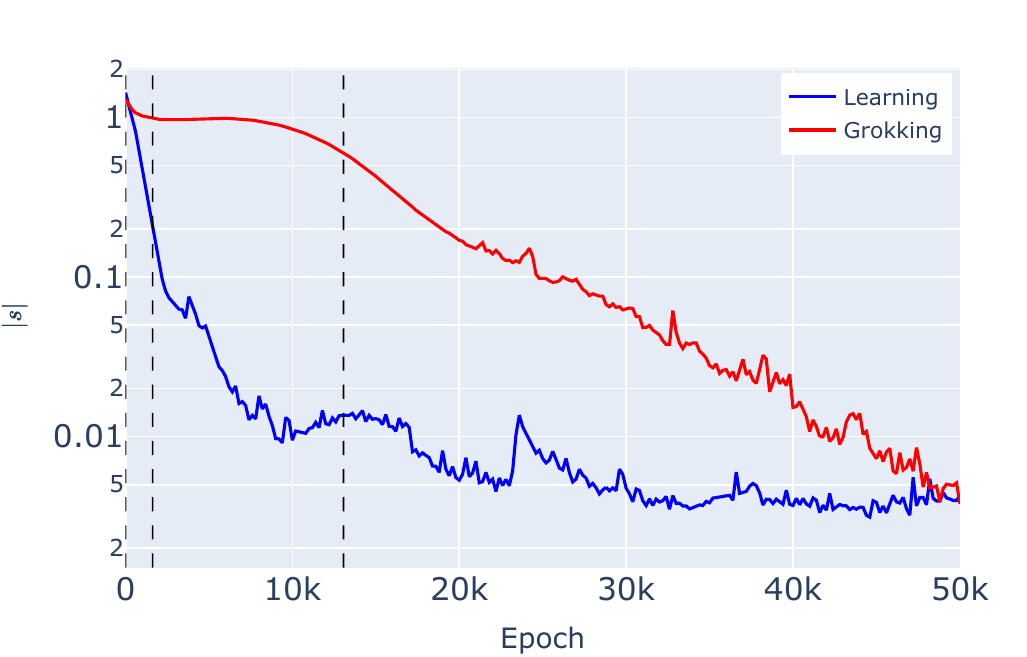}
        \caption{Euclidean Measure}
    \end{subfigure} 
    \caption{Step magnitudes of the models' trajectories through model space during training. Measures computed using the FIM $|s|_{FIM}$, or simply with the Euclidean assumption $|s|$, and averaged over the 10 runs.}
    \label{fig:dynamics_mag_FIMvsEuc}
\end{figure}

As an example, in \cref{fig:dynamics_mag_FIMvsEuc}, the step magnitudes for the
Ising task are shown suing the FIM-magnitude measure (as above in
\cref{fig:dynamics_mag_ising}) and also with the simplified Euclidean version.
One may observe the FIM measure better distinguishes between the grokking and
steady Learning regimes, and the `bump' feature as grokking begins is more
pronounced.

\clearpage
\subsection{Fisher step magnitudes}
Considering the spectrum of diagonal entries averaged over the 10 seed runs,
shown in appendix \ref{app:fim}, the spectra exhibit a common mass gap, such
that a subset of the parameters have relatively negligible importance. This
subset of parameters are labelled `sloppy', whilst the complement subset are
labelled `stiff'. Motivated by this mass gap, it is interesting to consider
splitting the model space into two subspaces\footnote{Here the full model space
is treated as a product space and with the stiff $\times$ sloppy subspaces
independent.}, one parameterised by the stiff parameters and one by the sloppy.
The model space positions and FIMs can be split onto these subspaces, and the
equivalent novel trajectory analysis measures reduced to be defined on each of
these subspaces also.

In both tasks a FIM diagonal threshold of 0.1 is used to discern whether a
network parameter is stiff or sloppy. The FIM diagonal is considered for the
final model after training, and parameters are assigned to be stiff (or sloppy)
when $g^{FIM}_{ii} \geq 0.1$ (or $< 0.1$). These assignments define the
subspaces, which are then fixed and tracked back through the training.

\begin{figure}[h!]
    \centering
    \begin{subfigure}{0.45\textwidth}
        \centering
        \includegraphics[trim=0 3 0 0, clip, width=0.94\textwidth]{plots/AvgMagnitues_FIM_Full_ising.pdf}
        \caption{Step Magnitude - Ising}\label{fig:dynamics_mag_ising}
    \end{subfigure} 
    \begin{subfigure}{0.45\textwidth}
        \centering
        \includegraphics[trim=0 19 0 0, clip, width=0.99\textwidth]{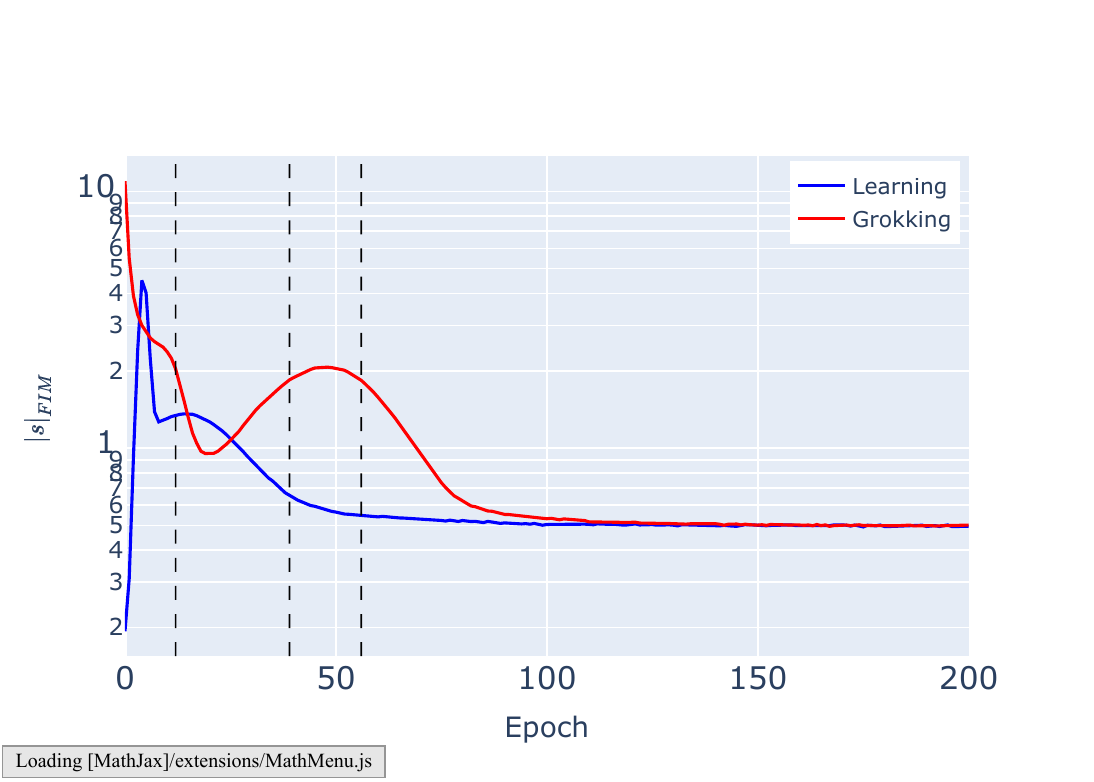}
        \caption{Step Magnitude - ModAdd}\label{fig:dynamics_mag_modadd}
    \end{subfigure}\\
    \begin{subfigure}{0.45\textwidth}
        \centering
        \includegraphics[trim=0 19 0 0, clip, width=0.99\textwidth]{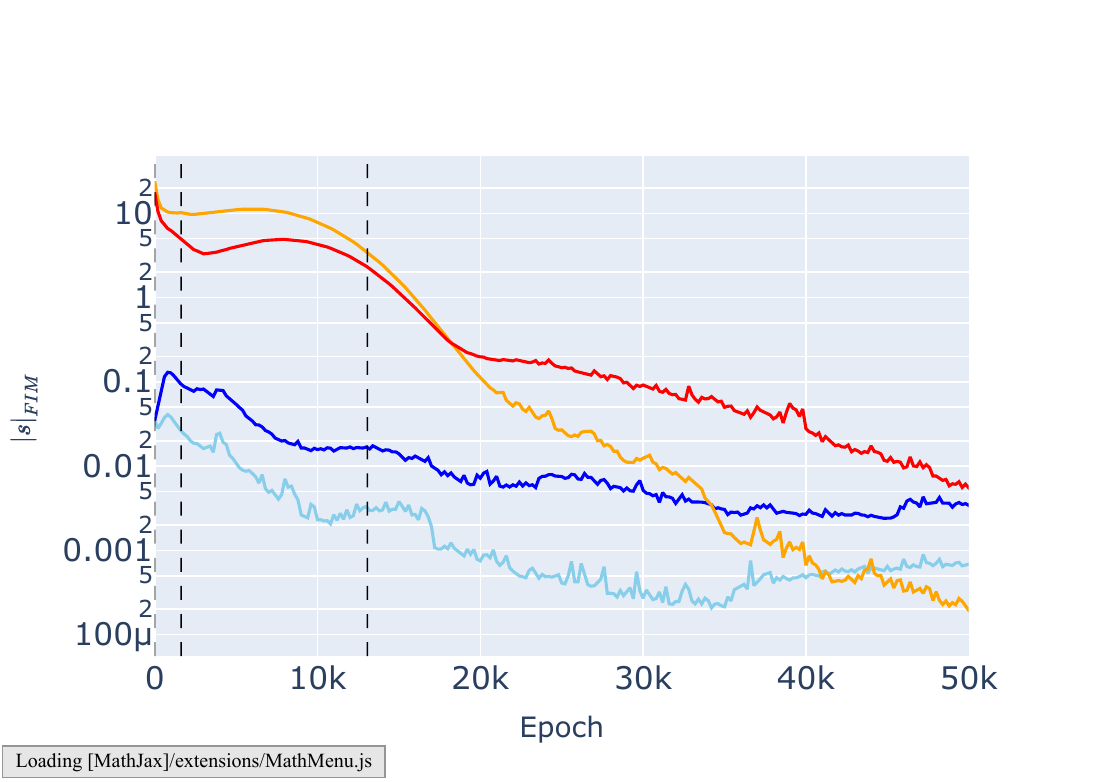}
        \caption{Split Step Magnitude - Ising}\label{fig:dynamics_splitmag_ising}
    \end{subfigure} 
    \begin{subfigure}{0.45\textwidth}
        \centering
        \includegraphics[trim=0 19 0 0, clip, width=0.99\textwidth]{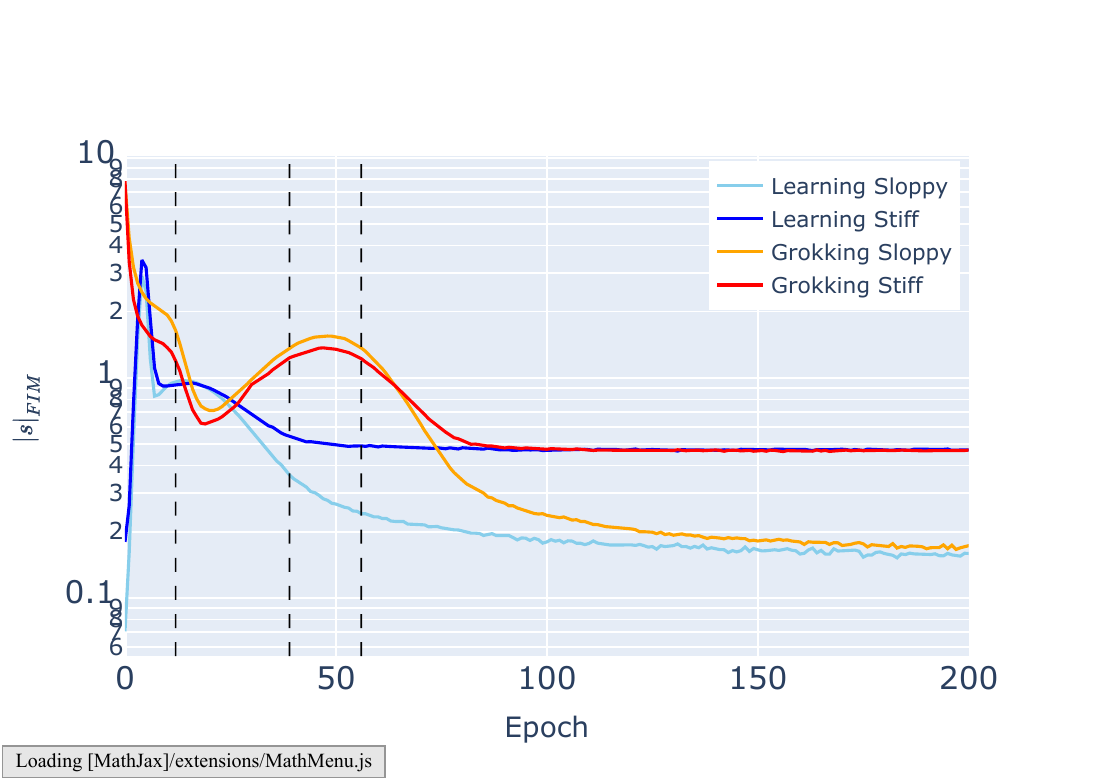}
        \caption{Split Step Magnitude - ModAdd}\label{fig:dynamics_splitmag_modadd}
    \end{subfigure}
    \caption{Step magnitudes, $|s|_{FIM}$, of the models' trajectories through model space during training. Measures computed using the FIM, averaged over the 10 runs. Plots show the behaviour of the full model (a) \& (b), as well as the sloppy and stiff submodels (c) \& (d); for both the learning and grokking regimes; for both the Ising (a) \& (c), and ModAdd (b) \& (d) investigations.}
    \label{fig:dynamics_mag}
\end{figure}

The first of the two novel dynamical measures introduced in this work considers
the training step FIM-magnitudes. Averaged over the 10 seed runs, these are
shown for both the Ising and modular addition tasks, with and without the
stiff-sloppy splitting, and for both the grokking and steady learning regimes,
in \cref{fig:dynamics_mag}.

In both tasks and both regimes the step magnitude generally decreases over the
training as the model improves the loss gradient drops and the steps become
smaller. The grokking regimes start with higher step magnitude values since
their initialisation uses larger weight multipliers, but towards the end of
training the step magnitudes are of the same order between the grokking and
learning regimes. The noticeable difference is that in the grokking regime the
step magnitude has a significant bump in value at the start of the grokking
period, where the step magnitude continues to increase during this period. Where
a similar behaviour seems to occur earlier in the Learning regime this is much
more pronounced in the grokking case, and may be attributed to the optimizer
finding an optimal trajectory for improvement which allows the momentum to grow
and step magnitude to increase. 

\begin{figure}[h!]
    \centering
    \begin{subfigure}{0.45\textwidth}
        \centering
        \includegraphics[trim=0 19 0 0, clip, width=0.98\textwidth]{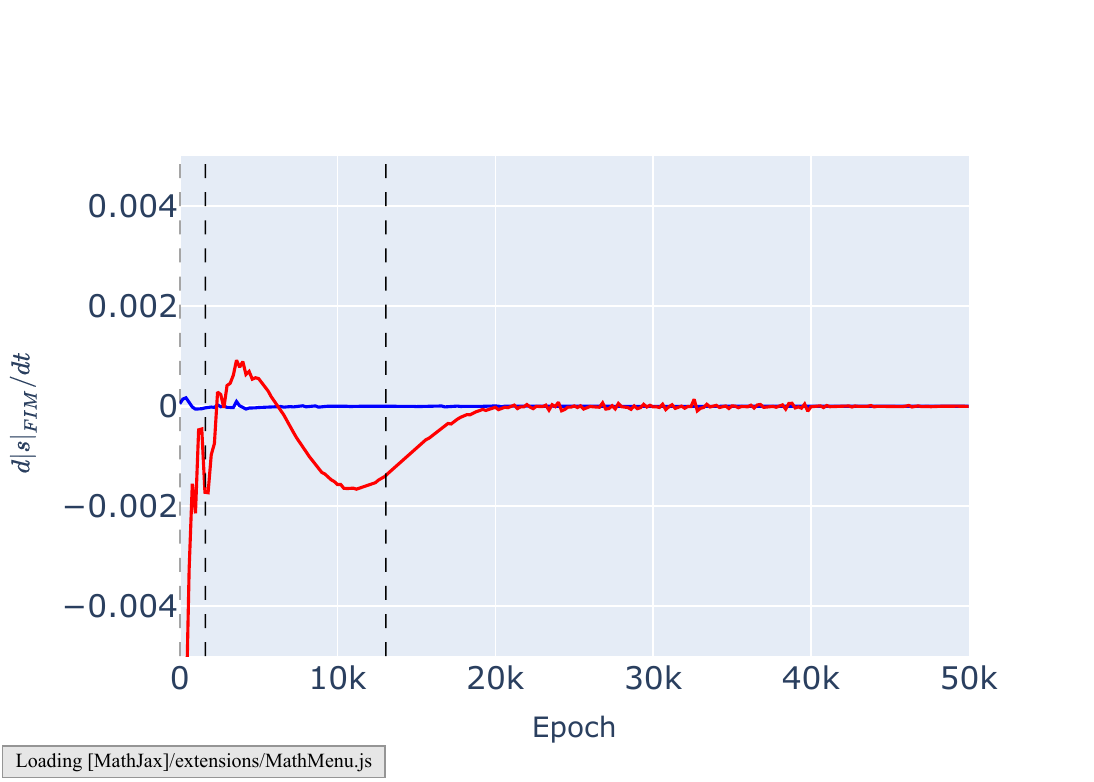}
        \caption{Ising}
    \end{subfigure} 
    \begin{subfigure}{0.45\textwidth}
        \centering
        \includegraphics[trim=0 19 0 0, clip, width=0.98\textwidth]{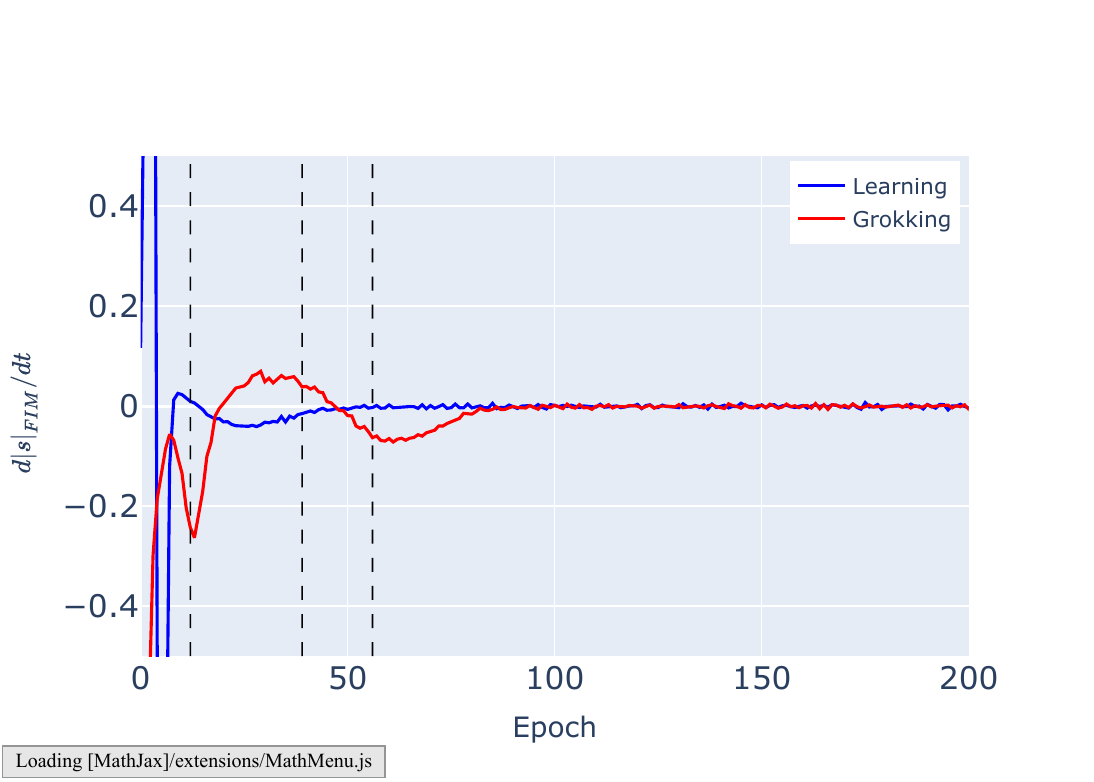}
        \caption{ModAdd}
    \end{subfigure} 
    \caption{Discrete gradients of FIM steps, $d|s|_{FIM}/dt$, along the model space learning trajectory, averaged over the 10 seed runs. Plots showing both the learning and grokking regimes; for both the Ising (a), and ModAdd (b) investigations. The axes are restricted to highlight the roots.}
    \label{fig:dynamics_mag_grad}
\end{figure}

From the previous observations, the bump in the FIM step magnitude appears to be
a good indicator for the grokking period, this we can make more explicit by
plotting the discrete gradient of this graph and considering the roots of the
line, as shown in \cref{fig:dynamics_mag_grad}. Both tasks show 2 significant
roots where the plot crosses gradients of zero before plateauing towards zero
gradient, which may be used to define the start and end of the grokking regimes.
For the Ising task the first root is between the epoch samples 2200 and 2400,
whilst the second root is between 6200 and 6400. For the modular arithmetic task
the first root is between epochs 19 and 20, whilst the second root is between
epochs 46 and 47. These epoch ranges for the grokking period roughly agree with
those specified previously with the measure defined in this paper, supporting
this as an alternative mechanism to define the grokking period.

Considering how the step magnitude analysis splits into the stiff and sloppy
subspaces, on both tasks the stiff step magnitudes converge to higher values
than the sloppy step magnitudes. Indicating that the optimizer identifies these
stiff directions and directs the parameter updates to only change those which
have a significant effect on the model. The bump increase in step magnitude
during the grokking period is exhibited in both the stiff and sloppy subspaces
indicating this is likely a universal feature of grokking across the parameters.
Furthermore, in the grokking regime the stiff and sloppy lines cross, indicating
that when a model is set up to grok the optimizer will first prioritize changing
sloppy features (where this line starts higher) and then will swap after the
grokking period has finished to prioritize stiff features.

\begin{figure}[h!]
    \centering
    \begin{subfigure}{0.45\textwidth}
        \centering
        \includegraphics[trim=0 19 0 0, clip, width=0.98\textwidth]{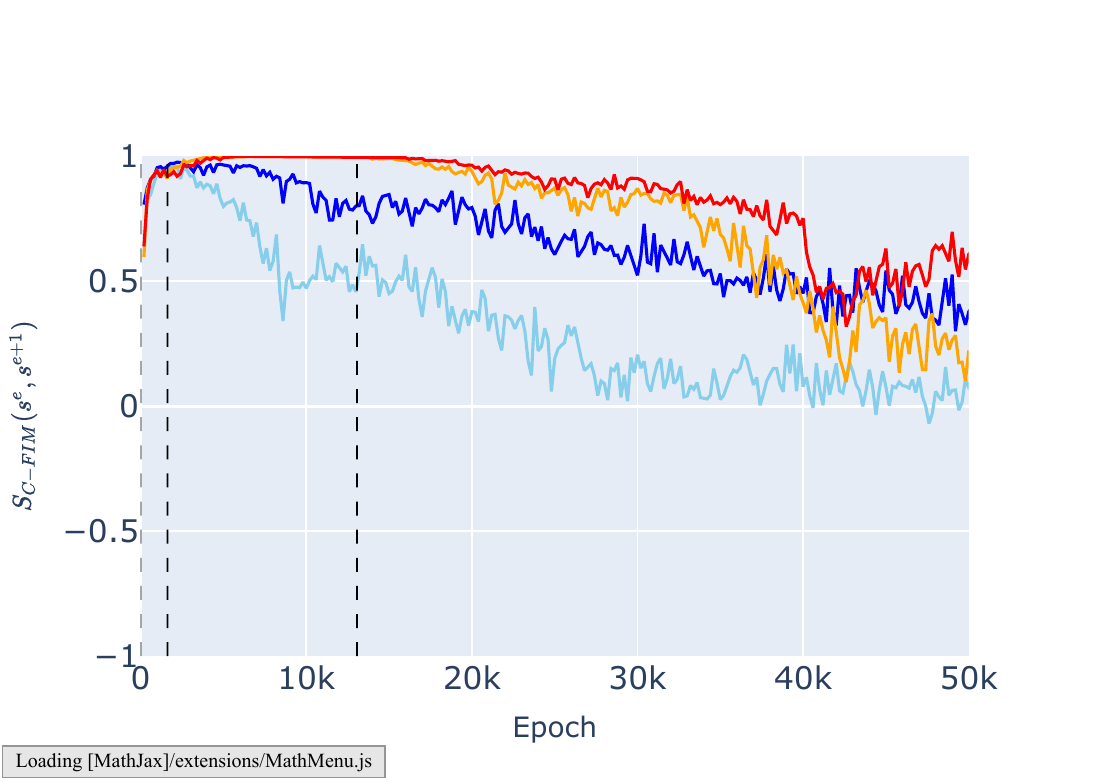}
        \caption{Ising}
    \end{subfigure} 
    \begin{subfigure}{0.45\textwidth}
        \centering
        \includegraphics[trim=0 19 0 0, clip, width=0.98\textwidth]{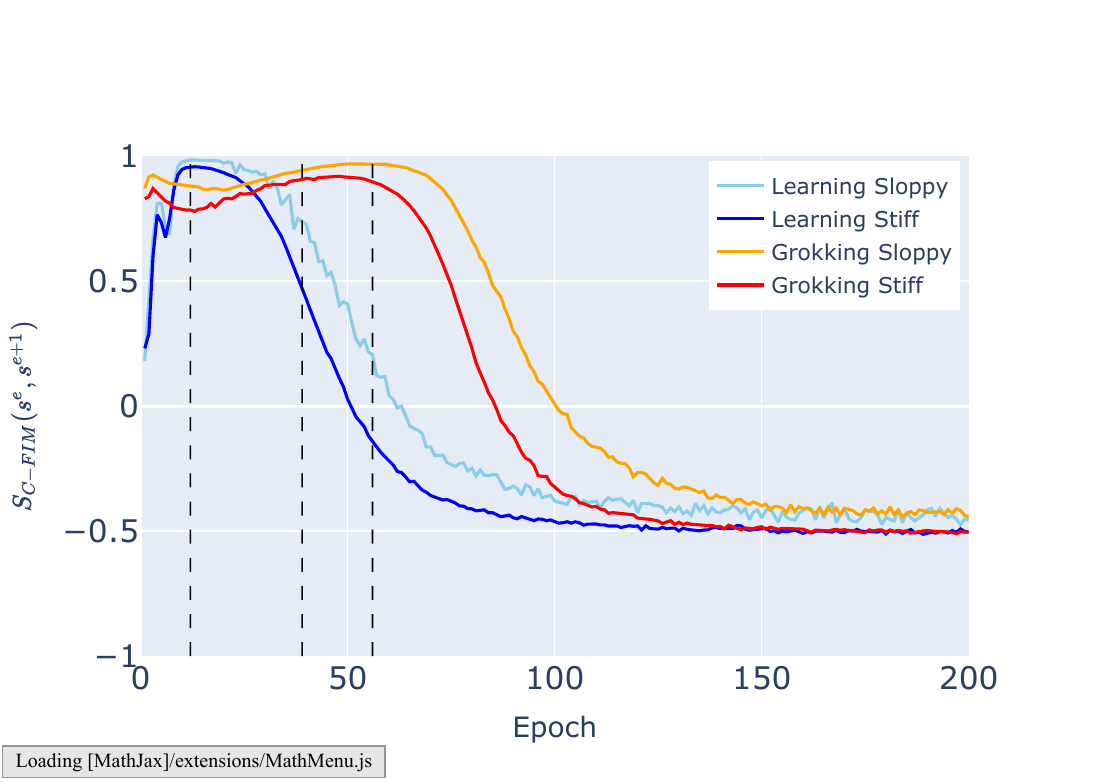}
        \caption{ModAdd}
    \end{subfigure} 
    \caption{Split FIM-cosine similarities between consecutive steps, $S_{C-FIM}(s^e,s^{e+1})$, for the sloppy and stiff submodels. Plots showing both the learning and grokking regimes; for both the Ising (a), and ModAdd (b) investigations.}
    \label{fig:dynamics_cosine_split}
\end{figure}

The second novel dynamical measure, considered primarily in the main body of the
text, may also be split to study the trajectories on the sloppy and stiff
submanifolds. Where here we only consider $S_{C-FIM}(s^e,s^{e+1})$ for step
vectors at consecutive epoch samples, as in the main body but without the origin
trajectory comparison. These equivalent plots are shown, averaged over the 10
seed runs, for both the Ising and modular addition tasks, with the stiff-sloppy
splitting, and for both the grokking and steady learning regimes, in
\cref{fig:dynamics_cosine_split}.

In the grokking regime, the behaviour is similar between the subspaces, whereas
for the learning regime in the Ising task, the stiff subspace has consistently
higher values - indicating that the grokking regime treats the model parameters
more symmetrically in the updates. 

\clearpage

\clearpage

\pagebreak

\end{document}